\documentclass[9.5pt,journal,compsoc]{IEEEtran}
\usepackage[table,xcdraw]{xcolor}

\usepackage{tikz}
\usetikzlibrary{matrix,shapes,arrows,fit,tikzmark}

\ifCLASSOPTIONcompsoc
  \usepackage[nocompress]{cite}
\else
  \usepackage{cite}
\fi

\usepackage{amsmath}
\usepackage{amssymb}
\usepackage[linguistics]{forest}

\ifCLASSOPTIONcompsoc
 \usepackage[font=footnotesize,labelfont=sf,textfont=sf]{subcaption}
\else
 \usepackage[font=footnotesize]{subcaption}
\fi
\usepackage{caption}

\usepackage{algorithm}
\usepackage[noend]{algpseudocode}

\usepackage{url}
\usepackage{amsmath}
\usepackage{booktabs}
\usepackage{soul}

\usepackage{color}
\definecolor{sarahcolor}{rgb}{0.858,0.188,0.478}
\definecolor{marccolor}{rgb}{0.45, 0.68, 0.0}

\newcommand{\review}[1]{#1}
\newcommand{\matt}[1]{#1}
\newcommand{\reviewfinal}[1]{#1} %

\newcommand{\myparagraph}[1]{\noindent\textbf{#1}}

\newcommand{\boldspacepar}[1]{\par\smallskip\noindent\textbf{#1}}
\newcommand{\emphspacepar}[1]{\par\smallskip\noindent\emph{#1}}
\newcommand{\NA}{n/a}

\usepackage{pifont}%
\newcommand{\xmark}{\ding{55}}%

\usepackage[normalem]{ulem}

\newcommand{\redline}[1]{{\color{red}\uline{{\color{black}#1}}}}
\newcommand{\greenline}[1]{{\color{green}\uline{{\color{black}#1}}}}

\definecolor{blue}{HTML}{4fc3f7}
\definecolor{red}{HTML}{d32f2f}
\definecolor{green}{HTML}{66bb6a}
\definecolor{yellow}{HTML}{ffeb3b}
\definecolor{purple}{HTML}{9575cd }

\definecolor{heatmap_green}{HTML}{388e3c}
\definecolor{heatmap_yellow}{HTML}{fff176}
\definecolor{heatmap_red}{HTML}{e74c3c}
\newcommand\marktopleft[1]{%
    \tikz[overlay,remember picture] 
        \node (marker-#1-a) at (0.5em,.1em) {};%
}
\newcommand\markbottomright[4]{%
    \tikz[overlay,remember picture] 
        \node (marker-#1-b) at (-.3em,.5em) {};%
    \tikz[overlay,remember picture,inner sep=3pt]
        \node[draw=#2,rounded corners,fit=(marker-#1-a.north west) (marker-#1-b.south east)] {};%
}

\newcommand\marktopleftTIGHT[1]{%
    \tikz[overlay,remember picture] 
        \node (marker-#1-a) at (0.8em,0em) {};%
}
\newcommand\markbottomrightTIGHT[4]{%
    \tikz[overlay,remember picture] 
        \node (marker-#1-b) at (-0.8em,.6em) {};%
    \tikz[overlay,remember picture,inner sep=3pt]
        \node[draw=#2,rounded corners,fit=(marker-#1-a.north west) (marker-#1-b.south east)] {};%
}

\makeatletter
\newcommand{\ccell}[3][]{%
  \kern-\fboxsep
  \if\relax\detokenize{#1}\relax
    \expandafter\@firstoftwo
  \else
    \expandafter\@secondoftwo
  \fi
  {\colorbox{#2}}%
  {\colorbox[#1]{#2}}%
  {#3}\kern-\fboxsep
}
\makeatother

\usepackage{eso-pic}
\definecolor{linkblue}{RGB}{0, 0, 255}

\AddToShipoutPictureBG{
  \AtPageLowerLeft{%
    \raisebox{0.4\baselineskip}{\makebox[\paperwidth]{\begin{minipage}{21cm}\centering
      {\scriptsize Copyright (c) 2021 IEEE. Personal use is permitted. For any other purposes, permission must be obtained from the IEEE by emailing pubs-permissions@ieee.org}
    \end{minipage}}}%
  }
}

\AddToShipoutPictureBG{
  \AtPageUpperLeft{%
    \raisebox{-1.2\baselineskip}{\makebox[\paperwidth]{\begin{minipage}{21cm}\centering
      {\scriptsize This is the author's version of an article that has been published in this journal. Changes were made to this version by the publisher prior to publication.\\
The final version of record is available at \quad {\color{linkblue} \url{http://dx.doi.org/10.1109/TPAMI.2021.3057446}}}
    \end{minipage}}}%
  }
}

\begin{document}

\title{A continual learning survey: \\ Defying forgetting in classification tasks}

\author{Matthias~De Lange, Rahaf Aljundi, Marc Masana,
        Sarah Parisot, Xu Jia,\\ Ale\v{s} Leonardis, Gregory Slabaugh,
        Tinne Tuytelaars

\IEEEcompsocitemizethanks{\IEEEcompsocthanksitem M. De Lange, R. Aljundi and T. Tuytelaars are with the Center for Processing Speech and Images, Department Electrical Engineering, KU Leuven. Correspondence to matthias.delange@kuleuven.be %
\IEEEcompsocthanksitem M. Masana is with Computer Vision Center, UAB. %
\IEEEcompsocthanksitem S. Parisot, X. Jia, A. Leonardis, G. Slabaugh are with Huawei.}%
\thanks{
Manuscript received 6 Sep. 2019; revised 28 Jul. 2020.
\hfil\break
Log number TPAMI-2020-02-0259.}
 }%

\IEEEtitleabstractindextext{%
\begin{abstract}
Artificial neural networks thrive in solving the classification problem for a particular rigid task, acquiring knowledge through generalized learning behaviour from a distinct training phase.
The resulting network resembles a static entity of knowledge, with endeavours to extend this knowledge without targeting the original task resulting in a catastrophic forgetting.
Continual learning shifts this paradigm towards networks that can continually accumulate knowledge over different tasks without the need to retrain from scratch.
We focus on task incremental classification, where tasks arrive sequentially and are delineated by clear boundaries. Our main contributions concern
(1) a taxonomy and extensive overview of the state-of-the-art; (2) a novel framework to continually determine the stability-plasticity trade-off of the continual learner; (3) a comprehensive experimental comparison of \matt{11} state-of-the-art continual learning methods and 4 baselines. 
\matt{
We empirically scrutinize method strengths and weaknesses on three benchmarks, considering Tiny Imagenet and large-scale unbalanced iNaturalist and a sequence of recognition datasets. %
}
We study the influence of model capacity, weight decay and dropout regularization, and the order in which the tasks are presented, and qualitatively compare methods in terms of required memory, computation time and storage.

\end{abstract}

}

\maketitle

\IEEEdisplaynontitleabstractindextext

\IEEEpeerreviewmaketitle

\IEEEraisesectionheading{\section{Introduction}\label{sec:introduction}}

In recent years, machine learning models 
have been reported to exhibit or even surpass %
human level  performance on individual tasks, such as Atari games~\cite{silver2018general} or object recognition%
~\cite{russakovsky2015imagenet}. 
\matt{
While these results are impressive, they are obtained with static models incapable of adapting their behavior over time. As such, this requires restarting the training process each time new data becomes available. In our dynamic world, this practice quickly becomes intractable for data streams or may only be available temporarily due to storage constraints or privacy issues.
This calls for systems that adapt continually and keep on learning over time. 
}

\matt{
Human cognition exemplifies such systems, with a tendency to learn concepts sequentially.
Revisiting old concepts by observing examples may occur, but is not essential to preserve this knowledge, and while humans may gradually forget old information, a complete loss of previous knowledge is rarely attested~\cite{french1999catastrophic}. 
}

\matt{
By contrast, artificial neural networks cannot learn in this manner: they suffer from catastrophic forgetting of old concepts as new ones are learned~\cite{french1999catastrophic}.
To circumvent this problem, research on artificial neural networks has focused mostly on static tasks, with usually shuffled data to ensure i.i.d. conditions, and vast performance increase by revisiting training data over multiple epochs.
}

\boldspacepar{Continual Learning}
\matt{
studies the problem of learning from an infinite stream of data, with the goal of gradually extending acquired knowledge and using it for future learning~\cite{chen2018lifelong}. Data can stem from changing input domains (e.g.~varying imaging conditions) or can be associated with different tasks (e.g.~fine-grained classification problems). Continual learning is also referred to as lifelong learning~\cite{aljundi2017expertgate, chen2018lifelong, chaudhry2018efficient, Parisi2018, silver2002task, Triki2017}, sequential learning~\cite{Aljundi2018_SLNI, mccloskey1989catastrophic, shin2017continual} or incremental learning~\cite{aljundi2018memory, chaudhry2018riemannian, gepperth2016bio, Rebuffi2017, rosenfeld2018incremental, Shmelkov2017, zhang2017survey}. 
The main criterion is the sequential nature of the learning process, with only a small portion of input data from one or few tasks available at once.
The major challenge is to learn without catastrophic forgetting: performance on a previously learned task or domain should not significantly degrade over time as new tasks or domains are added.
This is a direct result of a more general problem in neural networks, namely the stability-plasticity dilemma~\cite{GrossbergStephen1982Soma}, with plasticity referring to the ability of integrating new knowledge, and stability retaining previous knowledge while encoding it. 
}
\reviewfinal{
Albeit a challenging problem, progress in continual learning has led to real-world applications starting to emerge~\cite{lange2020unsupervised,lee2020clinical,gupta2020neural}.
}

\boldspacepar{Scope.}
To keep focus, we limit the scope of our study in two ways. 
First, we only consider the task incremental setting, where data arrives sequentially in batches and one batch corresponds to one task, such as a new set of categories to be learned. 
In other words, we assume for a given task, all data becomes available simultaneously for offline training. This enables learning for multiple epochs over all its training data, repeatedly shuffled to ensure i.i.d. conditions.
Importantly, data from previous or future tasks cannot be accessed. 
Optimizing for a new task in this setting will result in catastrophic forgetting, with significant drops in performance for old tasks, unless special measures are taken. The efficacy of those measures, under different circumstances, is exactly what this paper is about.
\review{
Furthermore, task incremental learning confines the scope to a multi-head configuration, with an exclusive output layer or head for each task.
This is in contrast to the even more challenging class incremental setup with all tasks sharing a single head. This introduces additional interference in learning and increases the number of output nodes to choose from. Instead, we assume it is known which task a given sample belongs to.
}

Second, we focus on classification problems only, as classification is arguably one of the most established tasks for artificial neural networks, with good performance using relatively simple, standard and well understood network architectures.
The setup is described in more detail in Section~\ref{sec:setting}, with Section~\ref{sec:future} discussing open issues towards tackling a more general setting.

\boldspacepar{Motivation.}
\reviewfinal{
There is limited consensus in the literature on experimental setups and datasets to be used. 
Although papers provide evidence for at least one specific setting of model architecture, combination of tasks and hyperparameters under which the proposed method reduces forgetting and outperforms alternative approaches,
there is no comprehensive experimental comparison performed to date. 
}

\boldspacepar{Findings.} 
To establish a fair comparison, the following question needs to be considered:
``How can the trade-off between stability and plasticity be set in a consistent manner, using only data from the current task?" 
For this purpose, we propose a generalizing framework that dynamically determines stability and plasticity for continual learning methods.
Within this principled framework, 
we scrutinize 11 representative approaches in the spectrum of continual learning methods, and find method-specific preferences for configurations regarding model capacity and combinations with typical regularization schemes such as weight decay or dropout.
All representative methods generalize well to the nicely balanced small-scale Tiny Imagenet setup. However, this no longer holds when transferring to large-scale real-world datasets, such as the profoundly unbalanced iNaturalist, and especially in the unevenly distributed RecogSeq sequence with highly dissimilar recognition tasks. Nevertheless, rigorous methods based on isolation of parameters seem to withstand generalizing to these challenging settings.
Additionally, we find the ordering in which tasks are presented to have insignificant impact on performance. 
\reviewfinal{A summary of our main findings can be found in Section~\ref{sec:exp:summary}.}

\boldspacepar{Related work.}  Continual learning has been the subject of several recent surveys~\cite{Parisi2018, lesort2019continual, pfulb2019a, farquhar2018towards}. Parisi et al.~\cite{Parisi2018} describe a wide range of methods and approaches, yet without an empirical evaluation or comparison. 
In the same vein, Lesort et al.~\cite{lesort2019continual} descriptively survey and formalize continual learning, but with an emphasis on dynamic environments for robotics.
Pf\"{u}lb and Gepperth~\cite{pfulb2019a} perform an empirical study  on catastrophic forgetting and develop a protocol for setting hyperparameters and method evaluation. However, they only consider two methods, namely Elastic Weight Consolidation (EWC)~\cite{kirkpatrick2017overcoming} and Incremental Moment Matching (IMM)~\cite{Lee2017}. Also, their evaluation is limited to small datasets. %
\matt{
Similarly, Kemker et al.~\cite{kemker2018measuring} compare only three  methods, EWC~\cite{kirkpatrick2017overcoming},  PathNet~\cite{Fernando2017} and 
GeppNet~\cite{gepperth2016bio}.
The experiments are performed on a simple fully connected network, including only 3 datasets: MNIST, CUB-200 and AudioSet.
}
Farquhar and Gal~\cite{farquhar2018towards} survey continual learning evaluation strategies and highlight shortcomings of the common practice to use one dataset with different pixel permutations
(typically permuted MNIST \cite{goodfellow2013empirical}). 
\matt{
In addition, they stress that task incremental learning under multi-head settings hides the true difficulty of the problem. 
While we agree with this statement, we still opted for a multi-head based survey, as it allows comparing existing methods without major modifications.
}
They also propose a couple of desiderata for evaluating continual learning methods. However, their study is limited to the fairly easy MNIST and Fashion-MNIST datasets with far from realistic data encountered in practical applications.
None of these works systematically addresses the questions raised above.

\boldspacepar{Paper overview. }
First, we describe in Section \ref{sec:setting} the widely adopted task incremental setting for this paper.
Next, Section~\ref{sec:continual_approaches} surveys different approaches towards continual learning, structuring them into three main groups: replay, regularization-based and parameter isolation methods. 
An important issue when aiming for a fair method comparison is the selection of hyperparameters, with in particular the learning rate and the stability-plasticity trade-off. Hence, Section~\ref{sec:framework} introduces a novel framework to tackle this problem without requiring access to data from previous or future tasks.
We provide details on the methods selected for our experimental evaluation in Section~\ref{sec:methods}, and describe the actual experiments, main findings, and qualitative comparison in Section~\ref{sec:experiments}.
We look further ahead in Section~\ref{sec:future}, highlighting additional challenges in the field, moving beyond the task incremental setting towards true continual learning. Then, we emphasize the relation with other fields in Section~\ref{sec:otherfields}, and conclude in Section~\ref{sec:conclusion}. Finally, implementation details and further experimental results are provided in supplemental material.
\review{
Code is publicly available for reproducibility 
\footnote{Code available at: \textsc{https://github.com/mattdl/CLsurvey}}.
}

\section{The Task Incremental Learning Setting}
\label{sec:setting}
\matt{
Due to the general difficulty and variety of challenges in continual learning,
many methods relax the general setting to an easier task incremental one.
The task incremental setting considers a sequence of tasks, receiving training data of just one task at a time to perform training until convergence. 
Data $(\mathcal{X}^{(t)},\mathcal{Y}^{(t)})$ is randomly drawn from distribution $D^{(t)}$, with $\mathcal{X}^{(t)}$ a set of data samples for  task $t$, and $\mathcal{Y}^{(t)}$ the corresponding ground truth labels.
The goal is to control the statistical risk of all seen tasks given limited  or no access to data $(\mathcal{X}^{(t)}, \mathcal{Y}^{(t)})$
from previous tasks $t < \mathcal{T}$:
}

\begin{equation}
\sum_{t=1}^\mathcal{T} \mathbb{E}_{(\mathcal{X}^{(t)},\mathcal{Y}^{(t)})}[\ell(f_t(\mathcal{X}^{(t)};\theta),\mathcal{Y}^{(t)})]
\label{eq:StatRisk}
\end{equation}
with loss function $\ell$, parameters $\theta$, $\mathcal{T}$ the number of tasks seen so far, and  $f_t$ representing the network function for task $t$.
\reviewfinal{The definition can be extended to zero-shot learning for $t > \mathcal{T}$, but mainly remains future work in the field as discussed in Section~\ref{sec:future}.
}
\matt{
For the current task $\mathcal{T}$, the statistical risk can be approximated by the empirical risk
}
\begin{equation}
\frac{1}{N_\mathcal{T}} \sum_{i=1}^{N_\mathcal{T}} \ell(f(x_i^{(\mathcal{T})};\theta),y_i^{(\mathcal{T})}) \ .
\label{eq:EmpRisk}
\end{equation}
\matt{
The main research focus in task incremental learning aims to determine optimal parameters $\theta^{*}$ by optimizing (\ref{eq:StatRisk}).
However, data is no longer available for old tasks, impeding evaluation of statistical risk for the new parameter values. 
}

The concept \emph{'task'} refers to an isolated training phase with a new batch of data, belonging to a new group of classes, a new domain, or a different output space.
\review{
Similar to~\cite{hsu2018re}, a finer subdivision into three categories emerges based on the marginal output and input distributions $P(\mathcal{Y}^{(t)})$ and $P(\mathcal{X}^{(t)})$ of a task $t$, with ${P(\mathcal{X}^{(t)})\neq P(\mathcal{X}^{(t+1)})}$.  
 First, \emph{class incremental learning} defines an output space for all observed class labels ${\{\mathcal{Y}^{(t)}\} \subset \{\mathcal{Y}^{(t+1)}\}}$ with ${P(\mathcal{Y}^{(t)})\ne P(\mathcal{Y}^{(t+1)})}$. 
\emph{Task incremental learning} defines $\{\mathcal{Y}^{(t)}\}\ne \{\mathcal{Y}^{(t+1)}\}$, which additionally requires task label $t$ to indicate the isolated output nodes $\mathcal{Y}^{(t)}$ for current task $t$.
\emph{Incremental domain learning} defines ${\{\mathcal{Y}^{(t)}\} = \{\mathcal{Y}^{(t+1)}\}}$ with ${P(\mathcal{Y}^{(t)})= P(\mathcal{Y}^{(t+1)})}$.
\reviewfinal{Additionally, \emph{data incremental learning}~\cite{lange2020proto} defines a more general continual learning setting for any data stream without notion of task, class or domain.}
Our experiments focus on the task incremental  learning setting for a sequence of classification tasks.
}

\section{Continual Learning Approaches}\label{sec:continual_approaches}
\matt{
Early works observed the catastrophic interference problem when sequentially learning  examples  of different input patterns.}  Several directions have been explored, such as reducing representation overlap~\cite{french1992semi, french1994dynamically, kruschke1992alcove, kruschke1993human, sloman1992reducing}, replaying samples or virtual samples from
the past~\cite{robins1995catastrophic, silver2002task} or introducing dual architectures~\cite{FRENCH1997recurrent, Rueckl1993Jumpnet, ans1997avoiding}. However, due to resource restrictions at the time, these works mainly considered few examples (in the order of tens) and were based on specific shallow architectures.

With the recent increased interest in neural networks, 
continual learning and catastrophic forgetting also received more attention. 
\matt{The influence of dropout~\cite{Srivastava2014Dropout} and different activation functions on forgetting have been studied empirically for two sequential tasks~\cite{goodfellow2013empirical}
, and several works studied task incremental learning from a theoretical perspective 
~\cite{pentina2014pac, alquier2016regret,pentina2015lifelong, balcan2015efficient}.}

More recent works have addressed continual learning with longer task sequences and a larger number of examples. 
In the following, we will review the most important works. We distinguish three families, based on how task specific information is stored and used throughout the sequential learning process: 
\begin{itemize}
    \item Replay methods
    \item Regularization-based methods
    \item Parameter isolation methods
\end{itemize}
Note that our categorization overlaps to some extent with that introduced in previous work~\cite{farquhar2018towards,Aljundi2019continualNoT}. However, we believe it offers a more general overview and covers most existing works. A summary can be found in Figure~\ref{fig:taxonomy}.

\definecolor{level1}{HTML}{f48e82}%
\definecolor{level2}{HTML}{63aff3}%
\definecolor{level3}{HTML}{74df7b}%
\definecolor{level4}{HTML}{ffe080}%
\definecolor{level5}{HTML}{cafcda}

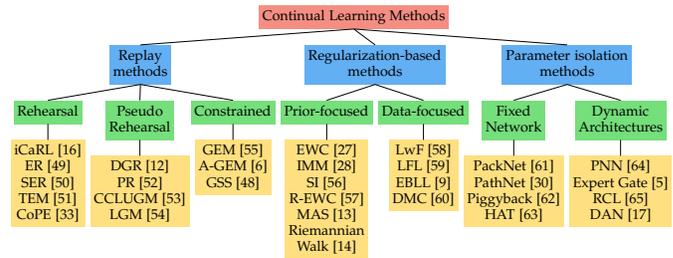
\begin{figure}

\centering
\medskip
\footnotesize

\resizebox{\linewidth}{!}{
 \begin{forest}
for tree={s sep=1mm, inner sep=2, l=1}
[
{Continual Learning Methods },fill=level1
[Replay\\ methods,fill=level2 
[Rehearsal,fill=level3 
[{iCaRL~\cite{Rebuffi2017}}\\ ER~\cite{rolnick2019experience}\\SER~\cite{isele2018selective}\\TEM\cite{chaudhry2019continual}\\CoPE~\cite{lange2020proto},fill=level4]]
[Pseudo\\Rehearsal,fill=level3
[DGR~\cite{shin2017continual}\\PR~\cite{Craig2018Recursal}\\CCLUGM~\cite{lavda2018continual}\\LGM~\cite{ramapuram2017lifelong} ,fill=level4]]
[Constrained,fill=level3 
[GEM~\cite{lopez2017gradient}\\A-GEM~\cite{chaudhry2018efficient}\\GSS~\cite{Aljundi2019continualNoT},fill=level4]
]
]
[{Regularization-based \\ methods},fill=level2 
[{Prior-focused},fill=level3 
[EWC~\cite{kirkpatrick2017overcoming}\\IMM~\cite{Lee2017}\\SI~\cite{zenke2017continual}\\R-EWC~\cite{Liu2018}\\MAS~\cite{aljundi2018memory}\\Riemannian\\Walk~\cite{chaudhry2018riemannian},fill=level4
]
]
[{Data-focused},fill=level3 
[LwF~\cite{li2016learning}\\LFL~\cite{Jung_LFL}\\EBLL~\cite{Triki2017}\\DMC~\cite{Junting2019Class-incremental},fill=level4]
]
]
[{Parameter isolation\\ methods},fill=level2 
[Fixed\\Network ,fill=level3
[PackNet~\cite{Mallya2017}\\PathNet~\cite{Fernando2017}\\Piggyback~\cite{Mallya2018}\\HAT~\cite{Serra2018},fill=level4]]
[Dynamic\\Architectures,fill=level3
[PNN~\cite{rusu2016progressive}\\Expert Gate~\cite{aljundi2017expertgate}\\RCL~\cite{xu2018reinforced}\\DAN~\cite{rosenfeld2018incremental},fill=level4]
]
]
]
\end{forest}
} %

\caption{A tree diagram illustrating the different continual learning families of methods and the different branches within each family. The leaves enlist example methods. }
\label{fig:taxonomy}

\end{figure}

\subsection{Replay methods}
\matt{
This line of work stores samples in raw format or generates pseudo-samples with a generative model. These previous task samples are replayed while learning a new task to alleviate forgetting. They are either reused as model inputs for rehearsal,
or to constrain optimization of the new task loss to prevent previous task interference.
}

{\em Rehearsal methods}~\cite{chaudhry2019continual,rolnick2019experience,isele2018selective,Rebuffi2017} explicitly retrain on a limited subset of stored samples while training on new tasks. The performance of these methods is upper bounded by joint training on previous and current tasks.
\reviewfinal{
Most notable is class incremental learner iCaRL~\cite{Rebuffi2017}, storing a subset of exemplars per class, selected to best approximate class means in the learned feature space. At test time, the class means are calculated for nearest-mean classification based on all exemplars.
In data incremental learning, Rolnick et al. \cite{rolnick2019experience} suggest reservoir sampling to limit the number of stored samples to a fixed budget assuming an i.i.d. data stream. 
Continual Prototype Evolution (CoPE)~\cite{lange2020proto} combines the nearest-mean classifier approach with an efficient reservoir-based sampling scheme.
Additional experiments on rehearsal for class incremental learning are provided in~\cite{masana2020class}.
}

While rehearsal might be prone to overfitting the subset of stored samples and seems to be bounded by joint training,  
\matt{
{\em constrained optimization} is an alternative solution leaving more leeway for backward/forward transfer.} As proposed in GEM~\cite{lopez2017gradient} under the task incremental setting, the key idea is to only constrain new task updates to not interfere with previous tasks. This is achieved through projecting the estimated gradient direction on the feasible region outlined by previous task gradients through first order Taylor series approximation. A-GEM~\cite{chaudhry2018efficient} relaxes the problem to project on one  direction estimated by randomly selected samples from a previous task data buffer. \matt{Aljundi et al. \cite{Aljundi2019continualNoT} extend this solution to a pure online continual learning setting without task boundaries, proposing to select sample subsets that maximally approximate the feasible region of historical data.}

In the absence of previous samples, {\em pseudo rehearsal} is an alternative strategy used in early works with shallow neural networks. \matt{The output of previous model(s) given random inputs are used to approximate previous task samples~\cite{robins1995catastrophic}. With deep networks and large input vectors (e.g. full resolution images), random input cannot cover the input space~\cite{Craig2018Recursal}. Recently,  generative models have shown the ability to generate high quality images~\cite{goodfellow2014generative,DBLP:conf/iclr/2014}, opening up possibilities to model the data generating distribution and retrain on generated examples \cite{shin2017continual}. However, this also adds complexity in training generative models continually, with extra care to balance retrieved examples and avoid mode collapse.}

\subsection{Regularization-based methods}
\matt{
This line of works avoids storing raw inputs, prioritizing privacy,  
and alleviating memory requirements. 
Instead, an extra regularization term is introduced in the loss function, consolidating previous knowledge when learning on new data. We can further divide these methods into data-focused and prior-focused methods.
}

\subsubsection{Data-focused methods}
The basic building block in data-focused methods is knowledge distillation from a previous model (trained on a previous task) to the model being trained on the new data. \matt{Silver et al.~\cite{silver2002task} first proposed to use previous task model outputs given new task input images, mainly for improving new task performance. It has been re-introduced by LwF~\cite{li2016learning} to mitigate forgetting and transfer knowledge, using the previous model output as soft labels for previous tasks. Other works~\cite{Jung_LFL,Junting2019Class-incremental} have been introduced with related ideas, however, it has been shown that this strategy is vulnerable to domain shift between tasks~\cite{aljundi2017expertgate}. In an attempt to overcome this issue, Triki et al.~\cite{Triki2017} facilitate incremental integration of shallow autoencoders to constrain task features in their corresponding learned low dimensional space.}

\subsubsection{Prior-focused methods}
\matt{
To mitigate forgetting, prior-focused methods estimate a distribution over the model parameters, used as prior when learning from new data. 
Typically, importance of all neural network parameters is estimated, with parameters assumed independent to ensure feasibility.
During training of later tasks, changes to important parameters are penalized.
Elastic weight consolidation (EWC) \cite{kirkpatrick2017overcoming} was the first to establish this approach.  Variational Continual Learning (VCL) introduced a variational framework for this family \cite{Nguyen2017}, sprouting a body of Bayesian based works 
\cite{ahn2019uncertainty}, \cite{zeno2018task}.  
Zenke et al.~\cite{zenke2017continual} estimate importance weights online during task training.
Aljundi et al.~\cite{aljundi2018memory} suggest unsupervised importance estimation, allowing increased flexibility and online user adaptation as in \cite{lange2020unsupervised}.
Further work extends this to task free settings~\cite{DBLP:journals/corr/abs-1812-03596}.
}

\subsection{Parameter isolation methods}
\matt{
This family dedicates different model parameters to each task, to prevent any possible forgetting.
When no constraints apply to architecture size, one can grow new branches for new tasks, while freezing previous task parameters~\cite{rusu2016progressive,xu2018reinforced}, or dedicate a model copy to each task~\cite{aljundi2017expertgate}.
Alternatively, the architecture remains static, with fixed parts allocated to each task. Previous task parts are masked out during new task training, either imposed at parameters level \cite{Mallya2017,Fernando2017}, or unit level~\cite{Serra2018}.
These works typically require a task oracle, activating corresponding masks or task branch during prediction.
Therefore, they \matt{are restrained to a multi-head setup, incapable to cope with a shared head between tasks.}
} 
Expert Gate~\cite{aljundi2017expertgate} avoids this problem through learning an auto-encoder gate.

\section{Continual Hyperparameter Framework}
\label{sec:framework}
\matt{
Methods tackling the continual learning problem typically involve extra hyperparameters to balance the stability-plasticity trade-off.  These hyperparameters are in many cases found via a grid search, using held-out validation data from all tasks. However, this inherently violates the main assumption in continual learning, namely no access to  previous task data. 
This may lead to overoptimistic results, that cannot be reproduced in a true continual learning setting.
As it is of our concern in this survey to provide a comprehensive and fair study on the different continual learning methods, we need to establish a principled framework to set hyperparameters without violating the continual learning setting.  Besides a fair comparison over the existing approaches, this general strategy dynamically determines the stability-plasticity trade-off, and therefore extends its value to real-world continual learners.
}

\matt{
First, our proposed framework assumes only access to new task data to comply with the continual learning paradigm. 
Next, the framework aims at the best trade-off in stability and plasticity in a dynamic fashion.
For this purpose, the set of hyperparameters $\mathcal{H}$ related to forgetting should be identified for the continual learning method, as we do
for all compared methods
in Section~\ref{sec:methods}. 
The hyperparameters are initialized to ensure minimal forgetting of previous tasks.
Then, if a predefined threshold on performance is not achieved, hyperparameter values are decayed until reaching the desired performance on the new task validation data.
Algorithm \ref{algor:framework} illustrates the two main phases of our framework, to be repeated for each new task:
}

\matt{
\textbf{Maximal Plasticity Search } first finetunes model copy $\theta^{t'}$ on new task data, given previous model parameters $\theta^{t}$.
The learning rate $\eta^*$ is obtained via  a coarse grid search, which aims for highest accuracy $A^*$ on a held-out validation set from the new task. 
The accuracy $A^*$ represents the best accuracy that can be achieved while disregarding the previous tasks. 
}

\textbf{Stability Decay. } \matt{In the second phase we train $\theta^{t}$ with the acquired learning rate $\eta^*$ using the considered continual learning method. Method specific hyperparameters are set to their highest values to ensure minimum forgetting. 
We define threshold $p$ indicating tolerated drop in new task performance compared to finetuning $A^*$.
When this threshold is not met,  
we decrease hyperparameter values in $\mathcal{H}$ by scalar multiplication with decay factor $\alpha$, and subsequently repeat this phase. This corresponds to increasing model plasticity in order to reach the desired performance threshold.
To avoid redundant stability decay iterations, decayed hyperparameters are propagated to later tasks.}

 \renewcommand{\algorithmicrequire}{\textbf{input }}
  \renewcommand{\algorithmicensure}{\textbf{require }}
  \newcommand{\mattalgorcomment}[1]{\Statex\textit{// #1}}
\algdef{SE}[DOWHILE]{Do}{doWhile}{\algorithmicdo}[1]{\algorithmicwhile\ #1}%

\begin{algorithm}
\begin{algorithmic}[1] %

\Require \matt{$\mathcal{H}$ hyperparameter set, $\alpha \in \left[ 0, 1\right]$ decaying factor, $p \in \left[ 0, 1\right]$ accuracy drop margin, $D^{t+1}$ new task data, \quad $\Psi$ coarse learning rate grid}

\Ensure $\theta^t$ previous task model parameters
\Ensure $CLM$ continual learning method
\mattalgorcomment{Maximal Plasticity Search} 
\State $A^* = 0$
\For{$\eta \in \Psi$} 
\State $A \leftarrow$ Finetune$\left( \: D^{t+1}, \, \eta; \, \theta^t  \:\right)$ \Comment{Finetuning accuracy}
\If{$A > A^*$}
\State $A^*,\, \eta^* \leftarrow A,\, \eta$ \Comment{Update best values}
\EndIf
\EndFor
\mattalgorcomment{Stability Decay}
\Do 
\State $A \leftarrow$ $CLM$$\left( \: D^{t+1}, \, \eta^*; \, \theta^t  \:\right)$
\If{$A < (1-p)A^*$}
\State $\mathcal{H} \leftarrow \alpha \cdot \mathcal{H}$ \Comment{Hyperparameter decay}
\EndIf
\doWhile{$A < (1-p)A^*$}

\end{algorithmic}
\caption{Continual Hyperparameter Selection Framework}
\label{algor:framework}
\end{algorithm}

\section{Compared Methods}
\label{sec:methods}%

In Section~\ref{sec:experiments}, we carry out a comprehensive comparison between representative methods from each of the three families of continual learning approaches introduced in Section~\ref{sec:continual_approaches}. For clarity, we first provide a brief description of the selected methods, highlighting their main characteristics.

\subsection{Replay methods}
\textbf{iCaRL}~\cite{Rebuffi2017} was the first replay method, focused on learning in a class-incremental way.
Assuming fixed allocated memory, it selects and stores samples (exemplars) closest to the feature mean of each class.
During training, the estimated loss on new classes is minimized, along with the distillation loss between targets obtained from previous model predictions and current model predictions on the previously learned classes. As the distillation loss strength correlates with preservation of the previous knowledge, this hyperparameter is optimized in our proposed framework.
In our study, we consider the task incremental setting, and therefore implement iCarl also in a multi-head fashion to perform a fair comparison with other methods.

\textbf{GEM}~\cite{lopez2017gradient} exploits exemplars to solve a constrained optimization problem, projecting the current task gradient in a feasible area, outlined by the previous task gradients. 
The authors observe increased backward transfer by altering the gradient projection with a small constant ${\gamma \ge 0}$, constituting $\mathcal{H}$ in our framework.

The major drawback of replay methods is limited scalability over the number of classes, requiring additional computation and storage of raw input samples.
Although fixing memory limits memory consumption, this also deteriorates the ability of exemplar sets to represent the original distribution.
Additionally, storing these raw input samples may also lead to privacy issues.

\subsection{Regularization-based methods}
This survey strongly focuses on regularization-based methods, comparing seven methods in this family. The regularization strength correlates to the amount of knowledge retention, and therefore constitutes $\mathcal{H}$ in our hyperparameter framework.

\textbf{Learning without Forgetting} (LwF)~\cite{li2016learning} retains knowledge of preceding tasks by means of knowledge distillation~\cite{Hinton2015}. Before training the new task, network outputs for the new task data are recorded, and are subsequently used during training to distill prior task knowledge.
However, the success of this method depends heavily on the new task data and how strong it is related to prior tasks. Distribution shifts with respect to the previously learned tasks can result in a gradual error build-up to the prior tasks as more dissimilar tasks are added~\cite{li2016learning, aljundi2017expertgate}. This error build-up also applies in a class-incremental setup, as shown in~\cite{Rebuffi2017}.
\matt{Another drawback resides in the additional overhead to forward all new task data, and storing the outputs.}
LwF is specifically designed for classification, but has also been applied to other problems, such as object detection~\cite{Shmelkov2017}.

\matt{\textbf{Encoder Based Lifelong Learning} (EBLL)~\cite{Triki2017} extends LwF by preserving important low dimensional feature representations of previous tasks.
For each task, an under-complete autoencoder is optimized end-to-end, projecting features on a lower dimensional manifold.
}
During training, an additional regularization term impedes the current feature projections to deviate from previous task optimal ones. 
Although required memory grows linearly with the number of tasks, autoencoder size constitutes only a small fraction of the backbone network. The main computational overhead occurs in autoencoder training, and collecting feature projections for the samples in each optimization step.

\matt{\textbf{Elastic Weight Consolidation} (EWC)~\cite{kirkpatrick2017overcoming} 
introduces network parameter uncertainty in the Bayesian framework~\cite{MacKay1992}.
Following sequential Bayesian estimation, the old posterior of previous tasks  $t < \mathcal{T}$ constitutes the prior for new task $\mathcal{T}$, founding a mechanism to propagate old task importance weights.
The true posterior is intractable, and is therefore estimated using a Laplace approximation with precision determined by the Fisher Information Matrix (FIM).
Near a minimum, this FIM shows equivalence to the positive semi-definite second order derivative of the loss~\cite{pascanu2013revisiting}, and is in practice typically approximated by the empirical FIM to avoid additional backward passes~\cite{Martens2014}:
\begin{equation} \label{eq_EWC_IW}
    \Omega^{\mathcal{T}}_k = \mathbb{E}_{(x,y)\sim D^{\mathcal{T}}} \left [ \left ( \frac{\delta \mathcal{L}}{\delta \theta^{\mathcal{T}}_k} \right )^2 \right ] \,,
\end{equation}
with importance weight $\Omega^{\mathcal{T}}_k$ calculated after training task $\mathcal{T}$. %
Although originally requiring one FIM per task, this can be resolved by propagating a single penalty \cite{huszar2018note}.
Further, the FIM is approximated after optimizing the task, inducing gradients close to zero, and hence very little regularization. This is inherently coped with in our framework, as the regularization strength is initially very high and lowered only to decay stability.
Variants of EWC are proposed to address these issues in~\cite{Schwarz2018},~\cite{Liu2018} and~\cite{chaudhry2018riemannian}. 
}

\textbf{Synaptic Intelligence} (SI)~\cite{zenke2017continual} breaks the EWC paradigm of determining the new task importance weights $\Omega^{\mathcal{T}}_k$ in a separate phase after training. Instead, they maintain an online estimate $\omega^{\mathcal{T}}$ during training to eventually attain
\begin{equation}
    \Omega^{\mathcal{T}}_k = \sum_{t=1}^{\mathcal{T}} \frac{\omega^t_k}{(\Delta \theta^t_k)^2 + \xi} \, ,
\end{equation}
with $\Delta \theta^t_k = \theta^t_k - \theta^{t-1}_k$ the task-specific parameter distance, and damping parameter $\xi$ avoiding division by zero. 
Note that the accumulated importance weights $\Omega^{\mathcal{T}}_k$ are still only updated after training task $\mathcal{T}$ as in EWC. 
Further, the knife cuts both ways for efficient online calculation. First, stochastic gradient descent incurs noise in the approximated gradient during training, and therefore the authors state importance weights tend to be overestimated. Second, catastrophic forgetting in a pretrained network becomes inevitable, as importance weights can't be retrieved.
In another work, Riemannian Walk~\cite{chaudhry2018riemannian} combines the SI path integral with an online version of EWC to measure parameter importance.

\textbf{Memory Aware Synapses} (MAS)~\cite{aljundi2018memory} redefines the parameter importance measure to an unsupervised setting. Instead of calculating gradients of the loss function $\mathcal{L}$ as in~(\ref{eq_EWC_IW}), the authors obtain gradients of the squared $L_2$-norm of the learned network output function $f_{\mathcal{T}}$:
\begin{equation} \label{eq_MAS_IW}
    \Omega^{\mathcal{T}}_k = \mathbb{E}_{x\sim D^{\mathcal{T}}} [ \frac{\delta \left\lVert f_{\mathcal{T}}(x;\theta)\right\rVert_2^2}{\delta \theta^{\mathcal{T}}_k}]\,.
\end{equation}
Previously discussed methods require supervised data for the loss-based importance weight estimations, and are therefore confined to available training data. By contrast, MAS enables importance weight estimation on an unsupervised held-out dataset, hence capable of user-specific data adaptation.

\textbf{Incremental Moment Matching} (IMM)~\cite{Lee2017} estimates Gaussian posteriors for task parameters, in the same vein as EWC, but inherently differs in its use of model merging.
In the merging step, the mixture of Gaussian posteriors is approximated by a single Gaussian distribution, i.e. a new set of merged parameters $\theta^{1:\mathcal{T}}$ and corresponding covariances $\Sigma^{1:\mathcal{T}}$.
Although the merging strategy implies a single merged model for deployment, it requires storing models during training for each learned task.
In their work, two methods for the merge step are proposed: \emph{mean-IMM} and \emph{mode-IMM}. In the former, weights $\theta_k$ of task-specific networks are averaged following the weighted sum:
\begin{equation}
    \theta^{1:\mathcal{T}}_k = \sum\nolimits^{\mathcal{T}}_{t}\alpha_k^t \theta_k^t\,,
\end{equation}
with $\alpha_k^t$ the mixing ratio of task $t$, subject to $\sum\nolimits_t^{\mathcal{T}}\alpha_k^t = 1$. 
Alternatively, the second merging method \emph{mode-IMM} aims for the mode of the Gaussian mixture,
with importance weights obtained from (\ref{eq_EWC_IW}):
\begin{equation}
    \theta^{1:\mathcal{T}}_k =  \frac{1}{\Omega^{1:\mathcal{T}}_k} \sum\nolimits_{t}^{\mathcal{T}}\alpha_k^t \Omega^t_k  \theta_k^t \,, \quad
    \Omega^{1:\mathcal{T}}_k = \sum\nolimits_{t}^{\mathcal{T}}\alpha_k^t \Omega^{t}_k \,.
\end{equation}
When two models converge to a different local minimum due to independent initialization, simply averaging the models might result in increased loss, as there are no guarantees for a flat or convex loss surface between the two points in parameter space~\cite{Goodfellow2014}. Therefore, IMM suggests three transfer techniques aiming for an optimal solution in the interpolation of the task-specific models:
i) \emph{Weight-Transfer} initializes the new task network with previous task parameters; ii) \emph{Drop-Transfer} is a variant of dropout~\cite{Hinton2012} with  previous task parameters as zero point; iii) \emph{L2-transfer} is a variant of L2-regularization, again with previous task parameters redefining the zero point.
In this study, we compare \emph{mean-IMM} and \emph{mode-IMM} with both weight-transfer and L2-transfer.
\matt{
As \emph{mean-IMM} is consistently outperformed by \emph{mode-IMM}, we refer for its full results to appendix.
}

\subsection{Parameter isolation methods}
\review{
This family of methods isolates parameters for specific tasks and can guarantee maximal stability by fixing the parameter subsets of previous tasks. Although this refrains stability decay, the capacity used for the new task can be minimized instead to avoid capacity saturation and ensuring stable learning for future tasks.
}

\textbf{PackNet}~\cite{Mallya2017} iteratively assigns parameter subsets to consecutive tasks by constituting binary masks. 
For this purpose, new tasks establish two training phases. First, the network is trained without altering previous task parameter subsets.
Subsequently, a portion of unimportant free parameters are pruned, measured by lowest magnitude. 
Then, the second training round retrains this remaining subset of important parameters.
The pruning mask preserves task performance as it ensures to fix the task parameter subset for future tasks.
PackNet allows explicit allocation of network capacity per task, and therefore inherently limits the total number of tasks.
\review{
$\mathcal{H}$ contains the per-layer pruning fraction.
}

\matt{
\textbf{HAT}~\cite{Serra2018} requires only one training phase, incorporating task-specific embeddings for attention masking. 
The per-layer embeddings are gated through a Sigmoid to attain unit-based attention masks in the forward pass. The Sigmoid slope is rectified through training of each epoch, initially allowing mask modifications and culminating into near-binary masks.
To facilitate capacity for further tasks, a regularization term imposes sparsity on the new task attention mask.
Core to this method is constraining parameter updates between two units deemed important for previous tasks, based on the attention masks. Both regularization strength and Sigmoid slope are considered in  $\mathcal{H}$. We only report results on Tiny Imagenet as difficulties regarding asymmetric capacity and hyperparameter sensitivity  prevent using HAT for more difficult setups \cite{masana2020ternary}. 
}
\review{
We refer to Appendix~B.5 and B.6 for a detailed analysis on capacity usage of parameter isolation methods.
}

\section{Experiments}
\label{sec:experiments}
\tikzset{   
        every picture/.style={remember picture,baseline},
        every node/.style={anchor=base,align=center,outer sep=1.5pt},
        every path/.style={thick},
        }
In this section we first discuss the experimental setup in Section~\ref{sec:exp:setup}, followed by a comparison of all the methods on a common \textsc{base} model in Section~\ref{sec:exp_base}. The effects of changing the capacity of this model are discussed in Section~\ref{sec:exp:capacity}.  Next, in Section~\ref{sec:exp:regul} we look at the effect of two popular methods for regularization. 
We continue in Section~\ref{sec:exp:largedataset}, scrutinizing the behaviour of continual learning methods in a real-world setup, abandoning the artificially imposed balance between tasks.
In addition, we investigate the effect of the task ordering in both the balanced and unbalanced setup in Section~\ref{sec:exp:taskorder}, and elucidate a qualitative comparison in Table~\ref{tab:qualitative}  of Section~\ref{sec:exp:qual}.
Finally, Section~\ref{sec:exp:summary} summarizes our main findings in Table~\ref{tab:summary}.

\subsection{Experimental Setup}
\label{sec:exp:setup}
\par\noindent
\textbf{Datasets.} We conduct image classification experiments on \matt{three} datasets, the main characteristics of which are summarized in Table~\ref{tab:exp:setup:ds}.
First, we use the \textbf{Tiny Imagenet} dataset~\cite{tinyimgnet}. This is a subset of 200 classes from ImageNet~\cite{deng2009imagenet}, rescaled to image size $64\times64$. 
Each class contains 500 samples subdivided into training ($80\%$) and validation ($20\%$), and 50 samples for evaluation. %
In order to construct a balanced dataset, we assign an equal amount of 20 randomly chosen classes to each task in a sequence of 10 consecutive tasks. %
\matt{
This task incremental setting allows using an oracle at test time for our evaluation per task, ensuring all tasks are roughly similar in terms of difficulty, size, and distribution, making the interpretation of the results easier.
}

The second dataset is based on \textbf{iNaturalist}~\cite{inaturalist}, which aims for a more real-world setting with a large number of fine-grained categories and highly imbalanced classes. On top, we impose task imbalance and domain shifts between tasks by assigning $10$ super-categories of species as separate tasks. We selected the most balanced $10$ super-categories from the total of $14$ and only retained categories with at least $100$ samples.
More details on the statistics for each of these tasks can be found in Table~\ref{tab:ds:inat}.
We only utilize the training data, subdivided in training ($70\%$), validation ($20\%$) and evaluation ($10\%$) sets, with all images measuring $800\times600$.

\matt{
Thirdly, we adopt a sequence of 8 highly diverse recognition tasks (\textbf{RecogSeq}) as in \cite{aljundi2017expertgate} and \cite{aljundi2018memory}, which also targets distributing an imbalanced number of classes, and reaches beyond object recognition with scene and action recognition.
This sequence is composed of 8 consecutive datasets, from fine-grained to coarse classification of objects, actions and scenes, going from flowers, scenes, birds and cars, to aircrafts, actions, letters and digits. Details are provided in Table~\ref{tab:recogseq}, with validation and evaluation split following \cite{aljundi2017expertgate}.
}

In this survey we scrutinize the effects of different task orderings for both Tiny Imagenet and iNaturalist in Section~\ref{sec:exp:taskorder}.
Apart from that section, discussed results \matt{on both datasets} are performed on a random ordering of tasks, \matt{and the fixed dataset ordering for RecogSeq.}

\begin{table}[]
\centering
\caption{\matt{The balanced Tiny Imagenet, and unbalanced iNaturalist and RecogSeq dataset characteristics.}}
\label{tab:exp:setup:ds}

\begin{tabular}{@{}llll@{}}
\toprule
                         & \textbf{Tiny Imagenet} & \textbf{iNaturalist} & \textbf{RecogSeq}\\ \midrule
Tasks                    & 10                     & 10    & 8               \\
Classes/task         & 20                     & 5 to 314        & 10 to 200      \\
Train data/task   & 8k                     & 0.6k to 66k       & 2k to 73k   \\
Val. data/task & 1k                     & 0.1k to 9k       & 0.6k to 13k    \\
Task selection        & random class & supercategory     & dataset   \\ \bottomrule
\end{tabular}%
\end{table}

\begin{table}[]
\centering
\caption{The RecogSeq dataset sequence details. Note that \emph{VOC Actions} represents the human action classification subset of the VOC challenge 2012~\cite{everingham2015pascal}.}
\label{tab:recogseq}
\begin{tabular}{@{}lllll@{}}
\toprule
\textbf{Task} & \textbf{Classes}         & \multicolumn{3}{c}{\textbf{Samples}}                                        \\ \midrule
\textit{}     &   \multicolumn{1}{|l|}{}  & \textit{Train} & \textit{Val} & \textit{Test} \\
\textbf{Oxford Flowers}~\cite{nilsback2008automated}       & \multicolumn{1}{|l|}{102} & 2040           & 3074         & 3075          \\
\textbf{MIT Scenes}~\cite{quattoni2009recognizing}        & \multicolumn{1}{|l|}{67}  & 5360           & 670          & 670           \\
\textbf{Caltech-UCSD Birds}~\cite{welinder2010caltech}         & \multicolumn{1}{|l|}{200} & 5994           & 2897         & 2897          \\
\textbf{Stanford Cars}~\cite{krause20133d}          & \multicolumn{1}{|l|}{196} & 8144           & 4020         & 4021    \\
 \textbf{FGVC-Aircraft}~\cite{maji2013fine}      & \multicolumn{1}{|l|}{100} & 6666           & 1666         & 1667          \\
\textbf{VOC Actions}~\cite{everingham2015pascal}       & \multicolumn{1}{|l|}{11}  & 3102           & 1554         & 1554          \\
\textbf{Letters}~\cite{de2009character}       & \multicolumn{1}{|l|}{52}  & 6850           & 580          & 570           \\
\textbf{SVHN}~\cite{netzer2011reading}          & \multicolumn{1}{|l|}{10}                       & 73257                               & 13016        & 13016         \\ \bottomrule
\end{tabular}%
\end{table}

\par\smallskip\noindent
\textbf{Models.} %
We summarize in Table~\ref{tab:exp:setup:models} the models used for experiments in this work.
Due to the limited size of \textbf{Tiny Imagenet} we can easily run experiments with different models. This allows to analyze the influence of model capacity (Section~\ref{sec:exp:capacity}) and regularization for each model configuration (Section~\ref{sec:exp:regul}).
This is important, as the effect of model size and architecture on the performance of different continual learning methods has not received much attention so far.
We configure a \textsc{base} model, two models with less (\textsc{small}) and more (\textsc{wide}) units per layer, and a \textsc{deep} model with more layers.
The models are based on a VGG configuration~\cite{Chung2018}, but with less parameters due to the small image size.
We reduce the feature extractor to comprise 4 max-pooling layers, each preceded by a stack of identical convolutional layers with a consistent $3\, \times\, 3$ receptive field. The first max-pooling layer is preceded by one conv. layer with 64 filters. Depending on the model, we increase subsequent conv. layer stacks with a multiple of factor 2.
The models have a classifier consisting of 2 fully connected layers with each 128 units for the \textsc{small} model and 512 units for the other three models.
\matt{
The multi-head setting imposes a separate fully connected layer with softmax for each task, with the number of outputs defined by the classes in that task.
}
The detailed description can be found in Appendix~\ref{apdx:implementation-details}.

The size of \textbf{iNaturalist} and \textbf{RecogSeq} impose arduous learning. Therefore, we conduct the experiments solely for AlexNet~\cite{Krizhevsky}, pretrained on ImageNet.

\begin{table*}[]
\centering
\caption{\matt{Models used for Tiny Imagenet, iNaturalist and RecogSeq experiments. }
}
\label{tab:exp:setup:models}
\resizebox{\textwidth}{!}{%
\begin{tabular}{@{}llllllllll@{}}
\toprule
                       & \textbf{Model} & \multicolumn{3}{c}{\textbf{Feature Extractor}}                 & \multicolumn{2}{c}{\textbf{Classifier (w/o head)}} & \textbf{Total Parameters} & \textbf{Pretrained}  & \textbf{Multi-head} \\ \midrule
\textit{\textbf{}}     & \textit{}      & \textit{Conv. Layers} & \textit{MaxPool} & \textit{Parameters} & \textit{FC layers}          & \textit{Parameters}          & \textit{}                 & \textit{}            & \textit{}           \\ \addlinespace
\textbf{Tiny Imagenet} & \textsc{small} & 6                     & 4                & 334k                & 2                           & 279k                         & 613k                      & \xmark               & \checkmark          \\
\textbf{}              & \textsc{base}  & 6                     & 4                & 1.15m               & 2                           & 2.36k                        & 3.51m                     & \xmark               & \checkmark          \\
\textbf{}              & \textsc{wide}  & 6                     & 4                & 4.5m                & 2                           & 4.46m                        & 8.95m                     & \xmark               & \checkmark          \\
\textbf{}              & \textsc{deep}  & 20                    & 4                & 4.28m               & 2                           & 2.36k                        & 6.64m                     & \xmark               & \checkmark          \\ \addlinespace
\textbf{iNaturalist/ RecogSeq}   & AlexNet        & 5                     & 3                & 2.47m               & 2 (with Dropout)            & 54.5m                        & 57.0m                     & \checkmark (Imagenet) & \checkmark          \\ \bottomrule
\end{tabular}%
}
\end{table*}

\par\smallskip\noindent
\textbf{Evaluation Metrics.}
To measure performance in the continual learning setup, we evaluate accuracy and forgetting per task, after training each task.
We define this measure of forgetting~\cite{chaudhry2018riemannian} as the difference between the expectation of acquired knowledge of a task, i.e. the accuracy when first learning a task, and the accuracy obtained after training one or more additional tasks.
In the figures, we focus on evolution of accuracy for each task as more tasks are added.
In the tables, we report average accuracy and average forgetting on the final model, obtained by evaluating each task after learning the entire task sequence.

\par\smallskip\noindent
\textbf{Baselines.} %
The discussed continual learning methods in Section~\ref{sec:methods} are compared against several baselines:
    
\begin{enumerate}[]
    \item \label{exp:setup:FT} \matt{\textbf{Finetuning} starts form the previous task model to optimize current task parameters. This baseline greedily trains each task without considering previous task performance, hence introducing catastrophic forgetting, and representing the minimum desired performance.}
    
    \item \textbf{Joint} training considers all data in the task sequence simultaneously, hence violating the continual learning setup (indicated with appended '$*$' in reported results). \matt{This baseline provides a target reference performance.}
\end{enumerate}

\noindent
For the replay methods we consider two additional finetuning baselines, extending baseline (\ref{exp:setup:FT}) with the benefit of using exemplars:
\begin{enumerate}[]
    \setcounter{enumi}{2}
    \item \textbf{Basic rehearsal with Full Memory (R-FM)} fully exploits total available exemplar memory $R$, incrementally dividing equal capacity over all previous tasks. This is a baseline for replay methods defining memory management policies to exploit all memory (e.g. iCaRL).
    \item \textbf{Basic rehearsal with Partial Memory (R-PM)} pre-allocates fixed exemplar memory $R / \mathcal{T}$ over all tasks, assuming the amount of tasks $\mathcal{T}$ is known beforehand. This is used by methods lacking memory management policies (e.g.~GEM).
\end{enumerate}

\par\noindent
\textbf{Replay Buffers.}
Replay methods (GEM, iCaRL) and corresponding baselines (R-PM, R-FM) can be configured with arbitrary replay buffer size. Too large buffers would result in unfair comparison to regularization and parameter isolation methods, with in its limit even holding all data for the previous tasks, as in joint learning, which is not compliant with the continual learning setup.
Therefore, we use the memory required to store the \textsc{base} model as a basic reference for the amount of exemplars. This corresponds to the additional memory needed to propagate importance weights in the prior-focused methods (EWC, SI, MAS). For Tiny Imagenet, this gives a total replay buffer capacity of $4.5$k exemplars. We also experiment with a buffer of $9$k exemplars to examine the influence of increased buffer capacity. 
Note that the \textsc{base} reference implies replay methods to use a significantly different amount of memory than the non-replay based methods; more memory for the {\sc small} model), and less memory for the {\sc wide} and {\sc deep} models.
Comparisons between replay and non-replay based methods should thus only be done for the {\sc base} model. 
In the following,  notation of the methods without replay buffer size refers to the default $4.5$k buffer.

\par\noindent
\textbf{Learning details.} %
During training the models are optimized with stochastic gradient descent with a momentum of $0.9$. Training lasts for $70$ epochs unless preempted by early stopping.
\matt{
Standard AlexNet is configured with dropout for iNaturalist and RecogSeq.
For Tiny Imagenet we do not use any form of regularization by default, except for Section~\ref{sec:exp:regul} explicitly scrutinizing regularization influence. 
}
The framework we proposed in Section~\ref{sec:framework} can only be exploited in the case of forgetting-related hyperparameters, in which case we set $p=0.2$ and $\alpha = 0.5$. All methods discussed in Section~\ref{sec:methods} satisfy this requirement, except for IMM with L2 transfer for which we could not identify a specific hyperparameter related to forgetting.
For specific implementation details about the continual learning methods we refer to Appendix~A.

\subsection{Comparing Methods on the {\sc Base} Network}
\label{sec:exp_base}
\boldspacepar{Tiny Imagenet.} We start evaluation of continual learning methods discussed in Section~\ref{sec:methods}
with a comparison using the \textsc{base} network, on the Tiny Imagenet dataset,
with random ordering of tasks. This provides a balanced setup, making interpretation of results easier.

\begin{figure*}[!hbtp]
    \center{\includegraphics[width=1.\textwidth]{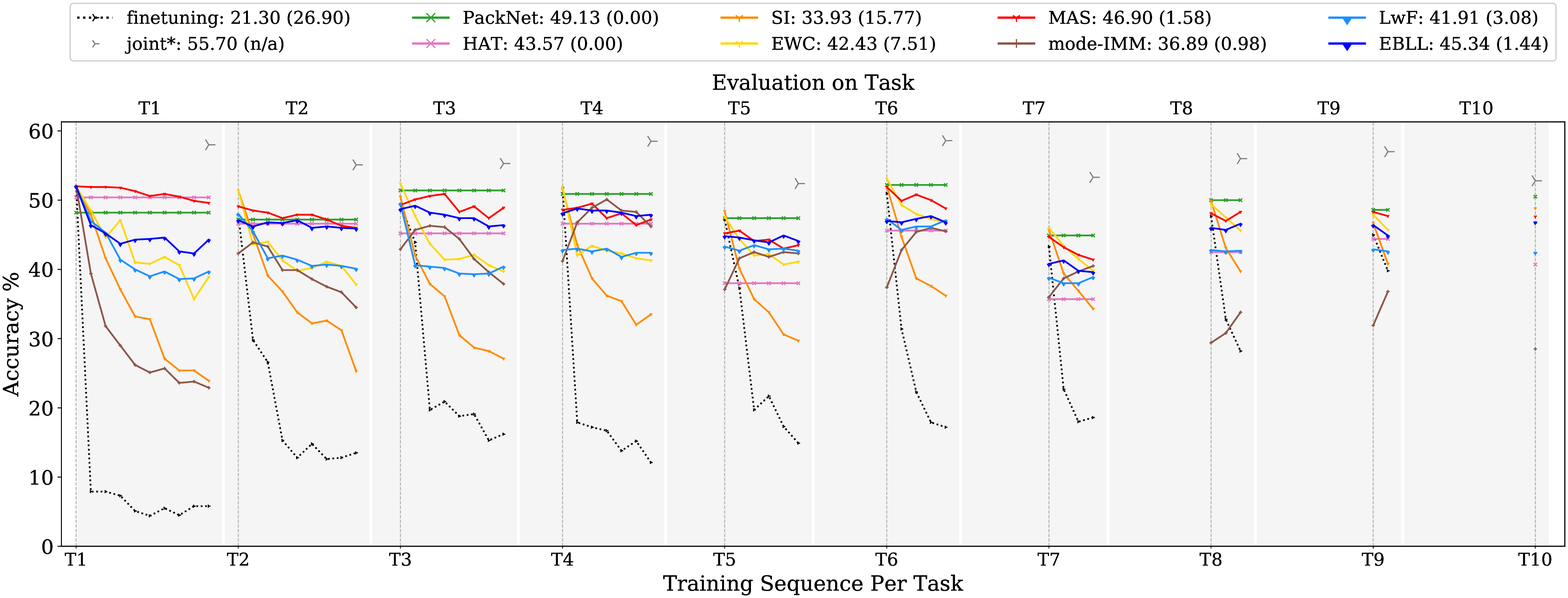}}
    \center{\includegraphics[width=1.\textwidth]{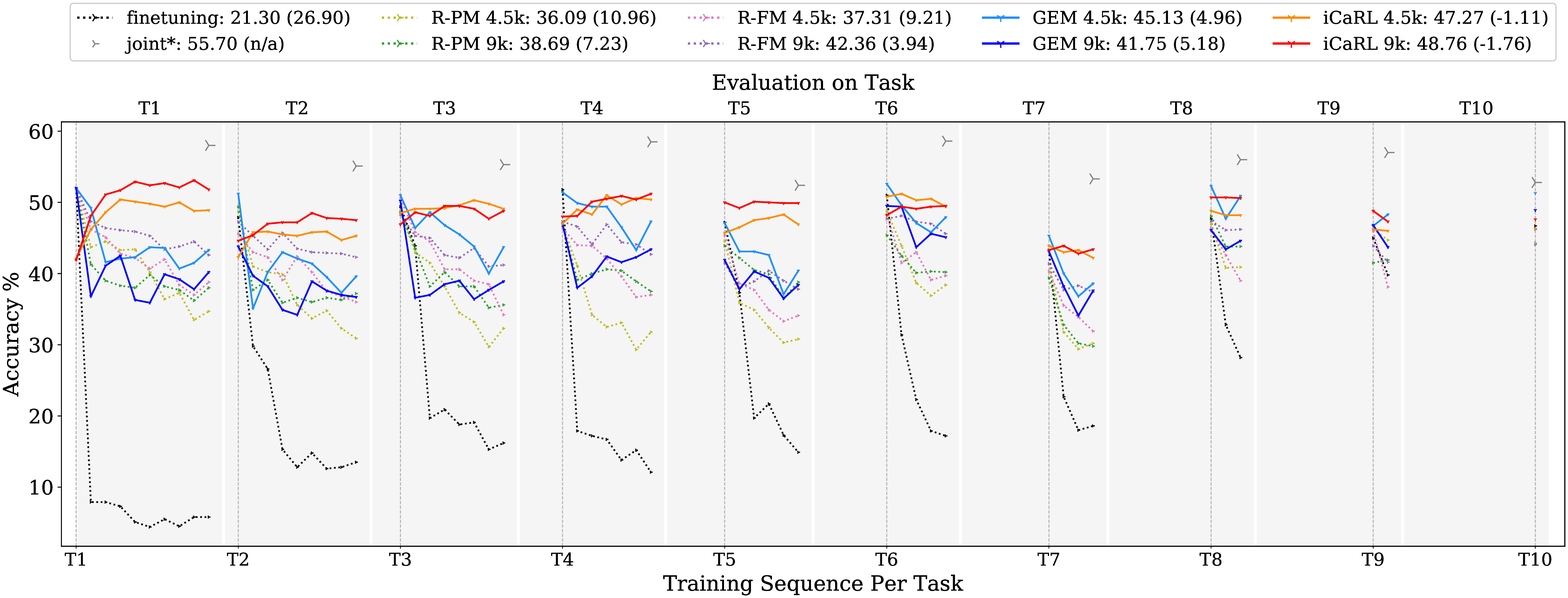}}
          \caption{\label{fig:tinyimgnet:small-model} \matt{Parameter isolation and regularization-based methods (top) and replay methods (bottom) on Tiny Imagenet for the \textsc{base} model with random ordering,
    reporting average accuracy (forgetting) in the legend.}}
  \end{figure*}

\matt{
Figure~\ref{fig:tinyimgnet:small-model} shows results for all three families.
Each of these figures consists of 10 subpanels, with each subpanel showing test accuracy evolution for a specific task (e.g.~Task~1 for the leftmost panel) as more tasks are added for training. 
Since the $n$-th task is added for training only after $n$ steps, curves get shorter as we move to subpanels on the right. 
}

\subsubsection{Discussion}
\emphspacepar{General observations.}
As a reference, it is relevant to highlight the soft upper bound obtained when training all tasks jointly.
This is indicated by the star symbol in each of the subpanels.
For the Tiny Imagenet dataset, all tasks have the same number of classes, same amount of training data and similar level of difficulty, resulting in similar accuracies for all tasks under the joint training scheme. The average accuracy for joint training on this dataset is $55.70\%$, 
while random guessing would result in $5\%$.

Further, as reported ample times in literature, the finetuning baseline suffers severely from catastrophic forgetting: initially good results are obtained when learning a task, but as soon as a new task is added, performance drops, resulting in poor average accuracy of only $21.30\%$ and $26.90\%$ average forgetting. 

With \matt{$49.13\%$} average accuracy, PackNet shows highest overall performance after training all tasks. 
When learning new tasks, due to compression PackNet only has a fraction of the total model capacity available. Therefore, it typically performs worse on new tasks compared to other methods. 
However, by fixing task parameters through masking, it allows complete knowledge retention until the end of the task sequence (no forgetting yields flat curves in the figure), resulting in highest accumulation of knowledge. This holds at least when working with long sequences, where forgetting errors gradually build up for other methods. 

MAS and iCaRL show competitive results w.r.t. PackNet (resp. $46.90\%$ and $47.27\%$).
iCaRL starts with a significantly lower accuracy for new tasks, due to its nearest-neighbour based classification, but improves over time, especially for the first few tasks.
\matt{Further, doubling replay-buffer size to $9$k enhances iCaRL performance to $48.76\%$.}

\emphspacepar{Regularization-based methods.}
 In prior experiments where we did a grid search over a range of hyperparameters over the whole sequence (rather than using the framework introduced in Section~\ref{sec:framework}), we observed MAS to be remarkably more robust to the choice of hyperparameter values compared to the two related methods EWC and SI. 
Switching to the continual hyperparameter selection framework described in Section~\ref{sec:framework},
this robustness leads to superior results for MAS with $46.90\%$ compared to the other two ($42.43\%$ and $33.93\%$). Especially improved forgetting stands out ($1.58\%$ vs $7.51\%$ and $15.77\%$).
Further, SI underperforms compared to EWC on Tiny Imagenet. We hypothesize this may be due to overfitting we observed when training on the \textsc{base} model (see results Appendix~B.2).
SI inherently constrains parameter importance estimation to the training data only, which is opposed to EWC and MAS able to determine parameter importance both on validation and training data in a separate phase after task training. 
\matt{
Further, mode-IMM catastrophically forgets in evaluation on the first task, but shows transitory recovering through backward transfer in all subsequent tasks.
Overall for Tiny Imagenet, IMM does not seem competitive with the other strategies for continual learning - especially if one takes into account that they need to store all previous task models, making them much more expensive in terms of storage.
}

The data-driven methods LwF and EBLL obtain similar results as EWC ($41.91\%$ and $45.34\%$ vs $42.43\%$). 
EBLL improves over LwF by residing closer to the optimal task representations, lowering average forgetting and improving accuracy.
Apart from the first few tasks, curves  for these methods are quite flat, indicating low levels of forgetting.

\emphspacepar{Replay methods.}
iCaRL starts from the same model as GEM and the regularization-based methods, but uses a feature classifier based on a nearest-neighbor scheme. As indicated earlier, this results in lower accuracies on the first task after training. Remarkably, for about half of the tasks, the iCaRL accuracy increases when learning additional tasks, resulting in a salient negative average forgetting of $-1.11\%$. Such level of backward transfer is unusual. After training all ten tasks, a competitive result of $47.27\%$ average accuracy can be reported. 
Comparing iCaRL to its baseline R-FM shows significant improvements over basic rehearsal ($47.27\%$ vs $37.31\%$).
Doubling the size of the replay buffer (iCaRL $9$k) increases performance even more, \matt{pushing iCarl closer to PackNet with $48.76\%$ average accuracy.}

The results of GEM are significantly lower than those of iCaRL ($45.13\%$ vs $47.27\%$). GEM is originally designed for an online learning setup, while in this comparison each method can exploit multiple epochs for a task. Additional experiments in Appendix~B.3 compare GEM sensitivity to the amount of epochs with iCaRL, from which we procure a GEM setup with 5 epochs for all the experiments in this work.
Furthermore, the lack of memory management policy in GEM gives iCaRL a compelling advantage w.r.t. the amount of exemplars for the first tasks, e.g. training Task 2 comprises a replay buffer of Task 1 with factor $10$ (number of tasks) more exemplars.
Surprisingly, GEM $9$k with twice as much exemplar capacity doesn't perform better than GEM $4.5$k. This unexpected behavior may be due to the random exemplar selection yielding a less representative subset.
Nevertheless, GEM convincingly improves accuracy of the basic replay baseline R-PM with the same partial memory scheme ($45.13\%$ vs $36.09\%$).

Comparing the two basic rehearsal baselines that use the memory in different ways, we observe the scheme exploiting full memory from the start (R-FM) giving significantly better results than R-PM for tasks 1 to 7, but not for the last three. 
As more tasks are seen, both baselines converge to the same memory scheme, where for the final task R-FM allocates an equal portion of memory to each task in the sequence, hence equivalent to R-PM.

\matt{
\emphspacepar{Parameter isolation methods.} Both HAT and PackNet exhibit the virtue of masking, reporting zero forgetting (see the two flat curves). HAT performs better for the first task, but builds up a salient discrepancy with PackNet for later tasks. PackNet rigorously assigns layer-wise portions of network capacity per task, while HAT imposes sparsity by regularization, which does not guarantee uniform capacity over layers and tasks.
}

\subsection{Effects of Model Capacity}
\label{sec:exp:capacity}
\boldspacepar{Tiny Imagenet.} 
A crucial design choice in continual learning concerns network capacity. Aiming to learn a long task sequence, a high capacity model seems preferable.
However, learning the first task using such model, with only data from a single task, holds the risk of overfitting, jeopardizing generalization performance.
So far, we compared all methods on the \textsc{base} model. Next, we study effects of extending or reducing capacity in this architecture. 
Details of the models are given in Section~\ref{sec:exp:setup} and Appendix~A. In the following, we discuss results of model choice, again using  the random ordering of Tiny Imagenet. These results are summarized in the top part of Table~\ref{tab:exp:tiny:rnd} for parameter isolation and regularization-based methods and Table~\ref{tab:exp:tiny:replay} for replay methods%
\footnote{\label{footnote:fwdref-orderin}Note that, for all of our observations, we also checked the results obtained for the two other task orders (reported in  %
the middle and bottom part of Table~\ref{tab:exp:tiny:rnd} and Table~\ref{tab:exp:tiny:replay}). The reader can check that the observations are quite consistent. }. 

\subsubsection{Discussion}
\emphspacepar{Overfitting.} 
We observed overfitting for several models and methods, and not only for the \textsc{wide} and \textsc{deep} models. Comparing different methods, SI seems quite vulnerable to overfitting issues, while PackNet prevents overfitting to some extent, thanks to the network compression phase. 

\emphspacepar{General Observations.} 
Selecting highest accuracy disregarding which model is used, iCaRL $9$k and PackNet remain on the lead with $49.94\%$ and \matt{$49.13\%$}. Further, LwF shows to be competitive with MAS (resp. $46.79\%$ and $46.90\%$).

\emphspacepar{Baselines.} 
Taking a closer look at the finetuning baseline results (see purple box in Table~\ref{tab:exp:tiny:rnd}), we observe that it does not reach the same level of accuracy with the \textsc{small} model as with the other models. In particular, the initial accuracies are similar, yet the level of forgetting is much more severe, due to the limited model capacity to learn new tasks. 

Joint training results on the \textsc{deep} network (blue box) are inferior compared to shallower networks, implying the deep architecture to be less suited to accumulate all knowledge from the task sequence.
As performance of joint learning serves as a soft upper bound for the continual learning methods, this already serves as an indication that a deep model may not be optimal.

In accordance, replay baselines R-PM and R-FM show to be quite model agnostic with all but the \textsc{deep} model performing very similar.
The baselines experience both less forgetting and increased average accuracy when doubling replay buffer size. 

\emphspacepar{\textsc{Small} model.} Finetuning and SI suffer from severe forgetting ($>20\%$) imposed by the decreased capacity of the network (see red underlining), making these combinations worthless in practice.
Other methods experience alleviated forgetting, with EWC most saliently benefitting from a small network (see green underlinings). %

\emphspacepar{\textsc{Wide} model.} Remarkable in \textsc{wide} model results is SI consistently outperforming EWC, in contrast with other model results (see orange box; even more clear in middle and bottom tables in Table~\ref{tab:exp:tiny:rnd}, which will be discussed later). 
EWC performs worse for the high capacity \textsc{Wide} and \textsc{deep} models and, as previously discussed, attains best performance for the  \textsc{small} model.  
LwF, EBLL and PackNet mainly reach their top performance when using the \textsc{wide} model, with SI performing most stable on both the \textsc{wide} and \textsc{base} models.
IMM also shows increased performance when using the \textsc{base} and \textsc{wide} model.

\emphspacepar{\textsc{Deep} model.} Over the whole line of methods (yellow box), extending the \textsc{base} model with additional convolutional layers results in lower performance. As we already observed overfitting on the \textsc{base} model, additional layers may introduce extra unnecessary layers of abstraction.
For the \textsc{Deep} model iCaRL outperforms all continual learning methods with both memory sizes. \matt{HAT deteriorates significantly for the \textsc{deep} model, especially in the random ordering, for which we observe issues regarding asymmetric allocation of network capacity. 
This makes the method sensitive to ordering of tasks, as early fixed parameters determine performance for later tasks at saturated capacity. We discuss this in more detail in Appendix~\ref{apdx:HAT}.
}

\emphspacepar{Model agnostic.}
Over all orderings which will be discussed in Section~\ref{sec:exp:taskorder}, some methods don't exhibit a preference for any model, except for the common aversion for the detrimental \textsc{deep} model. This is in general most salient for all replay methods in Table~\ref{tab:exp:tiny:replay}, but also for PackNet, MAS, LwF and EBLL in Table~\ref{tab:exp:tiny:rnd}.

\emphspacepar{Conclusion.} We can conclude that (too) deep model architectures do not provide a good match with the continual learning setup. For the same amount of feature extractor parameters, \textsc{wide} models obtain significant better results (on average $11\%$ better over all methods in Table~\ref{tab:exp:tiny:rnd} and Table~\ref{tab:exp:tiny:replay}). Also too small models should be avoided, as the limited available capacity can cause forgetting. On average we observe a modest $0.89\%$ more forgetting for the {\sc small} model compared to the \textsc{base} model. At the same time, for some models, poor results may be due to overfitting, which can possibly be overcome using regularization, as we will study next.

\begin{table*}[]
\centering
\caption{\textbf{Parameter isolation and regularization-based methods} results on Tiny Imagenet for different models with {\em random} (top), \emph{easy to hard} (middle) and \emph{hard to easy} (bottom) ordering of tasks, reporting average accuracy (forgetting).}
\label{tab:exp:tiny:rnd}
\resizebox{\textwidth}{!}{%

\begin{tabular}{@{}lllllllllll@{}}
\toprule
\textbf{Model} & \textbf{finetuning}                                         & \textbf{joint*}                           & \textbf{PackNet}  & \textbf{HAT} & \textbf{SI}                        & \textbf{EWC}                                   & \textbf{MAS}             & \textbf{LwF}              & \textbf{EBLL}                   & \textbf{mode-IMM}                           \\ \midrule
\textsc{small} & \marktopleft{FT}16.25 \redline{(34.84)}                     & \marktopleft{joint}57.00 (\NA)            & \matt{46.68} (0.00) & 44.19 (0.00)     & 23.91 \redline{(23.26)}            & 45.13 \greenline{(0.86)}                       & 40.58 \greenline{(0.78)} & 44.06 \greenline{(-0.44)} & 44.13 \greenline{(-0.53)}  & 29.63 (3.06)                                \\
\textsc{base}  & 21.30 (26.90)                                               & 55.70 (\NA)                               & \matt{49.13} (0.00) & 43.57 (0.00)     & 33.93 (15.77)                      & 42.43 (7.51)                                   & 46.90 (1.58)             & 41.91 (3.08)              & 45.34 (1.44)                        & 36.89 (0.98)                                \\
\textsc{wide}  & 25.28 (24.15)                                               & 57.29 (\NA)                               & \matt{47.64} (0.00) & 43.78 (0.50)     & \marktopleft{SIEWC}{33.86 (15.16)} & {31.10 (17.07)}\markbottomright{SIEWC}{orange} & 45.08 (2.58)             & 46.79 (1.19)              & 46.25 (1.72)                        & 36.42 (1.66)                                \\
\textsc{deep}  & \marktopleft{deep}20.82 (20.60)\markbottomright{FT}{purple} & 51.04 (\NA) \markbottomright{joint}{blue} & \matt{35.54} (0.00) & 8.06 (3.21)     & 24.53 (12.15)                      & 29.14 (7.92)                                   & 33.58 (0.91)             & 32.28 (2.58)              & 27.78 (3.14)                       & 27.51 (0.47) \markbottomright{deep}{yellow} \\ \bottomrule
\end{tabular}%
}
\rule{0pt}{1ex}  

\resizebox{\textwidth}{!}{%
\begin{tabular}{@{}lllllllllll@{}}
\toprule
\textbf{Model} & \textbf{finetuning} & \textbf{joint*} & \textbf{PackNet}  & \textbf{HAT} & \textbf{SI}   & \textbf{EWC}  & \textbf{MAS} & \textbf{LwF} & \textbf{EBLL}  & \textbf{mode-IMM} \\ \midrule
\textsc{small} & 16.06 (35.40)       & 57.00 (\NA)     & \matt{49.21} (0.00) & 43.89 (0.00)       & 35.98 (13.02) & 40.18 (7.88)  & 44.29 (1.77) & 45.04 (1.89) & 42.07 (1.73)  & 26.13 (3.03)      \\
\textsc{base}  & 23.26 (24.85)       & 55.70 (\NA)     & \matt{50.57} (0.00) & 43.93 (-0.02)       & 33.36 (14.28) & 34.09 (13.25) & 44.02 (1.30) & 43.46 (2.53) & 43.60 (1.71)  & 36.81 (-1.17)     \\
\textsc{wide}  & 19.67 (29.95)       & 57.29 (\NA)     & \matt{47.37} (0.00) & 41.63 (0.25)       & 34.84 (13.14) & 28.35 (21.16) & 45.58 (1.52) & 42.66 (1.07) & 44.12 (3.46)  & 38.68 (-1.09)     \\
\textsc{deep}  & 23.90 (17.91)       & 51.04 (\NA)     & \matt{34.56} (0.00) & 28.64 (0.87)       & 24.56 (14.41) & 24.45 (15.22) & 35.23 (2.37) & 27.61 (3.70) & 29.73 (1.95)  & 25.89 (-2.09)     \\ %
\end{tabular}%
}

\rule{0pt}{1ex}  

\resizebox{\textwidth}{!}{%
\begin{tabular}{@{}lllllllllll@{}}
\toprule
\textbf{Model} & \textbf{finetuning} & \textbf{joint*} & \textbf{PackNet} & \textbf{HAT} & \textbf{SI}   & \textbf{EWC}  & \textbf{MAS}  & \textbf{LwF} & \textbf{EBLL}  & \textbf{mode-IMM} \\ \midrule
\textsc{small} & 18.62 (28.68)       & 57.00 (\NA)     & \matt{44.42} (0.00) & 33.52 (1.55)   & 40.39 (4.50)  & 41.62 (3.54)  & 40.90 (1.35)  & 42.36 (0.63) & 43.60 (-0.05) & 24.95 (1.66)      \\
\textsc{base}  & 21.17 (22.73)       & 55.70 (\NA)     & \matt{43.17} (0.00) & 40.21 (0.18)   & 40.79 (2.58)  & 41.83 (1.51)  & 41.98 (0.14)  & 41.58 (1.40) & 41.57 (0.82)  & 34.58 (0.23)      \\
\textsc{wide}  & 25.25 (22.69)       & 57.29 (\NA)     & \matt{45.53} (0.00) & 41.76 (0.07)   & 37.91 (8.05)  & 29.91 (15.77) & 43.55 (-0.29) & 43.87 (1.26) & 42.42 (0.85)  & 35.24 (-0.82)     \\
\textsc{deep}  & 15.33 (20.66)       & 51.04 (\NA)     & \matt{30.73} (0.00) & 27.21 (3.21)   & 22.97 (13.23) & 22.32 (13.86) & 32.99 (2.19)  & 30.77 (2.51) & 30.15 (2.67)  & 26.38 (1.20)      \\ \bottomrule
\end{tabular}%
}

\end{table*}

\begin{table*}[]
\centering
\caption{\textbf{Replay methods} results on Tiny Imagenet for different models with {\em random} (top), \emph{easy to hard} (middle) and \emph{hard to easy} (bottom) ordering of tasks, reporting average accuracy (forgetting).}
\label{tab:exp:tiny:replay}
\resizebox{\textwidth}{!}{%
\begin{tabular}{@{}lllllllllll@{}}
\toprule
\textbf{Model} & \textbf{finetuning} & \textbf{joint*} & \textbf{R-PM 4.5k} & \textbf{R-PM 9k} & \textbf{R-FM 4.5k} & \textbf{R-FM 9k} & \textbf{GEM 4.5k} & \textbf{GEM 9k} & \textbf{iCaRL 4.5k} & \textbf{iCaRL 9k} \\ \midrule
\textsc{small}          & \marktopleft{FT}16.25 (34.84)       & \marktopleft{joint}57.00 (\NA)    & 36.97 (12.21)     & 40.11 (9.23)      & 39.60 (9.57)       & 41.35 (6.33)       & 39.44 (7.38)     & 41.59 (4.17)     & 43.22 (-1.39)      & 46.32 (-1.07)      \\
\textsc{base}           & 21.30 (26.90)       & 55.70 (\NA)    & 36.09 (10.96)     & 38.69 (7.23)      & 37.31 (9.21)       & 42.36 (3.94)       & 45.13 (4.96)     & 41.75 (5.18)     & 47.27 (-1.11)      & 48.76 (-1.76)      \\
\textsc{wide}           & 25.28 (24.15)       & 57.29 (\NA)    & 36.47 (12.45)     & 41.51 (5.15)      & 39.25 (9.26)       & 41.53 (6.02)       & 40.32 (6.97)     & 44.23 (3.94)     & 44.20 (-1.43)      & 49.94 (-2.80)      \\
\textsc{deep}           & \marktopleft{deep}20.82 (20.60)\markbottomright{FT}{purple}       & 51.04 (\NA)\markbottomright{joint}{blue}    & 27.61 (7.14)      & 28.99 (6.27)      & 32.26 (3.33)       & 33.10 (4.70)       & 29.66 (6.57)     & 23.75 (6.93)     & 36.12 (-0.93)      & 37.16 (-1.64) \markbottomright{deep}{yellow}     \\ \bottomrule
\end{tabular}%
} 
\rule{0pt}{1ex}  

\resizebox{\textwidth}{!}{%
\begin{tabular}{@{}lllllllllll@{}}
\toprule
\textbf{Model} & \textbf{finetuning} & \textbf{joint*} & \textbf{R-PM 4.5k} & \textbf{R-PM 9k} & \textbf{R-FM 4.5k} & \textbf{R-FM 9k} & \textbf{GEM 4.5k} & \textbf{GEM 9k} & \textbf{iCaRL 4.5k} & \textbf{iCaRL 9k} \\ \midrule
\textsc{small} & 16.06 (35.40)       & 57.00 (\NA)    & 37.09 (11.64)      & 40.82 (8.08)     & 38.47 (9.62)       & 40.81 (7.73)     & 39.07 (9.45)      & 42.16 (6.93)    & 48.55 (-1.17)       & 46.67 (-2.03)     \\
\textsc{base}  & 23.26 (24.85)       & 55.70 (\NA)    & 36.70 (8.88)       & 39.85 (6.59)     & 38.27 (8.35)       & 40.95 (6.21)     & 30.73 (10.39)     & 40.20 (7.41)    & 46.30 (-1.53)       & 47.47 (-2.22)     \\
\textsc{wide}  & 19.67 (29.95)       & 57.29 (\NA)    & 39.43 (9.10)       & 42.55 (5.32)     & 40.67 (7.52)       & 43.23 (4.09)     & 44.49 (4.66)      & 40.78 (8.14)    & 48.13 (-2.01)       & 44.59 (-2.44)     \\
\textsc{deep}  & 23.90 (17.91)       & 51.04 (\NA)    & 30.98 (6.32)       & 32.06 (4.16)     & 29.62 (6.85)       & 33.92 (3.56)     & 25.29 (12.79)     & 29.09 (6.30)    & 32.46 (-0.35)       & 34.64 (-1.25)     \\ 
\end{tabular}%
}
\rule{0pt}{1ex}  

\resizebox{\textwidth}{!}{%
\begin{tabular}{@{}lllllllllll@{}}
\toprule
\textbf{Model} & \textbf{finetuning} & \textbf{joint*} & \textbf{R-PM 4.5k} & \textbf{R-PM 9k} & \textbf{R-FM 4.5k} & \textbf{R-FM 9k} & \textbf{GEM 4.5k} & \textbf{GEM 9k} & \textbf{iCaRL 4.5k} & \textbf{iCaRL 9k} \\ \midrule
\textsc{small} & 18.62 (28.68)       & 57.00 (\NA)    & 33.69 (11.06)      & 37.68 (6.34)     & 37.18 (7.13)       & 38.47 (4.88)     & 38.87 (6.78)      & 39.28 (7.44)    & 46.81 (-0.92)       & 46.90 (-1.61)     \\
\textsc{base}  & 21.17 (22.73)       & 55.70 (\NA)    & 32.90 (8.98)       & 34.34 (7.50)     & 33.53 (9.08)       & 36.83 (5.43)     & 35.11 (7.44)      & 35.95 (6.68)    & 43.29 (-0.47)       & 44.52 (-1.71)     \\
\textsc{wide}  & 25.25 (22.69)       & 57.29 (\NA)    & 34.85 (7.94)       & 40.28 (6.67)     & 36.70 (6.63)       & 37.46 (5.62)     & 37.27 (7.86)      & 37.91 (6.94)    & 41.49 (-1.49)       & 49.66 (-2.90)     \\
\textsc{deep}  & 15.33 (20.66)       & 51.04 (\NA)    & 24.22 (6.67)       & 23.42 (5.93)     & 25.81 (6.06)       & 29.99 (2.91)     & 27.08 (6.47)      & 31.28 (5.52)    & 30.95 (0.85)        & 37.93 (-1.35)     \\ \bottomrule
\end{tabular}%
}

\end{table*}

\subsection{Effects of Regularization}
\label{sec:exp:regul}
\boldspacepar{Tiny Imagenet.} In the previous subsection we mentioned the problem of overfitting in continual learning. Although an evident solution would be to apply regularization, this might interfere with the continual learning methods.
Therefore, we investigate the effects of two popular regularization methods, namely dropout and weight decay, for the parameter isolation and regularization-based methods in Table~\ref{tab:exp:regul:regulbased}, and for the replay methods in Table~\ref{tab:exp:regul:replaybased}.
For dropout we set the probability of retaining the units to $p=0.5$, and weight decay applies a regularization strength of $\lambda=10^{-4}$.
\review{Any form of regularization is applied in both phases of our framework.}

\begin{table*}[!th]
\centering
\caption{\textbf{Parameter isolation and regularization-based} methods: \textbf{dropout} and \textbf{weight decay} for Tiny Imagenet models.}
\label{tab:exp:regul:regulbased}

\resizebox{\textwidth}{!}{%
\begin{tabular}{@{}llllllllllllll@{}}
\toprule
\textbf{Model}      & \textit{}                  & \textbf{finetuning} & \textbf{joint*} & \textbf{PackNet} & \textbf{HAT} & \textbf{SI}   & \textbf{EWC}  & \textbf{MAS}  & \textbf{LwF}  & \textbf{EBLL} & \textbf{mode-IMM} \\ \midrule
\textsc{small}  & \textit{Plain} & 16.25 (34.84)       & 57.00 (n/a)   & 49.09 (0.00) & 44.19 (0.00)     &\marktopleft{SI}23.91 (23.26) & 45.13 (0.86)  & 40.58 (0.78)  & 44.06 (-0.44) &44.13 (-0.53)     & 29.63 (3.06)         \\
\textbf{}      & \textit{Dropout}           & 19.52 (32.62)       & \redline{55.97} (n/a)   & 50.73 (0.00) & \redline{26.87} (2.46)     & 38.34 (9.24)  & \redline{40.02} \marktopleftTIGHT{a2}(7.57)\markbottomrightTIGHT{a2}{blue}  & \redline{40.26} \marktopleftTIGHT{a7}(7.63)\markbottomrightTIGHT{a7}{blue}  & \redline{31.56} \marktopleftTIGHT{a8}(18.38)\markbottomrightTIGHT{a8}{blue} &\redline{34.14} \marktopleftTIGHT{c1}(13.20)\markbottomrightTIGHT{c1}{blue}      & \redline{29.35} (0.90)         \\
\textbf{}      & \textit{Weight Decay}      & \redline{15.06} (34.61)       & \redline{56.96} (n/a)   & 49.90 (0.00) & \redline{42.44} (0.00)     & 37.99 (6.43)  & \redline{41.22} (1.52)  & \redline{37.37} (4.43)  & \redline{41.66} (-0.28) &\redline{42.67} (-0.63)    & 30.59 (0.93)         \\ \addlinespace
\textsc{base}   & \textit{Plain} & 21.30 (26.90)       & 55.70 (n/a)   & 47.67 (0.00) & 43.57 (0.00)     & 33.93 (15.77) & 42.43 (7.51)  & 46.90 (1.58)  & 41.91 (3.08)  &45.34 \marktopleftTIGHT{c7}(1.44)\markbottomrightTIGHT{c7}{blue}    & 36.89 (0.98)         \\
\textbf{}      & \textit{Dropout}           & 29.23 (26.44)       & 61.43 (n/a)   & 54.28 (0.00) & \redline{38.11} (0.36)     & 43.15 (10.83) & 42.09 \marktopleftTIGHT{a3}(12.54)\markbottomrightTIGHT{a3}{blue} & 48.98 (0.87)  & 41.49 \marktopleftTIGHT{b1}(8.72)\markbottomrightTIGHT{b1}{blue}  &\redline{44.66} \marktopleftTIGHT{c2}(7.66)\markbottomrightTIGHT{c2}{blue}     & \redline{34.20} (1.44)         \\
\textbf{}      & \textit{Weight Decay}      & \redline{19.14} (29.31)       & 57.12 (n/a)   & 48.28 (0.00) & 44.97 (0.19)     & 39.65 (8.11)  & 44.35 (3.51)  & \redline{44.29} (1.15)  & \redline{40.91} (1.29)  &\redline{41.26} (0.82)    & 37.49 (0.46)         \\ \addlinespace
\textsc{wide}  & \textit{Plain} & 25.28 (24.15)       & 57.29 (n/a)   & 48.39 (0.00) & 43.78 (0.50)     & 33.86 (15.16) & 31.10 (17.07) & 45.08 (2.58)  & 46.79 (1.19)  &46.25 (1.72)    & 36.42 (1.66)         \\
\textbf{}      & \textit{Dropout}           & 30.76 (26.11)       & 62.27 (n/a)   & 55.96 (0.00) & \redline{34.94} (0.81)     & 43.74 (8.80)  & 33.94 \marktopleftTIGHT{a4}(19.73)\markbottomrightTIGHT{a4}{blue} & 47.92 (1.37)  & \redline{45.04} \marktopleftTIGHT{b2}(6.85)\markbottomrightTIGHT{b2}{blue}  &\redline{46.19} \marktopleftTIGHT{c3}(5.31)\markbottomrightTIGHT{c3}{blue}      & 42.41 (-0.93)        \\
\textbf{}      & \textit{Weight Decay}      & \redline{22.78} (27.19)       & 59.62 (n/a)   & \redline{47.77} (0.00) & 44.12 (0.06)     & 42.44 (8.36)  & 37.45 (13.47) & 47.24 (1.60)  & 48.11 (0.62)  &48.17 (0.82)     & 39.16 (1.41)          \\ \addlinespace
\textsc{deep}  & \textit{Plain} & 20.82 (20.60)       & 51.04 (n/a)   & 34.75 (0.00) & 8.06 (3.21)     & 24.53 (12.15) & 29.14 (7.92)  & 33.58 (0.91)  & 32.28 (2.58)  &27.78 (3.14)     & 27.51 (0.47)         \\
\textbf{}      & \textit{Dropout}           & 23.05 (27.30)       & 59.58 (n/a)   & 46.22 (0.00) & 27.54 (2.79)     & 32.76 \marktopleftTIGHT{SIdrop}(15.09)\markbottomrightTIGHT{SIdrop}{blue} & 31.16 \marktopleftTIGHT{a5}(17.06)\markbottomrightTIGHT{a5}{blue} & 39.07 \marktopleftTIGHT{c6}(5.02)\markbottomrightTIGHT{c6}{blue}  & 37.89 \marktopleftTIGHT{b3}(7.78)\markbottomrightTIGHT{b3}{blue}  &36.85 \marktopleftTIGHT{c5}(4.87)\markbottomrightTIGHT{c5}{blue}     & 33.64 (-0.61         \\
\textbf{}      & \textit{Weight Decay}      & \redline{19.48} (21.27)       & 54.60 (n/a)   & 36.99 (0.00) & 30.25 (2.02)     & 26.04 (9.51) \markbottomright{SI}{green}  & \redline{22.47} (13.19) & \redline{19.35} (13.62) & 33.15 (1.16)  &31.71 (1.39)   & \redline{25.62} (-0.15)  \\ \bottomrule

\end{tabular}%
}
\end{table*}

\subsubsection{Discussion}
\emphspacepar{General observations.} 
In Table~\ref{tab:exp:regul:regulbased} and Table~\ref{tab:exp:regul:replaybased}, negative results (with regularization hurting performance) are underlined in red. This occurs repeatedly, especially in combination with the \textsc{small} model or with weight decay. 
Over all methods and models dropout mainly shows to be fruitful. This is consistent with earlier observations~\cite{goodfellow2013empirical}. There are, however, a few salient exceptions (discussed below).
Weight decay over the whole line mainly improves the wide network accuracies. In the following, we will describe the most notable observations and exceptions to the main tendencies.

\emphspacepar{Finetuning.} 
Goodfellow et al.~observe reduced catastrophic forgetting in a transfer learning setup with finetuning when using dropout~\cite{goodfellow2013empirical}. Extended to learning a sequence of 10 consecutive tasks in our experiments, finetuning consistently benefits from dropout regularization.  This is opposed to weight decay, resulting in increased forgetting and a lower performance on the final model. In spite of good results for dropout, we regularly observe an increase in the level of forgetting, which is compensated for by starting from a better initial model, due to a reduction in overfitting. Dropout leading to more forgetting is something we also observe for many other methods (see blue boxes),
and exacerbates as the task sequence grows in length.

\emphspacepar{Joint training and PackNet. } mainly relish higher accuracies with both regularization setups.
By construction, they benefit from the regularization without interference issues. PackNet even reaches a top performance of almost $56\%$, that is $7\%$ higher than closest competitor MAS.

\matt{
\emphspacepar{HAT} mainly favours weight decay of the model parameters, excluding embedding parameters. The embeddings are already prone to sparsity regularization, for which we found troublesome learning when included for weight decay as well.
Dropout results in worse results, probably due to interfering with the unit-based masking.
However, difficulties regarding asymmetric capacity in the \textsc{deep} model seem to be alleviated using either type of regularization.
}

\emphspacepar{SI } most saliently thrives with regularization (with increases in performance of around $10\%$, see green box), which we previously found to be sensitive to overfitting. The regularization aims to find a solution to a well-posed problem, stabilizing the path in parameter space. Therefore, this might provide a better importance weight estimation along the path. 
Nonetheless, SI remains noncompetitive with leading average accuracies of other methods.

\emphspacepar{EWC and MAS} suffer from interference with additional regularization. For dropout, more redundancy in the model means less unimportant parameters left for learning new tasks. For weight decay, parameters deemed important for previous tasks are also decayed in each iteration, affecting performance on older tasks. A similar effect was noticed in~\cite{Aljundi2018_SLNI}. 
However, in some cases, the effect of this interference is again compensated for by having better initial models.
EWC only benefits from dropout on the \textsc{wide} and \textsc{deep} models, and similar to MAS prefers no regularization on the \textsc{small} model.

\emphspacepar{LwF and EBLL} suffer from using dropout, with only the \textsc{deep} model significantly improving performance. For weight decay the methods follow the general trend, enhancing the \textsc{wide} net accuracies. 

\emphspacepar{IMM} with dropout exhibits higher performance only for the \textsc{wide} and \textsc{deep} model, coming closer to the performance obtained by the other methods.

\emphspacepar{iCaRL and GEM. }
Similar to the observations in Table~\ref{tab:exp:regul:regulbased}, there is a striking increased forgetting for dropout for all methods in Table~\ref{tab:exp:regul:replaybased}.
Especially iCaRL shows inreased average forgetting, albeit consistently accompanied with higher average accuracy. 
Except for the \textsc{small} model, all models for the replay baselines benefit from dropout. 
For GEM this benefit is only notable for the \textsc{wide} and \textsc{deep} models.
Weight decay doesn't improve average accuracy for the replay methods, nor forgetting, except mainly for the \textsc{wide} model.
In general, the influence of regularization seems limited for replay methods in Table~\ref{tab:exp:regul:replaybased} compared to non-replay based methods in Table~\ref{tab:exp:regul:regulbased}.

\begin{table}[]
\centering
\caption{\textbf{Replay} methods: \textbf{dropout} (${p=0.5}$) and \textbf{weight decay} (${\lambda=10^{-4}}$) regularization for Tiny Imagenet models.}
\label{tab:exp:regul:replaybased}
\resizebox{\linewidth}{!}{%
\begin{tabular}{@{}llllll@{}}
\toprule

 \textbf{Model}                 &       & \textbf{R-PM 4.5k} & \textbf{R-FM 4.5k} & \textbf{GEM 4.5k} & \textbf{iCaRL 4.5k} \\ \midrule
\textsc{small} & \textit{Plain} & 36.97 (12.21) & 39.60 (9.57)  & 39.44 (7.38)  & 43.22 (-1.39) \\
    & \textit{Dropout}           & \redline{35.50} (12.02) & \redline{35.75} (9.87)  & \redline{33.85} (6.25)  & 44.82 \marktopleftTIGHT{b1}(4.83)\markbottomrightTIGHT{b1}{blue} \\
   \cr   & \textit{Weight Decay}      & \redline{35.57} (13.38) & \redline{39.11} (9.49)  & \redline{37.64} (5.11)  & 44.90 (-0.81) \\ \addlinespace
\textsc{base}  & \textit{Plain} & 36.09 (10.96) & 37.31 (9.21)  & 45.13 (4.96)  & 47.27 (-1.11) \\
    & \textit{Dropout}           & 43.32 (10.59) & 45.76 (7.38)  & \redline{36.09} \marktopleftTIGHT{c3}(12.13)\markbottomrightTIGHT{c3}{blue} & 48.42 \marktopleftTIGHT{a7}(2.68)\markbottomrightTIGHT{a7}{blue}  \\
\cr    & \textit{Weight Decay}      & 36.11 (10.13) & 37.78 (8.01)  & \redline{38.05} (8.74)  & \redline{45.97} (-2.32) \\ \addlinespace
\textsc{wide}  & \textit{Plain} & 36.47 (12.45) & 39.25 (9.26)  & 40.32 (6.97)  & 44.20 (-1.43) \\
      & \textit{Dropout}           & 42.20 (12.31) & 45.51 (8.85)  & 42.76 (6.33)  & 45.59 \marktopleftTIGHT{a1}(2.74)\markbottomrightTIGHT{a1}{blue}  \\
\cr    & \textit{Weight Decay}      & 39.75 (8.28)  & \redline{38.73} (10.01) & 45.27 (5.92)  & 46.58 (-1.38) \\ \addlinespace
\textsc{deep}  & \textit{Plain} & 27.61 (7.14)  & 32.26 (3.33)  & 29.66 (6.57)  & 36.12 (-0.93) \\
      & \textit{Dropout}           & 34.42 \marktopleftTIGHT{d1}(9.57)\markbottomrightTIGHT{d1}{blue}  & 37.22 \marktopleftTIGHT{d2}(7.65)\markbottomrightTIGHT{d2}{blue}  & 32.75 \marktopleftTIGHT{c2}(8.15)\markbottomrightTIGHT{c2}{blue}  & 41.77 \marktopleftTIGHT{c1}(3.58)\markbottomrightTIGHT{c1}{blue}  \\
     \cr & \textit{Weight Decay}      & \redline{26.70} (9.01)  & \redline{31.70} (4.86)  & \redline{27.26} (5.94)  & \redline{33.70} (-0.47) \\ \bottomrule
\end{tabular}%
}
\end{table}

\subsection{Effects of a Real-world Setting} %
\label{sec:exp:largedataset}

\begin{figure*}%
\center
      \makebox[\textwidth][c]{\includegraphics[width=1\linewidth]{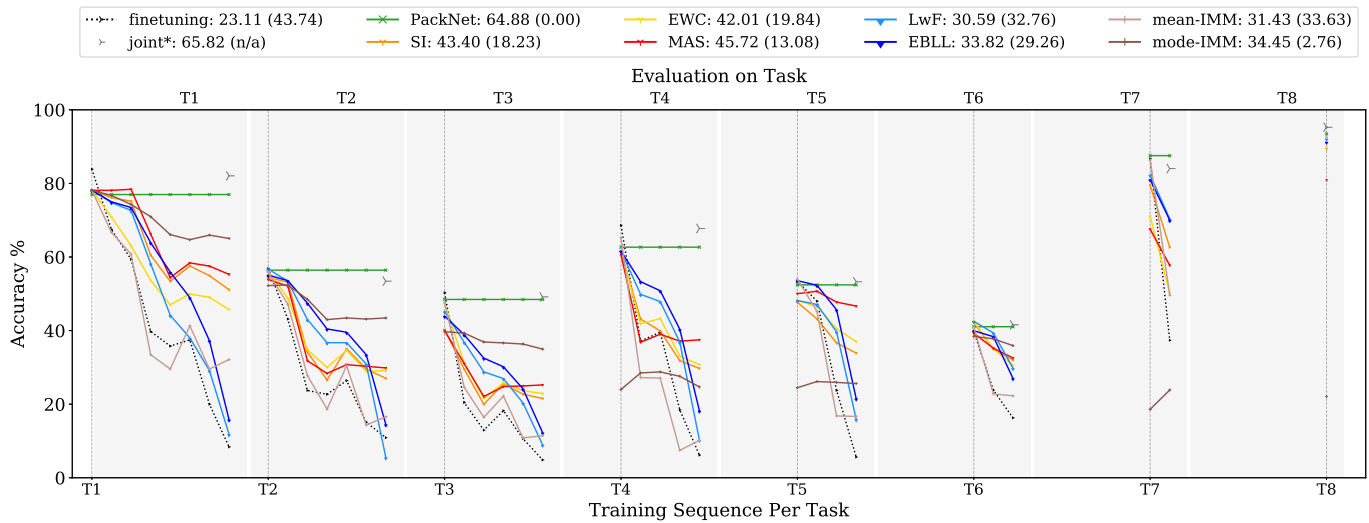}}
     \caption{RecogSeq dataset sequence results, reporting average accuracy (forgetting).}
\label{fig:exp:recogseq}
\end{figure*}

\boldspacepar{iNaturalist and RecogSeq.} Up to this point all experiments conducted on Tiny Imagenet are nicely balanced, with an equal amount of data and classes per task. In further experiments 
we scrutinize highly unbalanced task sequences, both in terms of classes and available data per task.
\matt{
We train AlexNet pretrained on Imagenet, and track task performance for RecogSeq and iNaturalist in Figure~\ref{fig:exp:recogseq} and Figure~\ref{fig:inat:ordering:rnd}.} The experiments exclude both
\matt{
noncompetitive HAT results due to problems of asymmetric capacity allocation discussed in Appendix~\ref{apdx:HAT}, and replay methods as they lack a policy to cope with unbalanced data, which would make comparison highly biased to our implementation.
}

\subsubsection{Discussion}
\boldspacepar{iNaturalist.}
Overall PackNet shows the highest average accuracy, tailgated by mode-IMM with superior forgetting by exhibiting positive backward transfer.
Both PackNet and mode-IMM attain accuracies very close or even surpassing joint training, such as for Task 1 and Task 8.

\emphspacepar{Performance drop.} Evaluation on the first 4 tasks shows salient dips when learning Task 5 for prior-based methods EWC, SI and MAS. In the random ordering Task 5 is an extremely easy task of supercategory 'Fungi' which contains only 5 classes and few data. Using expert gates to measure relatedness, the first four tasks show no particularly salient relatedness peaks or dips for Task 5. 
Instead, forgetting might rather be caused by the limited amount of training data, with only a few target classes, enforcing the network to overly fit to this task. 

\emphspacepar{SI.} For iNaturalist we did not observe overfitting, which might be the cause for stable SI behaviour in comparison to Tiny Imagenet. 

\emphspacepar{LwF and EBLL}. LwF catastrophically forgets on the unbalanced dataset with $13.77\%$ forgetting, which is in high contrast with the results acquired on Tiny Imagenet. The supercategories in iNaturalist constitute a completely different task, imposing severe distribution shifts between the tasks. In the contrary, Tiny Imagenet is constructed from a subset of randomly collected classes, implying similar levels of homogeneity between task distributions. 
On top, EBLL constraints the new task features to reside closely to the optimal presentation for previous task features, which for the random ordering enforces forgetting to nearly halve from $13.77\%$ to $7.51\%$, resulting in a striking $7.91\%$ increase in average accuracy over LwF (from $45.39\%$ to $53.30\%$).

\matt{
\boldspacepar{RecogSeq} shows similar findings to iNaturalist for LwF and EBLL, exacerbating in performance as they are subject to severe distribution shifts between tasks. This is most notable for the fierce forgetting when learning the last SVHN task. 
Besides the data-focused methods, we also observe deterioration for the prior-focused regularization-based methods. This shows increasing difficulty of these methods to cope with highly different recognition tasks.
By contrast, PackNet thrives in this setup, taking the lead by high margin and approaching joint performance. In its advantage, PackNet freezes the previous task parameters, to impose zero forgetting regardless of the task dissimilarities.
For all continual learning methods, Appendix~B.4 reports reproduced results similar to the original RecogSeq setup \cite{aljundi2017expertgate, aljundi2018memory}, showcasing hyperparameter sensitivity and an urge for hyperparameter robustness. 
}

\subsection{Effects of Task Order}
\label{sec:exp:taskorder}
In this experiment we scrutinize how changing the task order affects both the balanced (Tiny Imagenet) and unbalanced (iNaturalist) task sequences. Our hypothesis resembles curriculum learning \cite{curriculum}, implying knowledge is better captured starting with the general easier tasks followed by harder specific tasks.
Instead, experimental results on both datasets exhibit ordering agnostic behaviour, corresponding to similar previous findings \cite{nguyen2019toward}.

\par\noindent
\textbf{Tiny Imagenet.} The previously discussed results of parameter isolation and regularization-based methods on top of Table~\ref{tab:exp:tiny:rnd}, and replay methods on top of Table~\ref{tab:exp:tiny:replay} concern a random ordering.
We define a new order based on task difficulty, by measuring the accuracy over the 4 models obtained on held-out datasets for each of the tasks.
This results in an ordering from \emph{hard to easy} consisting of task sequence [5, 7, 10, 2, 9, 8, 6, 4, 3, 1] from the random ordering, with the inverse order equal to the \emph{easy to hard} ordering.

\boldspacepar{iNaturalist} constitutes three orderings defined as follows. 
First, we alphabetically order the supercategories to define the random ordering, as depicted in Figure~\ref{fig:inat:ordering:rnd}.
Further, the two other orderings start with task 'Aves' comprising most data, whereafter the remaining tasks are selected based on the average relatedness to the already included tasks in the sequence. Relatedness is measured using Expert Gate autoencoders following \cite{aljundi2017expertgate}. 
Selecting on highest relatedness results in the \emph{related ordering} in Figure~\ref{fig:inat:ordering:rel-to-unrel}, whereas selecting on lowest relatedness ends up with the  \emph{unrelated ordering} in Figure~\ref{fig:inat:ordering:unrel-to-rel}.
We refer to Table~\ref{tab:ds:inat} for 
the configuration of the three different orderings.

\begin{table}[]
\centering
\caption{The unbalanced iNaturalist task sequence details for random, related, and unrelated orderings.}
\label{tab:ds:inat}
\resizebox{\linewidth}{!}{%
\begin{tabular}{@{}l|l|lll|lll@{}}
\toprule
\multicolumn{1}{c}{\textbf{Task}} & \multicolumn{1}{c}{\textbf{Classes}}        & \multicolumn{3}{c}{\textbf{Samples}}        &        \multicolumn{3}{c}{\textbf{Ordering}}                              \\ \midrule
\textit{}     &                  & \textit{Train} & \textit{Val} & \textit{Test} & \textit{Rand.} & \textit{Rel.} & \textit{Unrel.} \\
Amphibia      & 28               & 5319              & 755                 & 1519          & 1               & 4                             & 10                            \\
Animalia      & 10               & 1388              & 196                 & 397           & 2               & 5                             & 9                             \\
Arachnida     & 9                & 1192              & 170                 & 341           & 3               & 8                             & 7                             \\
Aves          & 314              & 65897             & 9362                & 18827         & 4               & 1                             & 1                             \\
Fungi         & 5                & 645               & 91                  & 184           & 5               & 6                             & 2                             \\
Insecta       & 150              & 26059             & 3692                & 7443          & 6               & 9                             & 3                             \\
Mammalia      & 42               & 8646              & 1231                & 2469          & 7               & 2                             & 8                             \\
Mollusca      & 13               & 1706              & 241                 & 489           & 8               & 7                             & 4                             \\
Plantae       & 237              & 35157             & 4998                & 10045         & 9               & 10                            & 5                             \\
Reptilia      & 56               & 10173             & 1447                & 2905          & 10              & 3                             & 6                             \\ \bottomrule
\end{tabular}%
}
\end{table}

\subsubsection{Discussion}

\boldspacepar{Tiny Imagenet.}
The main observations on the random ordering for the \textsc{base} model in Section~\ref{sec:exp_base} and model capacity in Section~\ref{sec:exp:capacity} remain valid for the two additional orderings, with PackNet and iCaRL competing for highest average accuracy, subsequently followed by MAS and LwF. 
In the following, we will instead focus on general observations between the three different orderings.

\emphspacepar{Task Ordering Hypothesis.}
Starting from our curriculum learning hypothesis we would expect the easy-to-hard ordering to enhance performance w.r.t. the random ordering, and the opposite effect for the hard-to-easy ordering. 
However, especially SI and EWC show unexpected better results for the hard-to-easy ordering than for the easy-to-hard ordering. 
For PackNet and MAS, we see a systematic improvement when switching from hard-to-easy to easy-to-hard ordering. The gain is, however, relatively small. 
Overall, the impact of the task order seems insignificant.

\emphspacepar{Replay Buffer.}
Introducing the two other orderings, we now observe that iCaRL doesn't always improve when increasing the replay buffer size for the easy-to-hard ordering (see \textsc{small} and \textsc{wide} model). 
More exemplars induce more samples to distill knowledge from previous tasks, but might deteriorate stochasticity in the estimated feature means in the nearest-neighbour classifier.
GEM does not as consistently benefit from increased replay buffer size (GEM $9$k), which could also root in the reduced stochasticity of the constraining gradients from previous tasks.

\boldspacepar{iNaturalist.} The general observations for the random ordering remain consistent for the other orderings as well, with PackNet and mode-IMM the two highest performing methods. 

\matt{\emphspacepar{Surpassing Joint training. } In all three orderings, PackNet consistently performs slightly better than Joint training for the \emph{Mollusca} task. This is an easy task with only 13 classes to discriminate, with PackNet enabling beneficial forward transfer from previous tasks.
}

\begin{figure*}[!hbtp]
\center
\begin{subfigure}[b]{1\textwidth}
\centering
  \caption{Random ordering.}
      \makebox[\textwidth][c]{\includegraphics[width=1\linewidth]{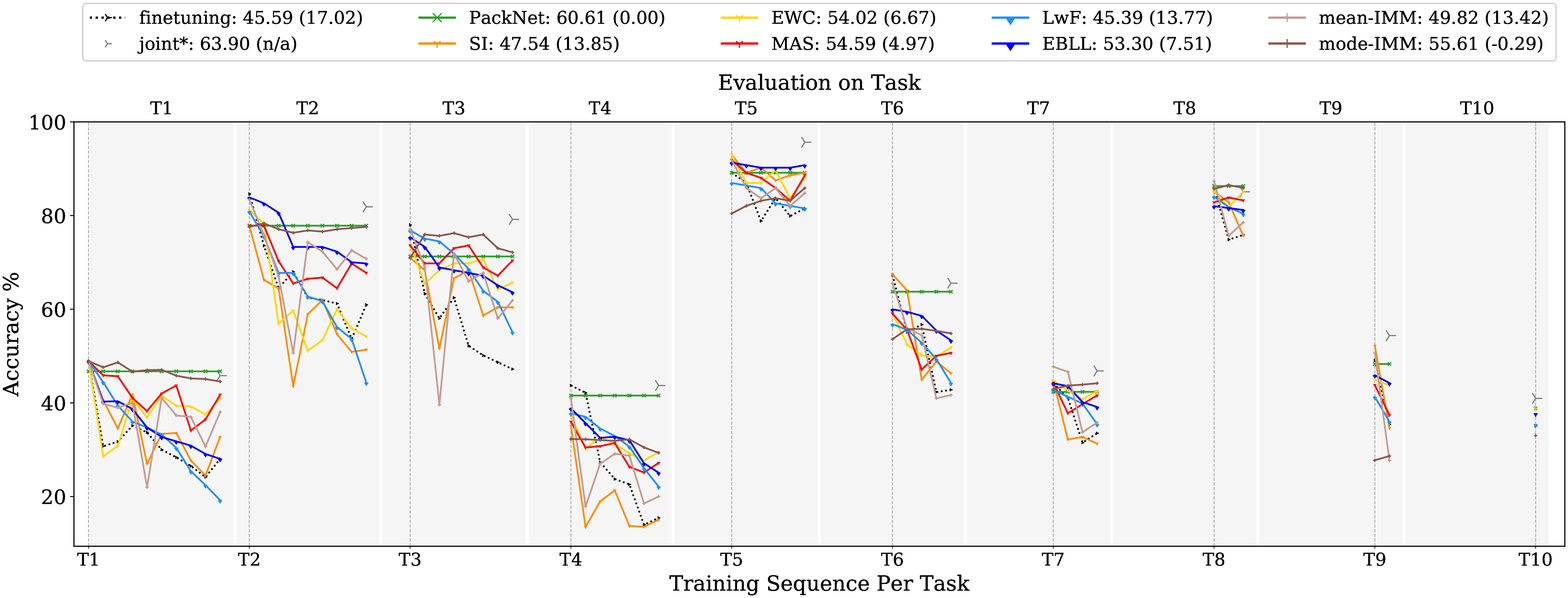}}
\label{fig:inat:ordering:rnd}
\end{subfigure}

\begin{subfigure}[b]{1\textwidth}
\centering
  \caption{Related ordering.}
          \makebox[\textwidth][c]{\includegraphics[width=1\linewidth]{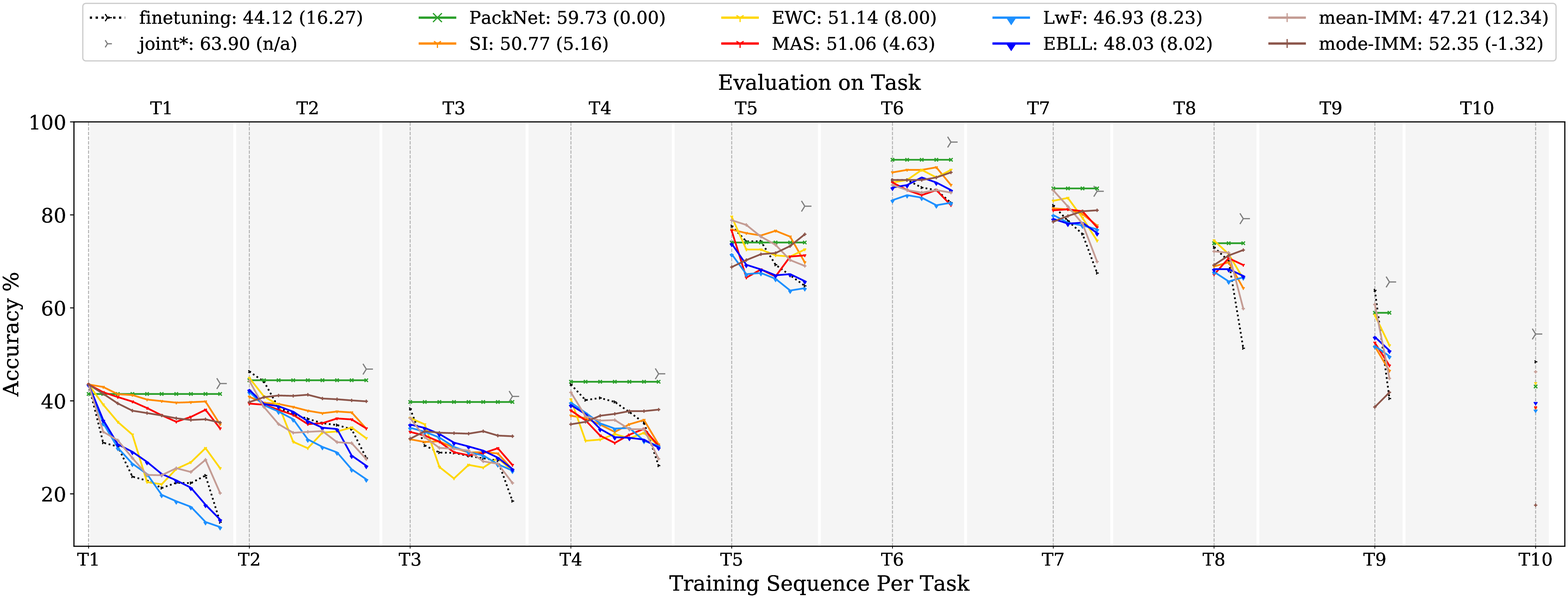}}
  \label{fig:inat:ordering:rel-to-unrel}
\end{subfigure}

\begin{subfigure}[b]{1\textwidth}
\centering
  \caption{Unrelated ordering.}
      \makebox[\textwidth][c]{\includegraphics[width=1\linewidth]{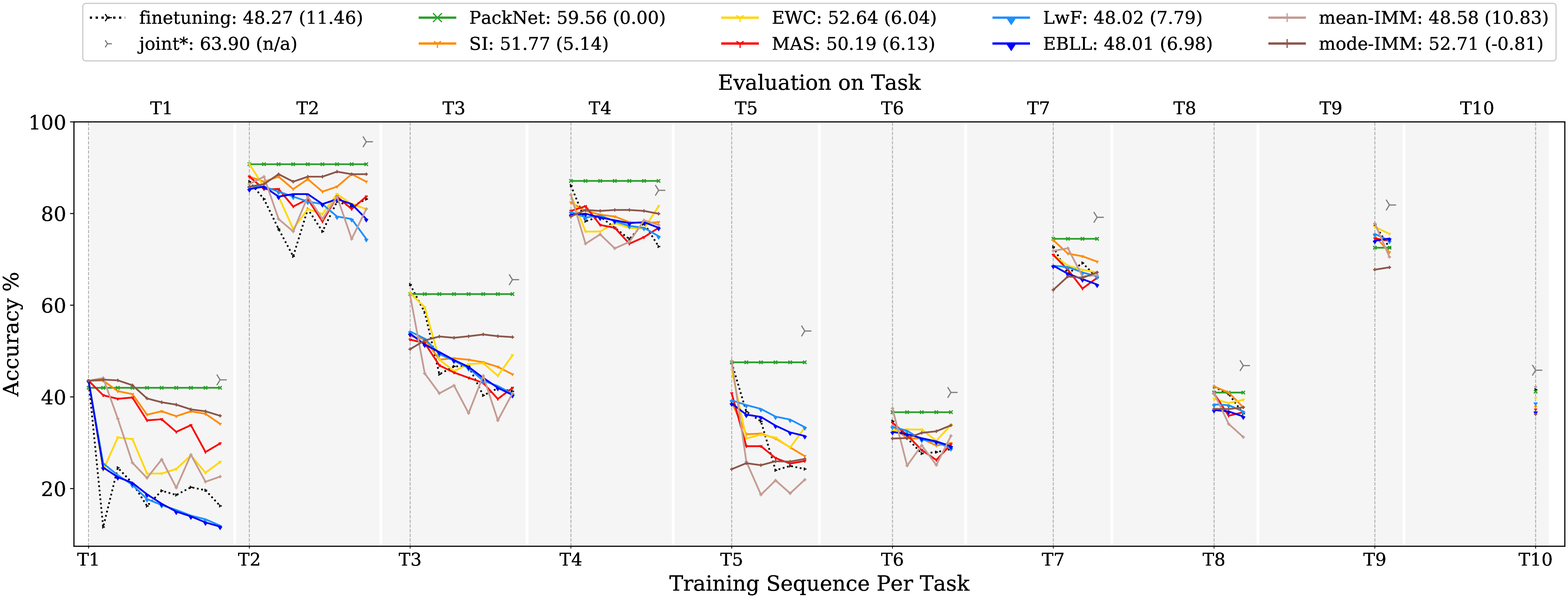}}

  \label{fig:inat:ordering:unrel-to-rel}
\end{subfigure}

\caption{Continual learning methods accuracy plots for 3 different orderings of the iNaturalist dataset.}
\label{fig:exp:capacity}
\end{figure*}

\emphspacepar{Small Task Dip.} 
For the random ordering we observed a dip in accuracies for all preceding tasks when learning the very small \emph{Fungi} task. Furthermore, for the unrelated ordering,  the \emph{Fungi} (Task 2) also experiences a performance dip for finetuning and EWC in Figure~\ref{fig:inat:ordering:unrel-to-rel} (see leftmost evaluation panel of Task 1).
On top, EWC shows the same behaviour for \emph{Fungi} (Task 6) also in the related ordering (Figure~\ref{fig:inat:ordering:rel-to-unrel}), and therefore seems to be most susceptible to highly varying task constitutions, and especially smaller tasks.

\emphspacepar{LwF and EBLL.} Where EBLL improves LwF with a significant $7.91\%$ accuracy in the random ordering, and 
$1.1\%$ in the related ordering, it performs similar to LwF for the unrelated ordering. %

\emphspacepar{SI and First Task.} As observed on Tiny Imagenet as well, SI is sensitive to overfitting when estimating importance weights through the path integral. 
The same behaviour can be observed for the random ordering which starts with a rather small task \emph{Amphibia} ($5$k training samples) showing very unstable learning curves.
The two other orderings start from task \emph{Aves} ($65$k training samples) from which the amount of data might act as a regularizer and prevent overfitting. Hence, the path integral can be stabilized, resulting in very competitive results compared to EWC and MAS.

\begin{table*}[ht]
\caption{\label{tab:qualitative}Qualitative comparison of the compared continual learning methods. The heatmap gives relative numbers to the minimal observed value in the column. \matt{Denoting number of seen tasks ($\mathcal{T}$), replay buffer size ($R$), autoencoder size ($A$), size for model parameters ($M$), with corresponding binary parameter mask $M[bit]$, and unit-based embeddings ($U$).
}}
\centering
\noindent\begin{minipage}{.9\linewidth}%
\centering
\begin{tabular}{@{}llllllccl@{}}
\toprule
\textbf{Category} & 
\textbf{Method} & 
\multicolumn{2}{c}{\textbf{Memory}} & 
\multicolumn{2}{c}{\textbf{Compute}} &
\textbf{\begin{tabular}[c]{@{}c@{}}Task-agnostic\\ possible\end{tabular}} & \textbf{\begin{tabular}[c]{@{}c@{}}Privacy\\ issues\end{tabular}} & 
\textbf{\begin{tabular}[c]{@{}c@{}}Additional required\\ storage\end{tabular}} \\ \midrule
 &  & \textit{train} & \textit{test} & \textit{train} & \textit{test} & &  & \\
\textbf{Replay-based}      & iCARL    & \cellcolor[HTML] {66a549} $1.24$ & \cellcolor[HTML] {d0da68} $1.00$ & \cellcolor[HTML] {f6ec73} $5.63$  & \cellcolor[HTML] {e74c3c} $45.61$ & \checkmark & \checkmark & $M + R$                       \\ 
\textbf{}                  & GEM      & \cellcolor[HTML] {459440} $1.07$ & \cellcolor[HTML] {feea73} $1.29$ & \cellcolor[HTML] {e74c3c} $10.66$ & \cellcolor[HTML] {4f9942} $3.64$  & \checkmark & \checkmark & $\mathcal{T} \cdot M + R$     \\ \hline
\textbf{Reg.-based}        & LwF      & \cellcolor[HTML] {44943f} $1.07$ & \cellcolor[HTML] {e2e26d} $1.10$ & \cellcolor[HTML] {43933f} $1.29$  & \cellcolor[HTML] {3f913e} $1.86$  & \checkmark & \xmark     & $M$                           \\
\textbf{}                  & EBLL     & \cellcolor[HTML] {9ec059} $1.53$ & \cellcolor[HTML] {dfe16c} $1.08$ & \cellcolor[HTML] {6ba74a} $2.24$  & \cellcolor[HTML] {3b8f3c} $1.34$  & \checkmark & \xmark     & $M + \mathcal{T} \cdot A$     \\
\textbf{}                  & SI       & \cellcolor[HTML] {4a9741} $1.09$ & \cellcolor[HTML] {dade6b} $1.05$ & \cellcolor[HTML] {3d903d} $1.13$  & \cellcolor[HTML] {3d903d} $1.61$  & \checkmark & \xmark     & $3 \cdot M$                   \\
\textbf{}                  & EWC      & \cellcolor[HTML] {4a9641} $1.09$ & \cellcolor[HTML] {d9de6b} $1.05$ & \cellcolor[HTML] {3c903d} $1.11$  & \cellcolor[HTML] {3f913e} $1.88$  & \checkmark & \xmark     & $2 \cdot M$                   \\
\textbf{}                  & MAS      & \cellcolor[HTML] {499641} $1.09$ & \cellcolor[HTML] {d9de6b} $1.05$ & \cellcolor[HTML] {3e913d} $1.16$  & \cellcolor[HTML] {3f913e} $1.88$  & \checkmark & \xmark     & $2 \cdot M$                   \\
\textbf{}                  & mean-IMM & \cellcolor[HTML] {398e3c} $1.01$ & \cellcolor[HTML] {d5dc69} $1.03$ & \cellcolor[HTML] {3b8f3d} $1.09$  & \cellcolor[HTML] {398e3c} $1.18$  & \checkmark & \xmark     & $\mathcal{T} \cdot M$         \\
\textbf{}                  & mode-IMM & \cellcolor[HTML] {398e3c} $1.01$ & \cellcolor[HTML] {d5dc69} $1.03$ & \cellcolor[HTML] {41923e} $1.24$  & \cellcolor[HTML] {388e3c} $1.00$  & \checkmark & \xmark     & $2 \cdot \mathcal{T} \cdot M$ \\ \hline
\textbf{Param. iso.-based} & PackNet  & \cellcolor[HTML] {388e3c} $1.00$ & \cellcolor[HTML] {ef8750} $1.94$ & \cellcolor[HTML] {7caf4f} $2.66$  & \cellcolor[HTML] {44943f} $2.40$  & \xmark     & \xmark     & $\mathcal{T} \cdot M[bit]$    \\
                           & HAT      & \cellcolor[HTML] {61a248} $1.21$ & \cellcolor[HTML] {f0e971} $1.17$ & \cellcolor[HTML] {388e3c} $1.00$  & \cellcolor[HTML] {41923e} $2.06$  &             \xmark& \xmark & $\mathcal{T} \cdot U$                              \\ 
\hline
\end{tabular}%
\end{minipage}%
\begin{minipage}{.1\linewidth} %
\begin{tikzpicture}[every node/.style={inner sep=0,outer sep=0}, baseline=(current bounding box.center)]
\draw[top color=heatmap_green,
      bottom color=heatmap_red, 
      middle color=heatmap_yellow,
      ]
 (0,0) rectangle (0.5,1.2);
 \node[text width=1.5cm, anchor=north, right] at (0.7,1) {Low};
  \node[text width=1.5cm, anchor=north, right] at (0.7,0.2) {High};
\end{tikzpicture}\hspace{-5mm} %

\end{minipage}
\vspace{1mm}
\end{table*}

\subsection{Qualitative Comparison}
\label{sec:exp:qual}
The previous experiments have a strong focus on coping with catastrophic forgetting, whereas by construction each method features additional advantages and limitations w.r.t. the other methods.
In Table~\ref{tab:qualitative} we visualize relative GPU memory requirements and computation time for each of the methods in a heatmap, and formalize the extra required storage during the lifetime of the continual learner. Further, we emphasize task-agnosticity and privacy in our comparison.
Note that we present the results in an indicative heatmap for a more high level relative comparison, as the results highly depend on our implementation.
Details of the setup are provided in Appendix~\ref{apdx:implementation-details}.

\emphspacepar{GPU Memory} for training is slightly higher for EBLL, which requires the autoencoder to be loaded in memory as well.
At test time all memory consumption is more or less equal, but shows some deviations for GEM, HAT and PackNet due to our implementation. 
PackNet and HAT respectively load additional masks and embeddings in memory for evaluation. However, the \textsc{base} model size (54MB), and hence additional mask or embedding size, are rather small compared to concurrently loaded data ($200$ images of size $64 \times 64 \times 3$ ). 
Double test time memory for PackNet is due to our implementation loading all masks at once for faster prediction, with the cost of increased memory consumption of the masks (factor $10$ increase, one mask per task).

\emphspacepar{Computation time} for training is doubled for EBLL and PackNet (see light green), as they both require an additional training step, respectively for the autoencoder and compression phase. 
iCaRL training requires additional forward passes for exemplar sets (factor 5 increase). 
On top of that, GEM requires backward passes to acquire exemplar set gradients, and solves a quadratic optimization problem (factor 10 increase). 
During testing, the iCaRL nearest-neighbor classifier is vastly outperformed by a regular softmax classifier in terms of computation time (factor 45 increase). 

\emphspacepar{Storage} indicates the additional memory required for the continual learning method. LwF and iCaRL both store the previous task model for knowledge distillation (although one could also store the recorded outputs instead). 
On top of that, EBLL stores previous task autoencoders, small in size $A$ relative to $M$. Further, iCarl and GEM store exemplars, with GEM additional  exemplar gradients.
Prior-focused methods EWC and MAS store importance weights and the previous task model, with SI also requiring a running estimate of the current task importance weights.
IMM, when naively implemented, stores for each task the model after training, with mode-IMM additionally storing importance weights. More storage efficient implementations propagate a merged model instead. %
Finally, PackNet requires task-specific masks equal to the amount parameters in the model ($M[bit]$), and HAT obtains unit-based embeddings.

\emphspacepar{Task-agnostic and privacy.} At deployment, PackNet and HAT require the task identifier for a given sample to load the appropriate masks and embeddings. Therefore, this setup inherently prevents task-agnostic inference with a single shared head between all tasks. Next, we emphasize the privacy issues for replay-based methods iCaRL and GEM, storing raw images as exemplars.

\makeatletter
\tikzset{vertical custom shading/.code={%
 \pgfmathsetmacro\tikz@vcs@middle{#1}
 \pgfmathsetmacro\tikz@vcs@bottom{\tikz@vcs@middle/2}
 \pgfmathsetmacro\tikz@vcs@top{(100-\tikz@vcs@middle)/2+\tikz@vcs@middle}
\pgfdeclareverticalshading[tikz@axis@top,tikz@axis@middle,tikz@axis@bottom]{newaxis}{100bp}{%
  color(0bp)=(tikz@axis@bottom);
  color(\tikz@vcs@bottom bp)=(tikz@axis@bottom);
  color(\tikz@vcs@middle bp)=(tikz@axis@middle);
  color(\tikz@vcs@top bp)=(tikz@axis@top);
  color(100bp)=(tikz@axis@top)}
  \pgfkeysalso{/tikz/shading=newaxis}
  }
}
\makeatother

\subsection{Experiments Summary}
\label{sec:exp:summary}
We summarize our main findings for each method in Table~\ref{tab:summary}.
\review{
Overall, PackNet seems the best performer by high margin on all three datasets, tailored for the task incremental multi-head setup.
However, it refrains applicability in class incremental setups and is confined to the initial model capacity, as exemplified in supplemental experiments in Appendix~B.6.
In the regularization-based methods, MAS exhibits hyperaparameter robustness and consistent lead performance over the datasets. %
Competing mode-IMM thrives in the iNaturalist setup, but lacks robustness to the other setups.
Further, LwF and EBLL show competitive performance for relatively similar tasks, but collapse for the significant distribution shifts in RecogSeq.
Replay-methods iCarl and GEM are not designed for a task incremental setup, but nonetheless show competitive performance with iCarl improving on larger memory size.
}

\review{
Weight decay and dropout show consistent merits only for specific methods and increased model capacity in \textsc{deep} models exhibits inferior performance in the scope of our experiments. Finally, presenting the tasks in different orderings to the continual learner has insignificant performance effects.
}

\begin{table*}[!ht]
\centering
\caption{Summary of our main findings. We report best results over all experiments, i.e. including regularization experiments for Tiny Imagenet. The \textsc{small, base, wide} models are denoted as \textsc{s,b,w}, and weight decay as L2. 
}
\begin{tabular}{@{}lllllll@{}}
\toprule

\multicolumn{1}{l}{\textbf{Method}} & \multicolumn{3}{c}{\textbf{\begin{tabular}[c]{@{}c@{}}Best Avg. Acc.\end{tabular}}} & \multicolumn{1}{c}{\textbf{\begin{tabular}[c]{@{}c@{}}Suggested\\Regularizer\end{tabular}}} & \multicolumn{1}{c}{\textbf{\begin{tabular}[c]{@{}c@{}}Suggested\\ Model\end{tabular}}} & \multicolumn{1}{c}{\textbf{Comments}} \\ 
& Tiny Imagenet  & iNaturalist  & RecogSeq   & \multicolumn{1}{c}{}&\multicolumn{1}{c}{}  & \\\midrule

\textbf{Replay} &  &  &   & &  & \begin{tabular}[t]{@{}l@{}} - Least sensitive model capacity/regularization \\ - Privacy issues storing raw images \\ - No clear policy for unbalanced tasks \end{tabular}  \\ \addlinespace
iCaRL 4.5k (9k) & 48.55 (49.94) & x & x & dropout & \textsc{s}/\textsc{b}/\textsc{w} & \begin{tabular}[t]{@{}l@{}}- Lead performance \\ - Designed for class incremental setup\\ - Continual exemplar management\\ - Nearest-neighbor classifier\end{tabular} \\ \addlinespace
GEM 4.5k (9k) & 45.27 (44.23) & x & x & none/dropout & \textsc{s}/\textsc{b}/\textsc{w} & \begin{tabular}[t]{@{}l@{}} - Designed for online continual setup \\ - Sensitive to amount of epochs\end{tabular} \\ \midrule
\multicolumn{2}{@{}l}{\textbf{Regularization-based}}  &  & & &  &  \\
LwF & 48.11 & 48.02 & 30.59 & L2 & \textsc{w} &  \begin{tabular}[t]{@{}l@{}} - Invigorated by \textsc{wide} model \\ - Requires sample outputs on previous model \\ \matt{- Forgetting buildup for older dissimilar tasks}\end{tabular}\\ \addlinespace
EBLL & 48.17 & 53.30 & 33.82  & L2 & \textsc{w} &  \begin{tabular}[t]{@{}l@{}} - Margin over LwF \\ - Autoencoder gridsearch for unbalanced tasks\end{tabular}\\ \addlinespace
SI & 43.74 & 51.77 & 43.40  & dropout/L2 & \textsc{b}/\textsc{w} & \begin{tabular}[t]{@{}l@{}}- Efficient training time over EWC/MAS \\ - Requires dropout or L2 (prone to overfitting) \\ - Most affected by task ordering \\ \end{tabular} \\ \addlinespace
EWC & 45.13 & 54.02 & 42.01  & none & \textsc{s} & \begin{tabular}[t]{@{}l@{}} - Invigorated by \textsc{small} capacity model \\ - Deteriorates on \textsc{wide} model\end{tabular} \\ \addlinespace
MAS & 48.98 & 54.59 & 45.72 & none & \textsc{b}/\textsc{w} & \begin{tabular}[t]{@{}l@{}}- Lead regularization-based performance \\ - Hyperparameter robustness \\ - Unsupervised importance weight estimation\end{tabular} \\ \addlinespace

mean-IMM & 32.42 & 49.82 & 31.43 & none & \textsc{b}/\textsc{w} &  - mode-IMM outperforms mean-IMM \\

mode-IMM & 42.41 & 55.61 & 34.45& none & \textsc{b}/\textsc{w} & \begin{tabular}[t]{@{}l@{}}
- Both require additional merging step\end{tabular} \\ \midrule
\textbf{Parameter isolation} &  &  &  & &  &  \begin{tabular}[t]{@{}l@{}} \matt{- Efficient memory
}\\ \review{- Prevents scalable class incremental setup} \\   \end{tabular}\\\addlinespace
PackNet & 55.96 & \matt{60.61} & \begin{tabular}[t]{@{}l@{}}
64.88 \end{tabular}& dropout/L2 & \textsc{s}/\textsc{b}/\textsc{w} & \begin{tabular}[t]{@{}l@{}}- Lead performance \\ - No forgetting (after compression) \\ 
- Requires retraining after compression \\ - Heuristic parameter pruning \end{tabular} \\ \addlinespace
HAT & 44.19 & x & x & L2 & \textsc{b}/\textsc{w}&  \begin{tabular}[t]{@{}l@{}}- Nearly no forgetting (nearly binary masks) \\ - End-to-end attention mask learning  \\ - Saturating low-level feature capacity \end{tabular}\\
\bottomrule
\end{tabular}
\label{tab:summary}
\end{table*}

\section{Looking ahead}
\label{sec:future}
In the  experiments performed in this survey, we have considered an task incremental learning setting. In this setting, tasks are received one at the time and offline training is performed on their associated training data until convergence. Knowledge of the task boundaries is required (i.e. when tasks switch) and allows for multiple passes over large batches of training data. Hence, it resembles a relaxation of the desired continual learning system that is more likely to be encountered in practice. Below, we describe the general setting in which continual learning methods are expected to be deployed, and outline the main characteristics to be considered by upcoming methods.
\newline
{\bf The General Continual Learning setting} considers an infinite stream of training data where at each time step, the system receives a (number of) new sample(s) drawn non i.i.d from a current distribution $D_t$ that could itself experience sudden or gradual changes. 
The main goal is to optimally learn a function that minimizes a predefined loss on the new sample(s) without interfering with, and possibly improving on, those that were learned previously.
Desiderata of an ideal continual learning scheme include:

\noindent\reviewfinal{
\textbf{1. Constant memory.} To avoid unbounded systems, the consumed memory should be constant w.r.t. the number of tasks or length of the data stream. \\
\textbf{2. No task boundaries.} Learning from the input data without requiring clear task divisions makes continual learning applicable to any never-ending data stream.\\
\textbf{3. Online learning} without demanding offline training of large batches or separate tasks introduces fast acquisition of new information.\\
\textbf{4. Forward transfer} or zero-shot learning indicates the importance of previously acquired knowledge to aid the learning of new tasks by increased data efficiency.\\
\textbf{5. Backward transfer} aims at retaining previous knowledge and preferably improving it when learning future related tasks.\\
\textbf{6. Problem agnostic} continual learning is not limited to a specific setting (e.g. only classification).\\
\textbf{7. Adaptive} systems learn from available unlabeled data as well, opening doors for adaptation to specific user data. \\
\textbf{8. No test time oracle} providing the task label
should be required for prediction.\\
\textbf{9. Task revisiting} of previously seen tasks should enable enhancement of the corresponding task knowledge.\\
\textbf{10. Graceful forgetting.} Given an unbounded system and infinite stream of data, selective forgetting of trivial information is an important mechanism to achieve balance between stability and plasticity. 
}

After establishing the general continual learning setting and the main desiderata, it is important to identify the critical differences between continual learning and other closely related machine learning fields that share some of the continual learning characteristics mentioned above.

\section{Related machine learning fields}
\label{sec:otherfields}
\begin{figure*}[!ht]
    \center{\includegraphics[clip,trim={3.6cm 15.7cm 0.2cm 3.3cm},width=.75\textwidth]{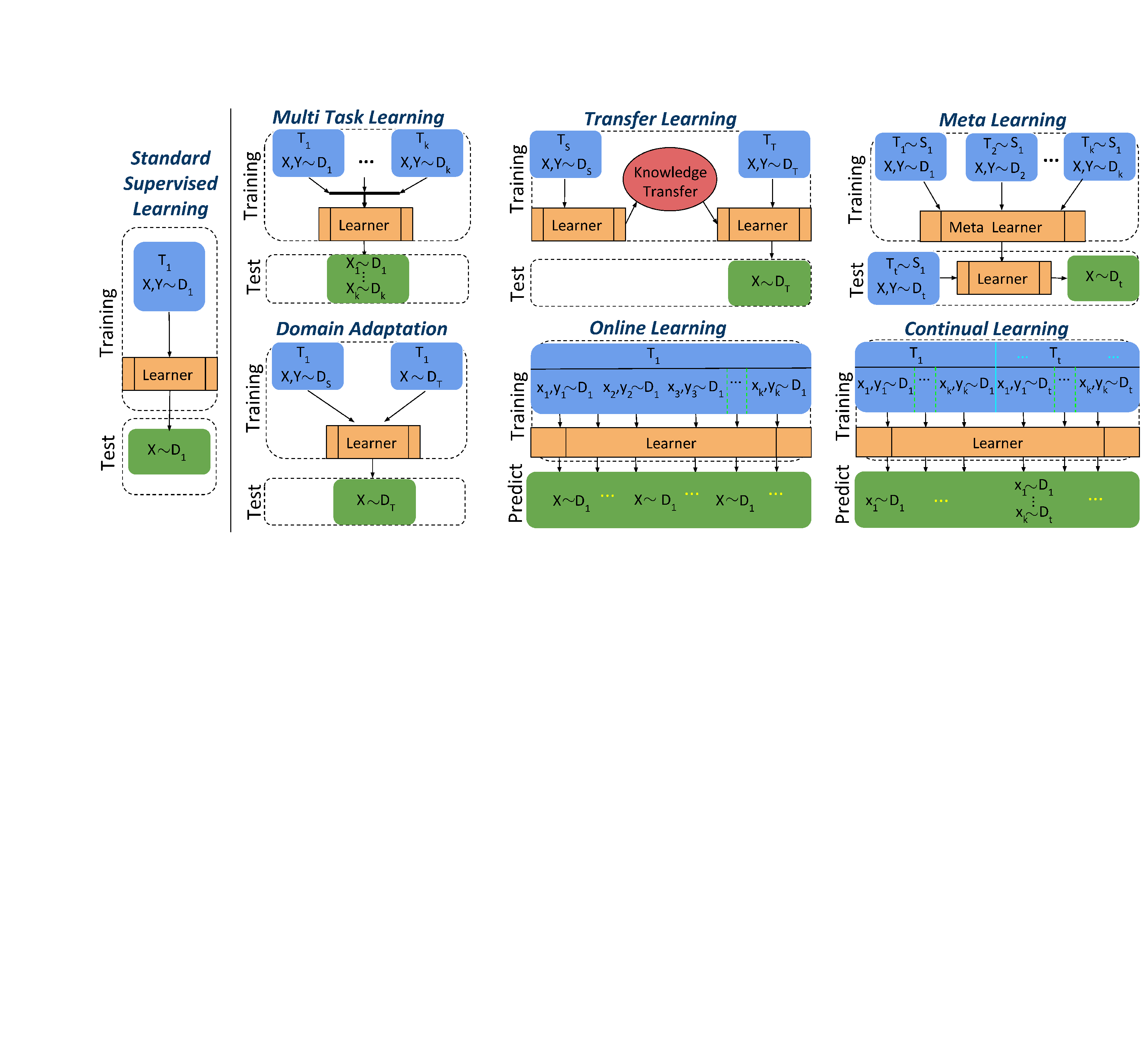}} 
    \caption{\label{fig:transfer_learning} Main setup of related machine learning fields, illustrating the differences with general continual learning settings.}
  \end{figure*}

The ideas depicted in the continual learning desiderata of knowledge sharing, adaptation and transfer have been studied previously in machine learning with development in related fields. We will describe each of them briefly and highlight the main differences with continual learning (see also Figure~\ref{fig:transfer_learning}).\\
\myparagraph{Multi-Task Learning} (MTL) considers learning multiple related tasks simultaneously using a set or subset of shared parameters. It aims for a better generalization and a reduced overfitting using shared knowledge extracted from the related tasks. We refer to~\cite{zhang2017survey} for a survey on the topic.
As in our experiments, an MTL baseline is represented by a \emph{joint} learning of all tasks simultaneously. There is no adaptation after the model has been deployed, as opposed to continual learning.  \\
\myparagraph{Transfer Learning}
 aims to aid the learning process of a given task (the target) by exploiting knowledge acquired from another task or domain (the source). %
Transfer learning is mainly concerned with the forward transfer desiderata of continual learning. However, it  does not involve any continuous adaptation after learning the target task. Moreover, performance on the source task(s) is not taken into account during transfer learning. We refer to \cite{pan2009survey} for a survey.\\
\myparagraph{Domain Adaptation} is a subfield of transfer learning  with the same source and target tasks, but drawn from different input domains. The goal is to adapt a model trained on the source domain to perform well on the target domain, which has only unlabelled data or has only few labels). In other words, it relaxes the classical machine learning assumption of having training and test data drawn from the same distribution~\cite{csurka2017domain}. As mentioned above for transfer learning, domain adaptation is unidirectional and does not involve any accumulation of knowledge~\cite{chen2018lifelong}.\\
\myparagraph{Learning to Learn (Meta Learning).}
The old definition of learning to learn was referring to the concept of improving the learning behaviour of a model with training experience. %
However, more recently, the common interpretation is the ability for a faster adaptation on a task with few examples given a large number of training tasks.
While these ideas seem quite close to continual learning, meta learning still follows the same  offline training assumption, but with data randomly drawn from a task training distribution, and test data being tasks with few examples. Hence, it is not  capable, alone, of preventing forgetting on those previous tasks.\\ %
\myparagraph{Online Learning.}
In traditional offline learning, the entire training data has to be made available prior to learning the task. On the contrary, online learning studies algorithms that learn to optimize predictive models over a stream of data instances sequentially. We refer to~\cite{shalev2012online,bottou1998online}
for surveys on the topic. 
Note that online learning assumes an i.i.d data sampling procedure and considers a single task domain, which sets it apart from continual learning.  \\
\myparagraph{Open World Learning}~\cite{DBLP:journals/corr/abs-1809-06004,bendale2015towards} deals with the problem of detecting new classes at test time, hence avoiding wrong assignments to known classes. When those new classes are then integrated in the model, it meets the problem of incremental learning.
As such, open world learning can be seen as a subtask of continual learning.

\section{Conclusion}
\label{sec:conclusion}%
In this work, we scrutinized recent state-of-the-art continual learning methods, confining the study to task incremental classification with a multi-head setup, as mainly applied in literature.
Within these outlines we proposed a constructive taxonomy and
tested each of the compared methods in an attempt to answer several critical questions in the field. 
In order to fulfill the urge for a fair comparison compliant with the continual learning paradigm, we proposed a novel continual hyperparameter framework which dynamically defines forgetting related hyperparameters, i.e. the stability-plasticity trade-off is determined in a continual fashion with only the current task data available.
The overview of our experimental findings in Table~\ref{tab:summary} shed an empirical light on method strengths and weaknesses, supplemented by recommendations for model capacity and the use of dropout or weight decay.
The experiments extend to three datasets, namely Tiny Imagenet, iNaturalist and RecogSeq, from which the latter two resemble real-world data sequences to challenge methods with an unbalanced data distribution and highly varying tasks. On top, we addressed the influence of task ordering and found it minimally influential towards the general trends of the compared continual learning methods.

Although the state-of-the-art made some significant progress to tackle catastrophic forgetting in neural networks, the majority of results originates from a highly confined setup, leaving numerous opportunities for further research to reach beyond classification, multi-head evaluation and a task incremental setup.

\ifCLASSOPTIONcompsoc
  \section*{Acknowledgments}
\else
  \section*{Acknowledgment}
\fi

The authors would like to thank Huawei for funding this research. Marc Masana acknowledges 2019-FI\_B2-00189 grant from Generalitat de Catalunya. Rahaf Aljundi is funded by an FWO scholarship

\ifCLASSOPTIONcaptionsoff
  \newpage
\fi


\begin{IEEEbiographynophoto}{Matthias~De Lange}
is a Ph.D. researcher at the Center for Processing Speech and Images of KU Leuven.
He received the M.Sc. degree and laureate from KU Leuven in Belgium 2018. His research interests encompass continual learning, computer vision and user adaptation.
\end{IEEEbiographynophoto}

\begin{IEEEbiographynophoto}
{Rahaf Aljundi} is a machine learning and vision researcher at Toyota Motor Europe.   She has a Ph.D. degree from KU Leuven, Belgium 2019 and joint master degree from Jean Monnet University, France and University of Alicante, Spain.  Her main research interests are continual learning, transfer learning and active learning.
\end{IEEEbiographynophoto}

\begin{IEEEbiographynophoto}{Marc Masana} 
 is a mathematician and computer science engineer working as a researcher at the Computer Vision Center. He obtained his Ph.D.\ 
 from Universitat Autònoma de Barcelona. His research interests concern machine learning, network compression and continual learning.
\end{IEEEbiographynophoto}

\begin{IEEEbiographynophoto}{Sarah Parisot} is Senior Research Scientist for Huawei Noah's Ark Lab.  She has a Ph.D. degree from Ecole Centrale Paris/INRIA and has held positions in both academia and industry.  Her research interests include few-shot/unsupervised learning, computer vision and graphs.
 
\end{IEEEbiographynophoto}

\begin{IEEEbiographynophoto}{Xu Jia} is senior researcher at Huawei Noah's Ark Lab. He received M.Sc. degree from Dalian University of Technology, China in 2013 and a Ph.D. degree from KU Leuven, Belgium in 2018. His research interests concern image processing, computer vision and machine learning.
\end{IEEEbiographynophoto}

\begin{IEEEbiographynophoto}{Ale\v{s}~Leonardis} is a Senior Research Scientist and Computer Vision Team Leader at Huawei Technologies Noah’s Ark Lab (London, UK). He is also Chair of Robotics at the School of Computer Science, University of Birmingham and Professor of Computer and Information Science at the University of Ljubljana. 
\end{IEEEbiographynophoto}

\begin{IEEEbiographynophoto}{Gregory Slabaugh} is Chief Scientist in Computer Vision (Europe) for Huawei Noah's Ark Lab.  He received a Ph.D. degree from Georgia Institute of Technology and has worked in academia and industry.  His research interests are computational photography and computer vision.
\end{IEEEbiographynophoto}

\begin{IEEEbiographynophoto}{Tinne Tuytelaars}
is a full professor at the Center for Processing Speech and Images of KU Leuven.
Her research interests encompass continual learning, robust image and video representations and multi-modal analysis.
\end{IEEEbiographynophoto}

\vfill

\cleardoublepage

\appendices
\section{Implementation Details}
\label{apdx:implementation-details}
\subsection{Model Setup}
\textbf{Tiny Imagenet} VGG-based model details. 

\noindent
Classifiers with number of fully connected units:
\begin{itemize}
    \item C1: [128, 128] + multi-head 
    \item C2: [512, 512] + multi-head 
\end{itemize}

\noindent
Full models (filter count, M = max pooling):
\begin{itemize}
    \item \textsc{small}: [64, M, 64, M, 64, 64, M, 128, 128, M] + C1
    \item \textsc{base}: [64, M, 64, M, 128, 128, M, 256, 256, M] + C2
    \item \textsc{wide}: [64, M, 128, M, 256, 256, M, 512, 512, M] + C2
    \item \textsc{deep}: [64, M, (64,)*6, M, (128,)*6, M, (256,)*6, M] + C2
\end{itemize}
\matt{
Dropout is applied on last and penultimate classifier layers for the regularization experiments.
}

\subsection{Framework Setup}
Tiny Imagenet, iNaturalist \matt{and RecogSeq} define maximal plasticity search with coarse learning rate grid $\Psi = \left\{ 1e^{-2}, 5e^{-3}, 1e^{-3}, 5e^{-4}, 1e^{-4} \right\}$. 
Tiny Imagenet starts from scratch and therefore applies 2 additional stronger learning rates $\Psi \cup \left\{ 1e^{-1}, 5e^{-2} \right\}$ for the first tasks.

We set the finetuning accuracy drop margin to $p = 0.2$ and decaying factor $\alpha = 0.5$, unless mentioned otherwise.
In case of multiple hyperparameters, multiple decay strategies are possible.
In our theoretical framework we propose scalar multiplication with all hyperparameters in the set.
In experiments we conduct a more careful approach, which first tries to decay each hyperparameter separately, and decays all if none of these individual decays achieved an accuracy within the finetuning margin. This process is then repeated until the stability decay criterion is met.

\subsection{Methods Setup}
\textbf{General setup} for all methods:
\begin{itemize}
    \item Pytorch implementation.
    \item Stochastic Gradient Descent with a momentum of $0.9$, and batch size 200.
    \item Max $70$ training epochs with early stopping and annealing of the learning rate: after 5 unimproved iterations of the validation accuracy the learning rate decays with factor $10$, after 10 unimproved iterations training terminates.
    \item The baselines and parameter-isolation methods start from scratch, other methods continue from the same model trained for the first task.
    \item All softmax temperatures for knowledge distillation are set to $2$.
\end{itemize}{}
We observed in our experiments that the initial hyperparameter values of the methods consistently decay to a certain order of magnitude. To avoid overhead, we start all methods with the observed upper bound, with an additional inverse decay step as margin.

\noindent
\boldspacepar{Replay setup.}
\noindent
GEM doesn't specify a memory management policy and merely divides an equal portion of memory to each of the tasks. The initial value for the forgetting-related hyperparameter is set to $1$.

In our task incremental setup, iCARL fully exploits the total available exemplar memory $M$, and incrementally divides the capacity equally over all the seen tasks. Similar to LwF, the initial knowledge distillation strength is $10$.

\noindent
\boldspacepar{Regularization-based setup.}
\noindent
Initial regularization strength based on consistently observed decays in the framework: EWC $400$, SI $400$, MAS $3$.
LwF and EBLL both start with the knowledge distillation strengths set to $10$, EBLL loss w.r.t. the code is initialized with $1$.
EBLL hyperparameters for the autoencoder are determined in a gridsearch with code dimension $\{100, 300\}$ for Tiny Imagenet and RecogSeq, and  $\{200, 500 \}$ for iNaturalist, all with reconstruction regularization strength in $\{0.01, 0.001\}$,  with learning rate $0.01$, and for $50$ epochs. RecogSeq required an additional learning rate $0.1$ in the gridsearch to enable autoencoder learning for the highly dissimilar tasks.

For IMM on Tiny Imagenet dataset we observed no clear influence of the L2 transfer regularization strength (for values $10e^{-2}$, $10e^{-3}$, $10e^{-4}$), and therefore executed all experiments with a fixed value of $10e^{-2}$ instead. In order to merge the models, all task models are equally weighted, as there are no indications in how the values should be determined in \cite{Lee2017}.

\boldspacepar{Parameter isolation setup.}
\noindent
For PackNet the initial value of the forgetting-related hyperparameter amounts to $90\%$ pruning per layer, decayed with factor $0.9$ if not satisfying finetuning threshold.

\matt{
HAT starts with sparsity regularization hyperparameter $c=2.5$ and $s_{max}=800$. This is the upper bound of stability in the hyperparameter range of good operation suggested by the authors \cite{Serra2018}.
Tiny Imagenet experiments required 10 warmup epochs for the first task starting from scratch, with $c=0$.
Afterwards the sparsity is enforced by employing the aforementioned $c$ values.
To enable training, the embedding initialization $\mathcal{U}(0,2)$  used in HAT compression experiments \cite{Serra2018} had to be adopted.
}

\subsection{Qualitative Comparison Setup}
Timing experiments for the heatmap are based on our experimental setup for Tiny Imagenet with the \textsc{base} model configured with batch size 200, all phases with multiple iterations have their epochs set to 10, gridsearches are confined to one node, and replay-buffers have 450 exemplars allocated per task.

\section{Supplemental Results}
\subsection{Complete Mean-IMM Results}
Table~\ref{apdx:tab:exp:tiny:rnd} compares the mean-IMM results with mode-IMM, and is consistently outperformed on Tiny Imagenet for all orderings. This is not surprising as mean-IMM merely averages models optimized for different task, while mode-IMM keeps track of per-parameter importance for each of the tasks. 

The influence of weight decay and dropout is also measured for mean-IMM in Table~\ref{apdx:tab:exp:regul:regulbased}. Salient is dropout increasing forgetting (blue box).

Finally, Table~\ref{apdx:tab:inat:meanimm} shows a comparison of both IMM merging methods on iNaturalist with the pretrained Alexnet, in all three orderings.
Both the iNaturalist and RecogSeq results are reported in the manuscript.

\begin{table}[]
\centering
\caption{\textbf{Mean-IMM results on Tiny Imagenet} for different model sizes with {\em random ordering} of tasks (top),
using the \emph{easy to hard} (middle) and \emph{hard to easy} (bottom). We report average accuracy (average forgetting), indicating mode-IMM to consistently surpass mean-IMM.}
\label{apdx:tab:exp:tiny:rnd}

\resizebox{\linewidth}{!}{%
\begin{tabular}{@{}lllll@{}}
\toprule
\textbf{Model} & \textbf{finetuning}                                         & \textbf{joint*}                        & \textbf{mean-IMM}       & \textbf{mode-IMM}                           \\ \midrule
\textsc{small} &16.25 (34.84)                   & 57.00 (\NA)            & 19.02 (27.03) & 29.63 (3.06)                                \\
\textsc{base}  & 21.30 (26.90)                                               & 55.70 (\NA)                               & 26.38 (22.43)           & 36.89 (0.98)                                \\
\textsc{wide}  & 25.28 (24.15)                                               & 57.29 (\NA)                               & 23.31 (26.50)           & 36.42 (1.66)                                \\
\textsc{deep}  & 20.82 (20.60) & 51.04 (\NA)  & 21.28 (18.25)           & 27.51 (0.47) \\ \bottomrule
\end{tabular}%
}

\rule{0pt}{1ex}  

\resizebox{\linewidth}{!}{%
\begin{tabular}{@{}lllllllllll@{}}
\toprule
\textbf{Model} & \textbf{finetuning} & \textbf{joint*} & \textbf{mean-IMM} & \textbf{mode-IMM} \\ \midrule
\textsc{small} & 16.06 (35.40)       & 57.00 (\NA)      & 14.62 (36.77)     & 26.13 (3.03)      \\
\textsc{base}  & 23.26 (24.85)       & 55.70 (\NA)      & 22.84 (24.33)     & 36.81 (-1.17)     \\
\textsc{wide}  & 19.67 (29.95)       & 57.29 (\NA)      & 25.42 (23.90)     & 38.68 (-1.09)     \\
\textsc{deep}  & 23.90 (17.91)       & 51.04 (\NA)      & 17.95 (22.38)     & 25.89 (-2.09)     \\ \bottomrule
\end{tabular}%
}

\rule{0pt}{1ex}  

\resizebox{\linewidth}{!}{%
\begin{tabular}{@{}lllllllllll@{}}
\toprule
\textbf{Model} & \textbf{finetuning} & \textbf{joint*} & \textbf{mean-IMM} & \textbf{mode-IMM} \\ \midrule
\textsc{small} & 18.62 (28.68)       & 57.00 (\NA)      & 17.44 (31.05)     & 24.95 (1.66)      \\
\textsc{base}  & 21.17 (22.73)       & 55.70 (\NA)       & 23.64 (20.17)     & 34.58 (0.23)      \\
\textsc{wide}  & 25.25 (22.69)       & 57.29 (\NA)       & 23.70 (22.15)     & 35.24 (-0.82)     \\
\textsc{deep}  & 15.33 (20.66)       & 51.04 (\NA)       & 16.20 (21.45)     & 26.38 (1.20)      \\ \bottomrule
\end{tabular}%
}

\end{table}

\begin{table}[]
\centering
\caption{\textbf{Mean-IMM}: \textbf{effects of dropout} ($p=0.5$) and \textbf{weight decay} (${\lambda=10^{-4}}$) regularization for the 4 model configurations on Tiny Imagenet. Red indicates deterioration in average accuracy w.r.t. no regularization.}
\label{apdx:tab:exp:regul:regulbased}

\resizebox{\linewidth}{!}{%
\begin{tabular}{@{}llllllllllllll@{}}
\toprule
\textbf{Model}      & \textit{}                  & \textbf{finetuning} & \textbf{joint*} & \textbf{mean-IMM} & \textbf{mode-IMM} \\ \midrule
\textsc{small}  & \textit{Plain} & 16.25 (34.84)       & 57.00 (n/a)   & 19.02 (27.03)     & 29.63 (3.06)         \\
\textbf{}      & \textit{Dropout}           & 19.52 (32.62)       & \redline{55.97} (n/a)    & 21.71 \marktopleftTIGHT{b3}(31.30)\markbottomrightTIGHT{b3}{blue}     & \redline{29.35} (0.90)         \\
\textbf{}      & \textit{Weight Decay}      & \redline{15.06} (34.61)       & \redline{56.96} (n/a)   & 22.33 (24.19)     & 30.59 (0.93)         \\ \addlinespace
\textsc{base}   & \textit{Plain} & 21.30 (26.90)       & 55.70 (n/a)  & 26.38 (22.43)     & 36.89 (0.98)         \\
\textbf{}      & \textit{Dropout}           & 29.23 (26.44)       & 61.43 (n/a)   & \redline{21.11} \marktopleftTIGHT{b4}(24.41)\markbottomrightTIGHT{b4}{blue}     & \redline{34.20} (1.44)         \\
\textbf{}      & \textit{Weight Decay}      & \redline{19.14} (29.31)       & 57.12 (n/a)   & \redline{25.85} (21.92)     & 37.49 (0.46)         \\ \addlinespace
\textsc{wide}  & \textit{Plain} & 25.28 (24.15)       & 57.29 (n/a)  & 23.31 (26.50)     & 36.42 (1.66)         \\
\textbf{}      & \textit{Dropout}           & 30.76 (26.11)       & 62.27 (n/a)  & 32.42 (22.03)     & 42.41 (-0.93)        \\
\textbf{}      & \textit{Weight Decay}      & \redline{22.78} (27.19)       & 59.62 (n/a)  & 28.33 (23.01)     & 39.16 (1.41)          \\ \addlinespace
\textsc{deep}  & \textit{Plain} & 20.82 (20.60)       & 51.04 (n/a) & 21.28 (18.25)     & 27.51 (0.47)         \\
\textbf{}      & \textit{Dropout}           & 23.05 (27.30)       & 59.58 (n/a)   & 26.38 \marktopleftTIGHT{b5}(23.38)\markbottomrightTIGHT{b5}{blue}     & 33.64 (-0.61         \\
\textbf{}      & \textit{Weight Decay}      & \redline{19.48} (21.27)       & 54.60 (n/a) & \redline{18.93} (18.64)     & \redline{25.62} (-0.15)  \\ \bottomrule

\end{tabular}%
}
\end{table}

\begin{table}[]
\centering
\caption{\textbf{Mean-IMM results on iNaturalist} with pretrained Alexnet, compared to mode-IMM and the baselines.}
\label{apdx:tab:inat:meanimm}
\resizebox{\linewidth}{!}{%
\begin{tabular}{@{}lllll@{}}
\toprule
\textbf{Ordering} & \textbf{finetuning} & \textbf{joint*} & \textbf{mean-IMM} & \textbf{mode-IMM} \\ \midrule
\emph{random }           & 45.59 (17.02)       & 63.90 (n/a)     & 49.82 (13.42)     & 55.61 (-0.29)     \\
\emph{related    }               & 44.12 (16.27)       & 63.90 (n/a)     & 47.21 (12.34)     & 52.35 (-1.32)     \\
\emph{unrelated   }                & 48.27 (11.46)       & 63.90 (n/a)     & 48.58 (10.83)     & 52.71 (-0.81)     \\ \bottomrule
\end{tabular}%
}
\end{table}

\subsection{Synaptic Intelligence (SI): Overfitting and Regularization}
\label{apdx:sup:SI-overfitting}
The observed overfitting for SI on Tiny Imagenet is illustrated in Table~\ref{tab:apdx:SI:overfit} for the \textsc{base} model.
Training accuracy without regularization tends to reach $100\%$ for all tasks, with the validation phase merely attaining about half of this accuracy. This results in an average discrepancy between training and validation accuracy of $48.8\%$, indicate significant overfitting on the training data. However, this discrepancy can be greatly reduced by the use of regularization, such as dropout ($20.6\%$) or weight decay ($30.8\%$).

Without regularization the overfitting results of SI are very similar to the finetuning baseline ($48.8\%$ and $48.0\%$ discrepancy). However, finetuning does not clearly benefit from regularization, retaining training accuracies near $100\%$. Observing different weight decay strengths in Table~\ref{tab:apdx:FT:regul-strengths}, severely increasing the standard weight decay strength in our experiments ($\lambda=0.0001$) by up to a factor $100$ merely  reduces average training accuracy to $85.5\%$, leaving a significant discrepancy with the validation accuracy of $37.8\%$. 

Continual learning methods EWC and LwF in the regularization-based family of SI, inherently reduce overfitting (resp. $38.4\%$ and $28.9\%$ discrepancy) without any use of the traditional weight decay and dropout regularizers.

\begin{table*}[]
\centering
\caption{We scrutinize the \textbf{level of overfitting for SI} compared to Finetuning, EWC and LwF.
Training (\textit{train}), validation (\textit{val}) accuracies and discrepancy (\textit{train}$-$ \textit{val}) are reported for the \textsc{Base} model on randomly ordered Tiny Imagenet. Accuracies are determined on data solely from the indicated current task. 
}
\label{tab:apdx:SI:overfit}
\begin{tabular}{@{}lllllllllllll|l@{}}
\toprule
                  &                            & \textbf{Accuracy}    & \textbf{T1} & \textbf{T2} & \textbf{T3} & \textbf{T4} & \textbf{T5} & \textbf{T6} & \textbf{T7} & \textbf{T8} & \textbf{T9} & \textbf{T10} & \textbf{Avg}  \\ \midrule
\textbf{SI}       & \textbf{No Regularization} & \textit{train}       & 97.3        & 99.9        & 99.6        & 90.0        & 99.9        & 98.8        & 99.7        & 99.9        & 100       & 98.4         & 98.4          \\
\textbf{}         & \textbf{}                  & \textit{val}         & 53.3        & 48.9        & 51.8        & 51.2        & 46.5        & 51.2        & 45.0        & 50.7        & 48.4        & 48.6         & 49.5          \\
\textbf{}         & \textbf{}                  & \textit{discrepancy} & 44.1        & 51.0        & 47.9        & 38.8        & 53.4        & 47.6        & 54.7        & 49.2        & 51.6        & 49.8         & \textbf{48.8} \\ \addlinespace
\textbf{}         & \textbf{Dropout}           & \textit{train}       & 97.9        & 85.0        & 65.6        & 70.6        & 71.3        & 70.0        & 68.3        & 76.1        & 71.6        & 72.6         & 74.9          \\
\textbf{}         & \textbf{}                  & \textit{val}         & 59.6        & 54.2        & 54.7        & 56.4        & 50.0        & 55.7        & 49.3        & 56.2        & 55.1        & 52.6         & 54.4          \\
\textbf{}         & \textbf{}                  & \textit{discrepancy} & 38.3        & 30.9        & 10.9        & 14.2        & 21.3        & 14.4        & 19.0        & 20.0        & 16.5        & 20.1         & \textbf{20.6} \\ \addlinespace
\textbf{}         & \textbf{L2}                & \textit{train}       & 99.9        & 97.2        & 75.3        & 75.2        & 73.0        & 76.3        & 68.1        & 67.5        & 71.7        & 82.0         & 78.6          \\
\textbf{}         & \textbf{}                  & \textit{val}         & 51.3        & 46.3        & 49.9        & 49.0        & 44.2        & 50.1        & 43.4        & 48.5        & 48.1        & 47.6         & 47.8          \\
\textbf{}         & \textbf{}                  & \textit{discrepancy} & 48.6        & 50.9        & 25.5        & 26.2        & 28.8        & 26.2        & 24.7        & 19.0        & 23.6        & 34.4         & \textbf{30.8} \\ \addlinespace
\midrule
\textbf{Finetuning} & \textbf{No Regularization} & \textit{train}       & 99.8        & 100       & 92.1        & 100       & 99.6        & 100       & 100       & 87.9        & 100       & 85.8         & 96.5          \\
\textbf{}         & \textbf{}                  & \textit{val}         & 53.5        & 47.4        & 49.1        & 51.8        & 44.0        & 51.4        & 42.9        & 49.3        & 47.9        & 47.9         & 48.5          \\
\textbf{}         & \textbf{}                  & \textit{discrepancy} & 46.3        & 52.6        & 43.0        & 48.2        & 55.6        & 48.6        & 57.1        & 38.6        & 52.1        & 37.9         & \textbf{48.0} \\ \addlinespace
\textbf{}         & \textbf{Dropout}           & \textit{train}       & 94.2        & 96.2        & 98.3        & 98.0        & 97.7        & 97.9        & 94.9        & 98.8        & 98.2        & 93.4         & 96.7          \\
\textbf{}         & \textbf{}                  & \textit{val}         & 58.4        & 57.2        & 60.2        & 59.6        & 53.4        & 58.0        & 51.9        & 59.4        & 57.1        & 55.9         & 57.1          \\
\textbf{}         & \textbf{}                  & \textit{discrepancy} & 35.8        & 39.0        & 38.1        & 38.4        & 44.3        & 40.0        & 43.0        & 39.4        & 41.1        & 37.6         & 39.7          \\ \addlinespace
\textbf{}         & \textbf{L2}                & \textit{train}       & 99.8        & 99.9        & 99.9        & 100       & 100       & 100       & 100       & 72.7        & 99.8        & 100        & 97.2          \\
\textbf{}         & \textbf{}                  & \textit{val}         & 52.8        & 49.8        & 53.9        & 51.4        & 45.2        & 50.1        & 45.2        & 48.1        & 47.9        & 46.2         & 49.0          \\
\textbf{}         & \textbf{}                  & \textit{discrepancy} & 47.0        & 50.1        & 46.0        & 48.6        & 54.9        & 49.9        & 54.9        & 24.6        & 52.0        & 53.8         & 48.2          \\ \addlinespace
\midrule
\textbf{EWC}      & \textbf{No Regularization} & \textit{train}       & 97.3        & 99.8        & 95.8        & 69.4        & 97.4        & 92.7        & 78.7        & 85.8        & 84.7        & 84.4         & \textbf{88.6}          \\
\textbf{}         & \textbf{}                  & \textit{val}         & 53.3        & 48.4        & 52.7        & 51.7        & 46.3        & 52.5        & 45.3        & 53.1        & 49.8        & 49.5         & 50.2          \\
\textbf{}         & \textbf{}                  & \textit{discrepancy} & 44.1        & 51.4        & 43.1        & 17.7        & 51.1        & 40.2        & 33.5        & 32.7        & 34.9        & 34.9         & 38.4          \\ \addlinespace
\textbf{}         & \textbf{Dropout}           & \textit{train}       & 97.9        & 94.8        & 96.8        & 94.8        & 87.7        & 84.4        & 77.6        & 83.0        & 76.4        & 87.3         & 88.1          \\
\textbf{}         & \textbf{}                  & \textit{val}         & 59.6        & 56.6        & 57.9        & 58.0        & 50.4        & 55.4        & 49.3        & 55.7        & 55.3        & 54.1         & 55.2          \\
\textbf{}         & \textbf{}                  & \textit{discrepancy} & 38.3        & 38.2        & 39.0        & 36.8        & 37.3        & 29.0        & 28.3        & 27.3        & 21.1        & 33.2         & 32.9          \\ \addlinespace
\textbf{}         & \textbf{L2}                & \textit{train}       & 99.9        & 100       & 97.8        & 94.6        & 71.9        & 93.8        & 68.8        & 82.9        & 66.0        & 74.4         & 85.0          \\
\textbf{}         & \textbf{}                  & \textit{val}         & 51.3        & 48.9        & 50.3        & 49.5        & 44.8        & 48.7        & 41.1        & 50.0        & 48.1        & 47.4         & 48.0          \\
\textbf{}         & \textbf{}                  & \textit{discrepancy} & 48.6        & 51.1        & 47.5        & 45.1        & 27.2        & 45.1        & 27.7        & 32.9        & 17.9        & 27.1         & 37.0          \\ \addlinespace
\midrule
\textbf{LwF}      & \textbf{No Regularization} & \textit{train}       & 99.8        & 100       & 67.4        & 100       & 50.3        & 99.6        & 63.5        & 58.9        & 60.4        & 70.8         & \textbf{77.1}          \\
\textbf{}         & \textbf{}                  & \textit{val}         & 53.5        & 50.8        & 53.2        & 51.7        & 42.9        & 47.8        & 40.6        & 48.1        & 47.3        & 46.5         & 48.2          \\
\textbf{}         & \textbf{}                  & \textit{discrepancy} & 46.3        & 49.3        & 14.3        & 48.3        & 7.4         & 51.8        & 22.9        & 10.9        & 13.1        & 24.4         & 28.9          \\ \addlinespace
\textbf{}         & \textbf{Dropout}           & \textit{train}       & 94.2        & 96.7        & 96.1        & 97.4        & 58.6        & 93.0        & 88.1        & 67.9        & 89.4        & 98.7         & 88.0          \\
\textbf{}         & \textbf{}                  & \textit{val}         & 58.4        & 56.4        & 55.9        & 55.6        & 49.0        & 54.2        & 49.8        & 54.4        & 55.4        & 52.7         & 54.2          \\
\textbf{}         & \textbf{}                  & \textit{discrepancy} & 35.8        & 40.3        & 40.2        & 41.8        & 9.7         & 38.8        & 38.3        & 13.5        & 34.0        & 46.1         & 33.9          \\ \addlinespace
\textbf{}         & \textbf{L2}                & \textit{train}       & 99.8        & 100       & 64.1        & 100       & 61.6        & 99.8        & 50.8        & 69.9        & 68.4        & 62.0         & 77.6          \\
\textbf{}         & \textbf{}                  & \textit{val}         & 52.8        & 50.0        & 47.6        & 47.5        & 42.0        & 46.4        & 38.4        & 46.6        & 46.7        & 43.2         & 46.1          \\
\textbf{}         & \textbf{}                  & \textit{discrepancy} & 47.0        & 50.0        & 16.5        & 52.5        & 19.6        & 53.4        & 12.5        & 23.3        & 21.8        & 18.9         & 31.5          \\ \bottomrule
\end{tabular}%
\end{table*}

\begin{table*}[]
\centering
\caption{\textbf{Finetuning} with different \textbf{weight decay strengths ($\lambda$)}, reported for the \textsc{Base} model on randomly ordered Tiny Imagenet. $\lambda = 0.0001$ is the standard setup for our experiments.}
\label{tab:apdx:FT:regul-strengths}
\begin{tabular}{@{}llllllllllll|l@{}}
\toprule
\textbf{$\lambda$}     & \textbf{Accuracy}    & \textbf{T1} & \textbf{T2} & \textbf{T3} & \textbf{T4} & \textbf{T5} & \textbf{T6} & \textbf{T7} & \textbf{T8} & \textbf{T9} & \textbf{T10} & \textbf{Avg} \\ \midrule
\textbf{0.0001} & \textit{train}       & 99.8 & 99.9  & 99.9  & 100 & 100 & 100 & 100 & 72.7 & 99.8  & 100 & 97.2 \\
\textbf{}       & \textit{val}         & 52.8 & 49.8  & 53.9  & 51.4  & 45.2  & 50.1  & 45.2  & 48.1 & 47.9  & 46.2  & 49.0 \\ 
\textbf{}       & \textit{discrepancy} & 47.0 & 50.1  & 46.0  & 48.6  & 54.9  & 49.9  & 54.9  & 24.6 & 52.0  & 53.8  & 48.2 \\ \addlinespace
\textbf{0.001}  & \textit{train}       & 99.6 & 100 & 100 & 100 & 100 & 100 & 50.0  & 99.7 & 100 & 100 & 94.9 \\
\textbf{}       & \textit{val}         & 52.9 & 51.1  & 54.3  & 53.2  & 47.5  & 50.6  & 43.0  & 51.2 & 50.9  & 49.1  & 50.4 \\
\textbf{}       & \textit{discrepancy} & 46.7 & 48.9  & 45.7  & 46.8  & 52.5  & 49.5  & 7.0   & 48.5 & 49.1  & 51.0  & 44.5 \\ \addlinespace
\textbf{0.01}   & \textit{train}       & 69.2 & 93.6  & 96.2  & 99.7  & 52.9  & 75.0  & 74.7  & 93.5 & 100 & 100 & 85.5 \\
\textbf{}       & \textit{val}         & 49.4 & 47.4  & 51.2  & 48.4  & 43.3  & 49.1  & 43.6  & 50.0 & 48.8  & 46.4  & 47.7 \\
\textbf{}       & \textit{discrepancy} & 19.9 & 46.3  & 45.0  & 51.3  & 9.6   & 25.9  & 31.2  & 43.5 & 51.2  & 53.6  & 37.8 \\ \bottomrule
\end{tabular}%
\end{table*}

\subsection{Extra Replay Experiments: Epoch Sensitivity}
\label{apdx:GEM:epochs}
In initial experiments with 70 epochs we observed inferior performance of GEM w.r.t. iCARL and the rehearsal baselines R-PM and R-PM. 
In the original GEM setup \cite{lopez2017gradient} only a single epoch is assumed for each task, while in our experiments methods get the advantage of multiple epochs. This might impose a disadvantage when comparing GEM to the other methods, wherefore we conduct this extra experiment to attain a fair comparison (Table~\ref{tab:apdx:extra-gem-exp}). As in the other experiments in this work, we apply early stopping after 10 non-improved epochs (indicated with $*$), which makes epochs $20$ to $70$ upper bounds of the actual trained epochs.
The GEM setup with only 5 epochs shows superior average accuracies, indicating a better trade-off between the online setup GEM was designed for and exploiting several epochs to optimize for the current task. Therefore, we conduct all GEM experiments in this work with 5 epochs. 

The other replay method iCARL, on the other hand, mainly benefits from more epochs, stagnating at more than 10 epochs, with only a small increase in average forgetting.

\begin{table}%
\centering
\caption{GEM and iCARL sensitivity analysis on the amount of epochs for average accuracy (average forgetting) on the  \textsc{base} network for Tiny Imagenet. Both exemplar memories of size 4.5k and 9k are considered using the standard setup of our experiments, i.e. early stopping after 10 non-improved epochs (indicated with $*$). }
\label{tab:apdx:extra-gem-exp}
\begin{tabular}{@{}lllll@{}}
\toprule
\textbf{Epochs} & \textbf{GEM 4.5k} & \textbf{GEM 9k} & \textbf{iCARL 4.5k} & \textbf{iCARL 9k} \\ \midrule
1               & 35.05 (4.04)      & 38.23 (5.00)    & 41.18 (-1.02)       & 41.05 (-0.68)     \\
5               & 42.05 (6.40)      & 43.85 (4.42)    & 46.97 (-1.95)       & 47.89 (-2.57)     \\
10              & 38.09 (10.05)     & 43.19 (7.12)    & 46.82 (-1.78)       & 47.55 (-2.06)     \\
20*             & 40.54 (9.08)      & 34.79 (12.07)   & 47.44 (-1.23)       & 48.78 (-2.12)     \\
30*             & 38.87 (9.28)      & 29.11 (13.33)   & 47.43 (-1.88)       & 49.27 (-2.58)     \\
50*             & 32.47 (12.50)     & 38.89 (9.70)    & 47.22 (-1.43)       & 46.09 (-1.01)      \\
70*             & 25.83 (14.67)     & 40.65 (7.17)    & 47.27 (-1.11)       & 48.76 (-1.76)     \\ \bottomrule
\end{tabular}%
\end{table}

\subsection{Reproduced RecogSeq Results}
\begin{figure*}%
\center
\caption{RecogSeq dataset sequence results with low learning rates, designed specifically for this setup. Real continual learners do not have this opportunity of cherry-picking, and instead should consider a coarse wide range of learning rates to adapt to the tasks, as in our reported experiments.}
      \makebox[\textwidth][c]{\includegraphics[width=1\linewidth]{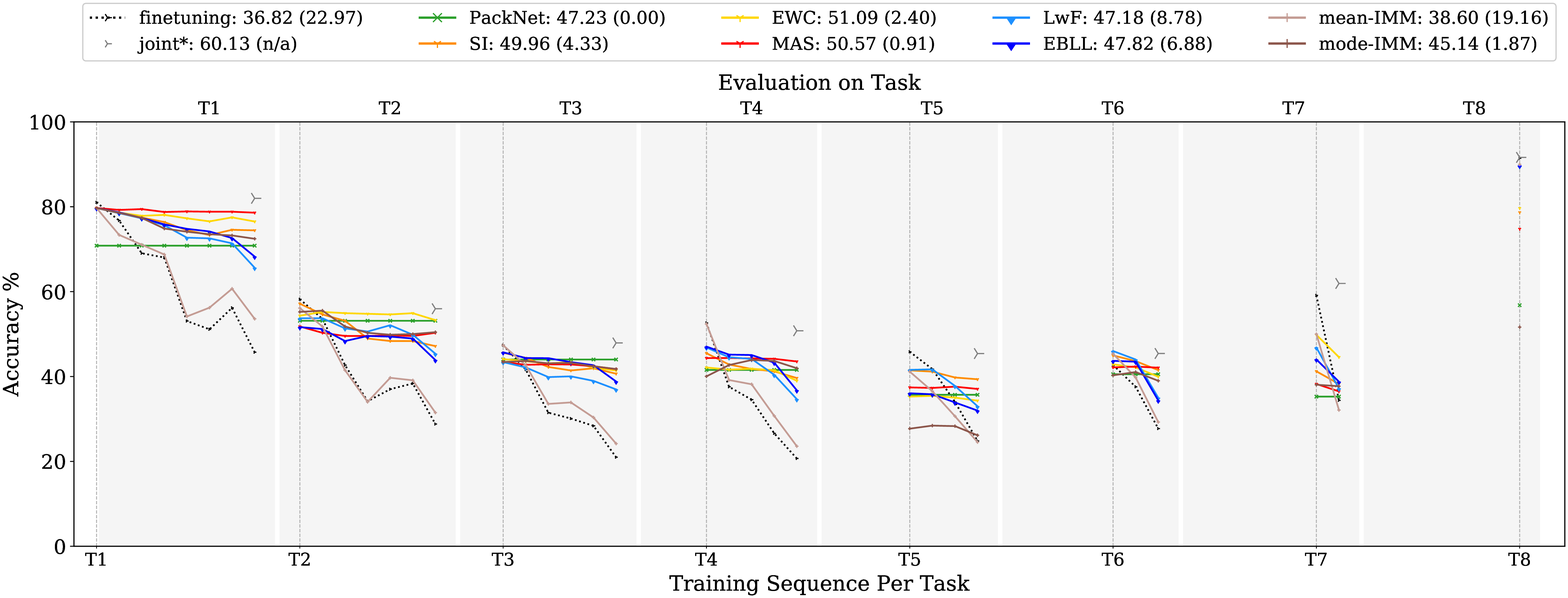}}
\label{fig:exp:recogseq:extra}
\end{figure*}

Figure~\ref{fig:exp:recogseq:extra} indicates RecogSeq results similar to the original proposed setup \cite{aljundi2017expertgate, aljundi2018memory}. In our work we consistently use the same coarse grid of learning rates ${\Psi = \left\{ 1e^{-2}, 5e^{-3}, 1e^{-3}, 5e^{-4}, 1e^{-4} \right\}}$. However, to reproduce the original results, we used softer learning rates ${\Psi = \left\{ 5e^{-5}, 1e^{-4} \right\}}$ for maximal plasticity search.
This resembles hyperparameters from the original work obtained via a gridsearch over all validation data, and hence violating the continual learning assumption.
These results show less forgetting for the regularization-based approaches, and decreased PackNet accuracy.
Firstly, this elicits significant hyperparameter sensitivity for regularization-based approaches to favor low learning rates, and PackNet thriving only with higher learning rates. 
Secondly, this illustrates the common overoptimistic results in literature, using validation data of all tasks to determine the hyperparameters. Instead, for a real-world continual learner only data of the current task is available, without previous and future task data, and additionally no knowledge of future tasks.
We address this real-world allocation of hyperparameters with our proposed continual hyperparameter framework.

\subsection{HAT Analysis: Asymmetric Capacity Allocation}
\label{apdx:HAT}

\noindent
\textbf{Results.}
We included results for HAT on \textbf{Tiny Imagenet}, with similar competitive findings as in \cite{masana2020ternary}.
The similar task distributions allow efficient reuse of previous task parameters.
We initialized the maximal hyperparameters, following the HAT ranges of good operation $\left(c=2.5, s_{max}=800\right)$ \cite{Serra2018}, decayed whenever the stability decay criterion is not met in our framework, as for all methods.

For \textbf{iNaturalist and RecogSeq} we performed numerous gridsearches, with different initializations, hyperparameters defined by the authors, and even outside the operation ranges. The main problem we observed through all the experiments was the unequal distribution of capacity in the network.

\begin{figure*}%
\center
\caption{Layer-wise HAT cumulative weight usage with sequential task learning, for the \textbf{\textsc{deep} model} on Tiny Imagenet, after training each task. The first two tasks already require over $80\%$ of the Conv0 layer capacity, leading to virtually no capacity left after training task 3. This impedes further tasks to learn anything at all.}
      \makebox[\textwidth][c]{\includegraphics[width=1\linewidth]{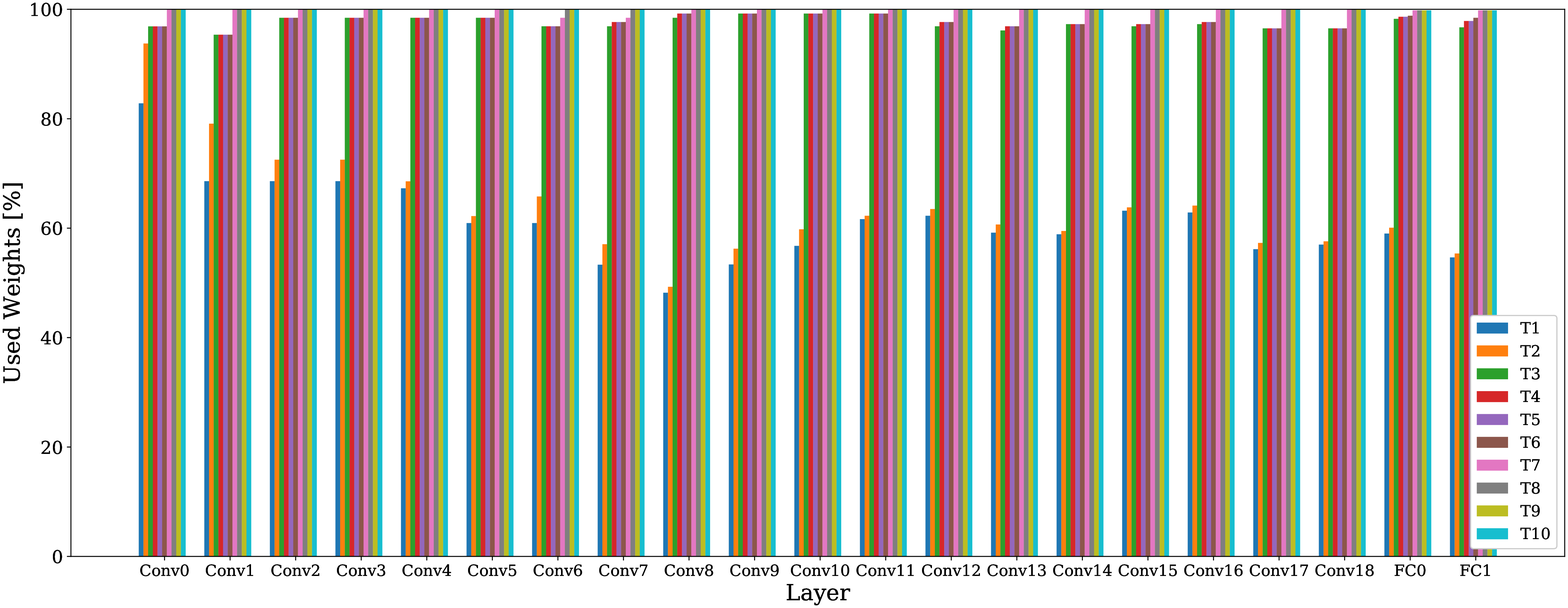}}
\label{fig:HATcompr:deep}
\end{figure*}

\begin{figure*}%
\center
\caption{Layer-wise cumulative HAT weight usage with sequential task learning, for the \textbf{\textsc{small} model} on Tiny Imagenet, after training each task. The first convolutional layer (Conv0) only saturates in capacity at the last task. All other layers retain sufficient capacity, ranging between $50\%$ and $80\%$.}
      \makebox[\textwidth][c]{\includegraphics[width=1\linewidth]{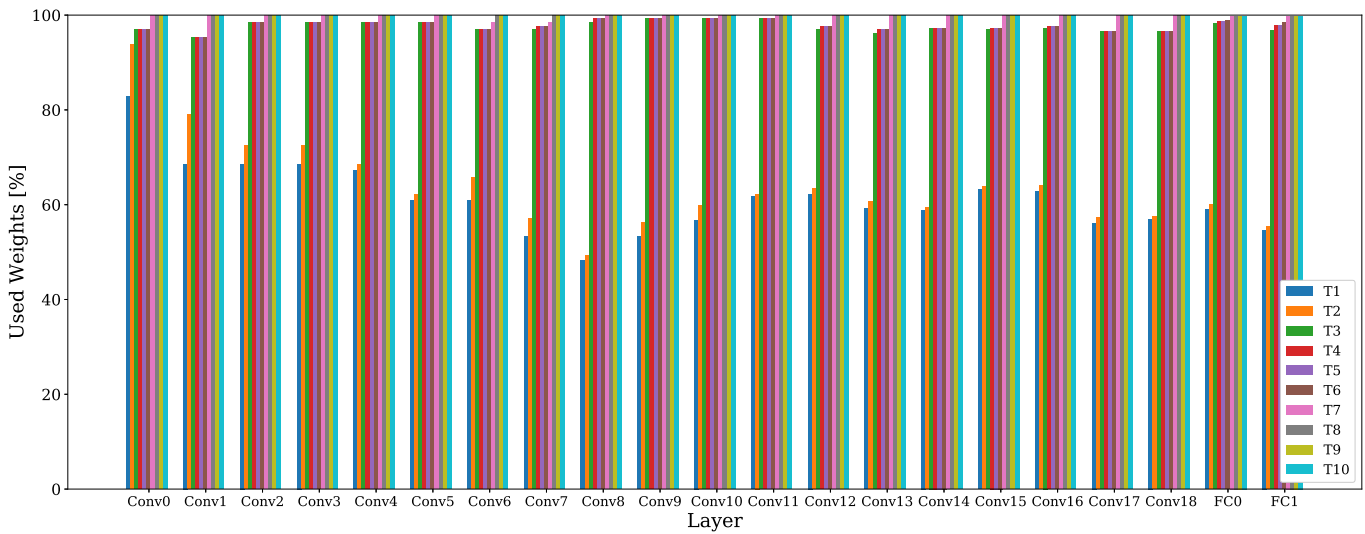}}
\label{fig:HATcompr:small}
\end{figure*}

\noindent
\textbf{Asymmetric Capacity Allocation.}
The authors of HAT conduct an experiment similar in size to our Tiny Imagenet experiments, and present the cumulative capacity used in each layer when progressing over the tasks. Their experiment indicates earlier layers to saturate first, i.e. using $100\%$ capacity, while the latter layers retain over $20\%$ free capacity. This might be the main cause for difficulties we observe in large-scale iNaturalist and RecogSeq setups, and for the \textsc{deep} model in Tiny Imagenet.
We exemplify these difficulties for the \textsc{deep} model in randomly ordered Tiny Imagenet in Figure~\ref{fig:HATcompr:deep}, which was not encountered for the other model configurations such as the \textsc{small} model in Figure~\ref{fig:HATcompr:small}. 
Further, our iNaturalist and RecogSeq setups comprise very differing and difficult large-scale tasks, requiring most of the low-level feature capacity starting from the first task. 
This issue cannot be resolved by use of the hyperparameters, as we discuss in the following.

\noindent
\textbf{Hyperparameters.}
The main implication resides in the problem that the HAT paper performs a gridsearch of all task validation sets and tries to find a optimal hyperparameter setting, where capacity is allocated in such a way that only at the end of the sequence saturation occurs. By contrast, our framework assumes that at each point only the current task validation data is available, but this makes it really hard to set the two hyperparameters for HAT to assign a specified amount of capacity to a task, with a sensitive interplay between the sparsity regularization strength and slope of the gates.
To free capacity in the network, we could decay $s_{max}$, or increase sparsity regularization strength $c$, but we observe both options to provide limited gains. 

\noindent
\textbf{Decay $s_{max}$ to free capacity.}
Even though in the near-binary mask previous task masks can be softened by altering the progress of the slope range for the Softmax gate ($s_{max}$). This $s_{max}$ has to decrease significantly, subsequently leading to catastrophic forgetting. For example, starting with really high hyperparameters minimizes the forgetting, but once capacity saturates in the early layers, lowering the hyperparameters to allow more forgetting results in catastrophic forgetting of previous tasks. This is probably due to the near-binary nature of the masks, where lowering the $s_{max}$ to reduce capacity for previous tasks, results in releasing either none or all of the masks. This would basically come down to a choice between no capacity for the new task, or catastrophically forgetting all previous knowledge.

\noindent
\textbf{Increase sparsity regularization strength $c$ to free capacity.}
To overcome the capacity problem, we also attempted some additional schemes, where for the first task we try to find the most sparse possible model by use of the hyperparameters. Nonetheless, also in this case the subsequent tasks would require all remaining early layer capacity for operation.
As shown in Table~\ref{tab:firsttask:inat-recog} for Alexnet on iNaturalist and RecogSeq setups, 
increasing the sparsity regularization strength helps to some extent, but is limited. We found high sparsity of $90\%$ for the first task such as for PackNet cannot be imposed without significant accuracy loss. 
This is in contrast with Tiny Imagenet, which does allow for higher sparsity with competitive performance in the first task model configuration as shown in Table~\ref{tab:firsttask:tiny}.
As we can avoid the capacity problem for most models on Tiny Imagenet (except the \textsc{deep} model), we include the competitive HAT results in the manuscript, whereas for iNaturalist and RecogSeq we limit experiments to this capacity analysis. 

\noindent
\textbf{Correspondence in literature.}
Besides the original results by Serra et al. \cite{Serra2018}, our results for Tiny Imagenet, and difficulties for other setups correspond to those found in other work\cite{masana2020ternary}.

\begin{table*}[]
\centering
\caption{
Tiny Imagenet random ordering with \textsc{base} model. Reporting validation accuracy on the first task, minimum and maximum capacity percentage left over all layers, and the total remaining capacity of the model. We consider both hyperparameters $c$ and $s_{max}$, respectively influencing sparsity regularization and the gate Sigmoid slope.
We perform a compression gridsearch to find initial hyperparameters on the first task with maximal capacity left, without significantly deteriorating accuracy. All model configurations can gain over $90\%$ capacity on the first task with minimal accuracy loss, which is in contrast with the iNaturalist and RecogSeq setups.
}
\label{tab:firsttask:tiny}
\begin{tabular}{@{}lllllll@{}}
\toprule
\textbf{model} & $c$ & \textbf{$s_{max}$} & \multicolumn{3}{c}{\textbf{capacity left} ($\%$)} & \textbf{acc.} \\ 
&  &  & \emph{min.} & \emph{max.} & \emph{total} &  \\\midrule

\textsc{small}& \underline{2.5}  & 800.0 & 56.2     & 97.3     & \underline{95.8}       & 49.00 \\
& 5.0  & 800.0 & 84.4     & 99.4     & 99.1       & 27.30 \\
& 7.5  & 800.0 & 87.5     & 99.7     & 99.5       & 24.15 \\
& 10.0 & 800.0 & 92.2     & 100.0    & 99.9       & 11.95 \\
& 12.5 & 800.0 & 0.0      & 3.1      & 0.6        & 11.05 \\
& 15.0 & 800.0 & 0.0      & 3.1      & 0.6        & 11.05 \\
& 17.5 & 800.0 & 0.0      & 3.1      & 0.6        & 11.05 \\
& 20.0 & 800.0 & 0.0      & 3.1      & 0.6        & 11.05 \\ \addlinespace

\textsc{base} & 2.5        & 800.0         & 17.2                       & 70.8                       & 63.8                    & 52.40        \\
& 5.0        & 800.0         & 20.3                       & 85.1                       & 80.1                    & 50.65        \\
& \underline{7.5}        & 800.0         & 39.1                       & 95.0                       & \underline{92.1}                    & 49.95        \\
& 10.0       & 800.0         & 46.9                       & 98.5                       & 96.9                    & 49.40        \\
& 12.5       & 800.0         & 51.6                       & 99.5                       & 98.1                    & 49.65        \\
& 15.0       & 800.0         & 54.7                       & 99.7                       & 98.9                    & 48.60        \\
& 17.5       & 800.0         & 65.6                       & 99.9                       & 99.4                    & 49.50        \\
& 20.0       & 800.0         & 75.0                       & 99.9                       & 99.7                    & 42.40        \\ \bottomrule
\end{tabular}%
\hspace{3em}
\begin{tabular}{@{}lllllll@{}}
\toprule
\textbf{model} & $c$ & \textbf{$s_{max}$} & \multicolumn{3}{c}{\textbf{capacity left} ($\%$)} & \textbf{acc.} \\ 
&  &  & \emph{min.} & \emph{max.} & \emph{total} &  \\\midrule
\textsc{wide}&2.5  & 800.0 & 12.5     & 62.8     & 59.9       & 51.95 \\
&5.0  & 800.0 & 23.4     & 80.5     & 78.6       & 51.70 \\
&7.5  & 800.0 & 29.7     & 87.7     & 86.8       & 50.55 \\
&\underline{10.0} & 800.0 & 32.8     & 96.0     & \underline{94.6}       & 49.60 \\
&12.5 & 800.0 & 42.2     & 96.0     & 94.8       & 50.10 \\
&15.0 & 800.0 & 53.1     & 99.1     & 98.5       & 47.80 \\
&17.5 & 800.0 & 59.4     & 99.7     & 99.2       & 47.30 \\
&20.0 & 800.0 & 62.5     & 99.9     & 99.6       & 47.70 \\ \addlinespace

\textsc{deep}&2.5  & 800.0 & 17.2     & 63.6     & 53.6       & 38.15 \\
&5.0  & 800.0 & 34.4     & 81.1     & 76.5       & 34.85 \\
&\underline{7.5}  & 800.0 & 48.4     & 91.8     & \underline{89.1}       & 35.00 \\
&10.0 & 800.0 & 65.6     & 99.7     & 99.2       & 36.15 \\
&12.5 & 800.0 & 57.8     & 94.4     & 92.4       & 34.55 \\
&15.0 & 800.0 & 65.6     & 99.7     & 99.3       & 39.60 \\
&17.5 & 800.0 & 70.3     & 97.8     & 97.2       & 36.40 \\
&20.0 & 800.0 & 81.2     & 99.9     & 99.6       & 33.40 \\ \bottomrule
\end{tabular}%
\end{table*}

\begin{table*}[]
\centering
\caption{
RecogSeq (left) and iNaturalist (right) random ordering with AlexNet. Reporting validation accuracy on the first task, minimum and maximum capacity percentage left over all layers, and the total remaining capacity of the model.
We perform a compression gridsearch to find initial hyperparameters on the first task with maximal capacity left, without significantly deteriorating validation accuracy. RecogSeq cannot attain more than $30\%$ remaining capacity. iNaturalist can attain more free capacity on the first task, but loses performance. In both cases, due to the difficult tasks, early subsequent tasks after this first task fill all remaining capacity. Hence, this scheme does not solve the capacity problem for these datasets.
}
\label{tab:firsttask:inat-recog}

\begin{tabular}{@{}llllll@{}}
\toprule
 $c$ & \textbf{$s_{max}$} & \multicolumn{3}{c}{\textbf{capacity left} ($\%$)} & \textbf{acc.} \\ 
 &  & \emph{min.} & \emph{max.} & \emph{total} &  \\\midrule
2.5        & 800.0         & 1.6               & 5.8               & 4.6                 & 39.75        \\
5.0        & 800.0         & 2.1               & 6.7               & 5.5                 & 32.66        \\
7.5        & 800.0         & 2.6               & 8.7               & 7.5                 & 42.16        \\
10.0       & 800.0         & 2.1               & 8.9               & 8.3                 & 40.27        \\
12.5       & 800.0         & 2.1               & 11.2              & 10.3                & 44.89        \\
15.0       & 800.0         & 3.1               & 11.3              & 10.7                & 40.40        \\
17.5       & 800.0         & 3.1               & 12.7              & 11.7                & 42.09        \\
20.0       & 800.0         & 4.3               & 14.6              & 12.9                & 44.63        \\ \addlinespace
22.5       & 800.0         & 5.1               & 15.6              & 14.4                & 45.44        \\
25.0       & 800.0         & 5.9               & 16.6              & 15.3                & 46.51        \\
27.5       & 800.0         & 5.9               & 16.3              & 15.1                & 42.87        \\
30.0       & 800.0         & 7.0               & 18.4              & 16.8                & 45.64        \\
32.5       & 800.0         & 6.2               & 18.0              & 16.3                & 41.67        \\
35.0       & 800.0         & 5.9               & 17.9              & 15.1                & 36.20        \\
37.5       & 800.0         & 7.4               & 19.8              & 17.9                & 40.27        \\
40.0       & 800.0         & 7.4               & 21.3              & 19.0                & 43.81        \\ \addlinespace

50.0  & 800.0 & 8.9  & 23.3 & 21.5 & 43.85 \\
60.0  & 800.0 & 9.4  & 24.0 & 22.7 & 41.96 \\
70.0  & 800.0 & 6.6  & 18.6 & 17.6 & 18.08 \\
80.0  & 800.0 & 10.9 & 26.7 & 24.8 & 39.88 \\
90.0  & 800.0 & 9.9  & 23.8 & 22.4 & 28.65 \\
100.0 & 800.0 & 10.2 & 25.1 & 23.7 & 30.61 \\ \addlinespace

110.0 & 800.0 & 10.5 & 26.3 & 24.9 & 30.38 \\
120.0 & 800.0 & 14.1 & 34.2 & 32.0 & 41.99 \\
130.0 & 800.0 & 9.9  & 24.0 & 22.3 & 16.46 \\
140.0 & 800.0 & 11.3 & 26.2 & 24.9 & 20.03 \\
150.0 & 800.0 & 12.1 & 30.1 & 27.7 & 26.96 \\
160.0 & 800.0 & 12.1 & 30.2 & 27.7 & 24.46 \\
170.0 & 800.0 & 11.7 & 27.8 & 25.8 & 19.94 \\
180.0 & 800.0 & 11.3 & 25.8 & 24.6 & 13.85 \\
190.0 & 800.0 & 12.0 & 28.6 & 26.6 & 17.82 \\
200.0 & 800.0 & 10.9 & 24.9 & 23.4 & 8.84 \\ \bottomrule

\end{tabular}%
\hspace{3em}
\begin{tabular}{@{}llllll@{}} %
\toprule
 $c$ & \textbf{$s_{max}$} & \multicolumn{3}{c}{\textbf{capacity left} ($\%$)} & \textbf{acc.} \\ 
 &  & \emph{min.} & \emph{max.} & \emph{total} &  \\\midrule
2.5        & 800.0         & 4.7               & 16.5              & 15.0                & 40.92        \\
5.0        & 800.0         & 4.7               & 21.2              & 20.5                & 40.66        \\
7.5        & 800.0         & 4.7               & 22.4              & 21.1                & 38.94        \\
10.0       & 800.0         & 4.7               & 27.8              & 27.0                & 39.47        \\
12.5       & 800.0         & 4.7               & 33.3              & 30.9                & 40.26        \\
15.0       & 800.0         & 4.7               & 42.9              & 38.9                & 41.58        \\
17.5       & 800.0         & 6.2               & 45.5              & 42.6                & 42.78        \\
20.0       & 800.0         & 4.7               & 46.3              & 42.3                & 41.32        \\ \addlinespace

22.5       & 800.0         & 4.7               & 44.8              & 41.0                & 40.13        \\
25.0       & 800.0         & 7.8               & 43.8              & 40.9                & 40.52        \\
27.5       & 800.0         & 6.2               & 49.4              & 46.6                & 40.79        \\
30.0       & 800.0         & 9.4               & 49.8              & 44.9                & 38.41        \\
32.5       & 800.0         & 9.4               & 54.5              & 51.6                & 41.85        \\
35.0       & 800.0         & 7.8               & 57.7              & 53.0                & 41.45        \\
37.5       & 800.0         & 12.5              & 50.9              & 46.1                & 36.68        \\
40.0       & 800.0         & 10.9              & 56.7              & 53.2               & 40.52        \\ \addlinespace

50.0       & 800.0         & 21.9              & 58.8              & 53.9                & 37.61        \\
60.0       & 800.0         & 18.8              & 61.9              & 57.3                & 36.68        \\
70.0       & 800.0         & 37.5              & 77.3              & 75.2                & 36.82        \\
80.0       & 800.0         & 34.4              & 76.2              & 74.6                & 35.23        \\
90.0       & 800.0         & 39.1              & 80.3              & 78.7                & 34.30        \\
100.0      & 800.0         & 40.6              & 83.4              & 82.1                & 33.37        \\ \addlinespace
110.0      & 800.0         & 47.3              & 78.2              & 76.0                & 29.53        \\
120.0      & 800.0         & 48.4              & 83.8              & 82.3                & 29.53        \\
130.0      & 800.0         & 48.4              & 83.3              & 81.7                & 29.00        \\
140.0      & 800.0         & 43.8              & 74.0              & 71.7                & 25.82        \\
150.0      & 800.0         & 48.4              & 82.3              & 80.9                & 26.09        \\
160.0      & 800.0         & 44.5              & 74.9              & 73.0                & 25.03        \\
170.0      & 800.0         & 48.8              & 80.3              & 78.3                & 25.56        \\
180.0      & 800.0         & 49.2              & 78.8              & 77.3                & 24.76        \\
190.0      & 800.0         & 56.2              & 87.5              & 87.3                & 23.31        \\
200.0      & 800.0         & 60.9              & 93.0              & 92.4                & 23.31        \\ \bottomrule
\end{tabular}%
\end{table*}

\newpage
\subsection{PackNet Long Sequence Capacity Analysis}
The main issue with the parameter isolation methods is the capacity left to learn new tasks. We consider two options to test the robustness of the method to this regard. First, we decrease the capacity of the model for a given number of tasks. As PackNet allocates a relative percentage of capacity for each task, this will result in the tasks getting assigned a smaller number of model parameters. However, for two models with different capacity, the rate of capacity saturation will be the same for the same pruning percentage. 
Secondly, we therefore increase the number of tasks for the same model, where we can scrutinize at what point free capacity is depleted and how this affects learning of further tasks.
We elaborate on our findings in the following.

\subsubsection{Decreased Model Capacity}
First, we can decrease the model capacity for a fixed number of 10 tasks, which we conduct in Tiny Imagenet experiments by reducing a \textsc{base} model to a \textsc{small} model. With the number of model parameters significantly reduced for this experiment, we can already observe a smaller gain in performance of $1.55\%$ for the \textsc{small} model, where PackNet has $46.68\%$ accuracy w.r.t. $45.13\%$ for best performing other method EWC. In contrast, the \textsc{base} net gains $2.23\%$ with $49.13\%$ over $46.90\%$ of best performing method MAS.
We refer to Section 6.3 for the full analysis on model capacity.

\subsubsection{Long Task Sequence Experiment} 
\begin{figure*}[]
    \center{\includegraphics[width=1.\textwidth]{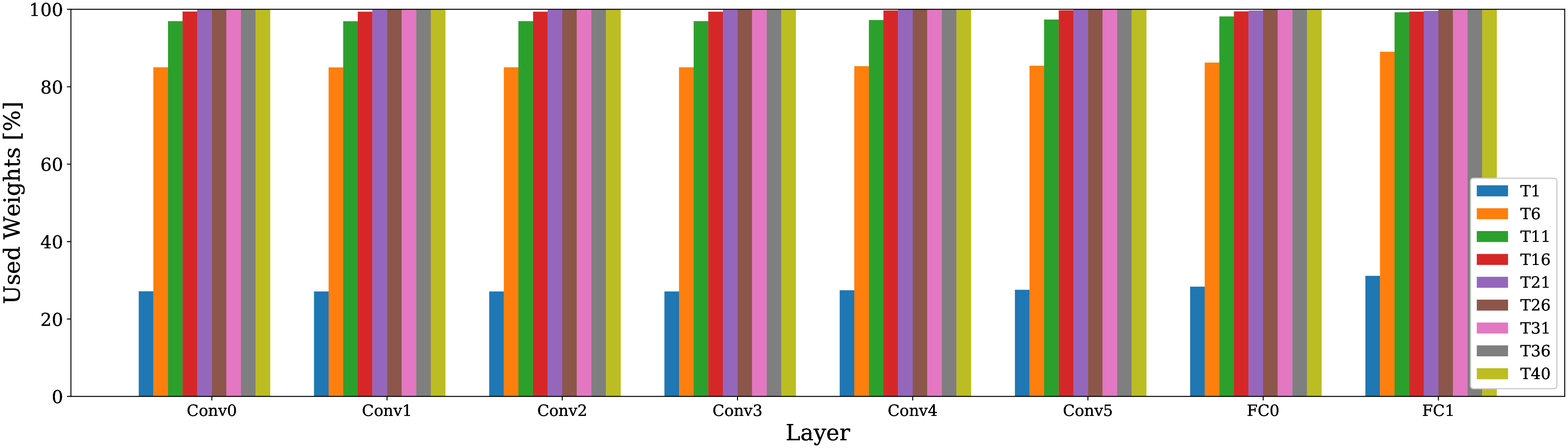}}
        \center{\includegraphics[width=1.\textwidth]{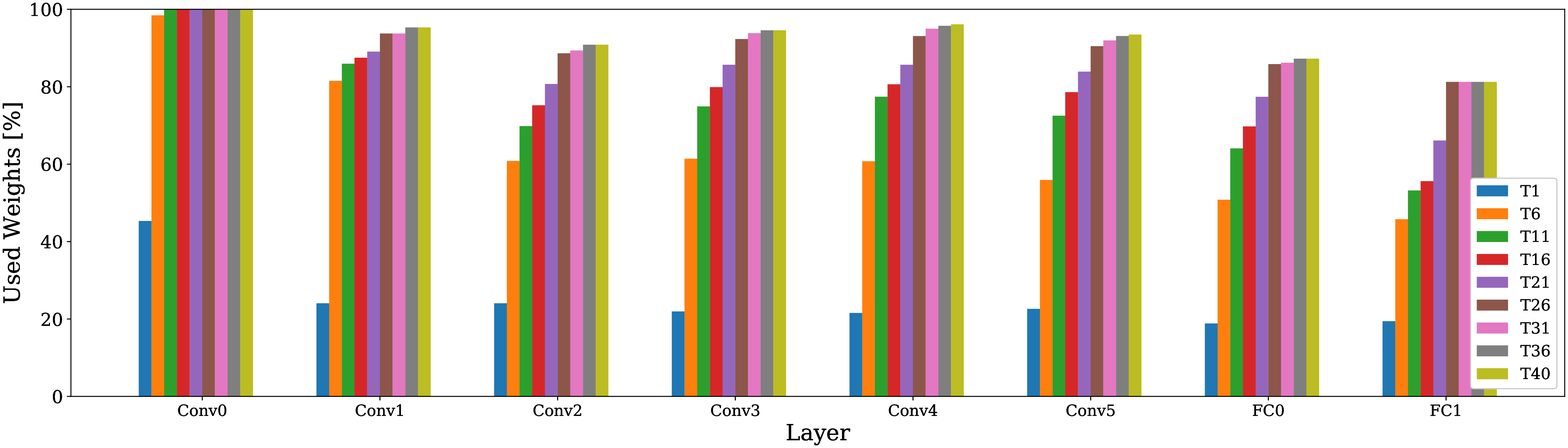}}
          \caption{\label{fig:packnet_capacity} Layer-wise cumulative PackNet (top) and HAT (bottom) weight usage with sequential task learning, for the \textbf{\textsc{base} model} on Tiny Imagenet with 40 tasks, after training each task. 
          PackNet allocates each task an  equal percentage of layer-capacity, in contrast to the asymmetrical capacity allocation observed in HAT. 
        }
  \end{figure*}

Secondly, we can increase the number of tasks, from which each task is allocated capacity to avoid forgetting. Here, the capacity of the model remains fixed, but we extend the task sequence.
\boldspacepar{Saturation in 10 tasks.} 
For example, we identify difficulties in the parameter isolation method HAT, which performs fine for the first few tasks, but deteriorates before ending the full sequence of 10 tasks in Tiny Imagenet. This already indicates and showcases the limitations of the parameter isolation methods. We include a thorough elaboration and analyze the limits of HAT in Appendix B.5, indicating an asymmetric capacity location which rapidly saturates the early layers' capacity.
PackNet seems to use the parameters more efficiently, and foremost equally allocates capacity in a layer-wise fashion. This avoids asymmetric capacity allocation as shown in Figure~\ref{fig:packnet_capacity}, with each layer capacity saturating at en equal rate.
Furthermore, the new task capacity is based only on the remaining free weights, which diminish by each learned task. Therefore, the capacity assigned to new tasks decreases, which preserves capacity for longer task sequences. 
This is a second contributing factor for PackNet not saturating in the 10 task sequences, compared to HAT.

\begin{table}[!ht]
\centering
\caption{Overview of the obtained datasets, with the balanced Tiny Imagenet, and unbalanced iNaturalist and RecogSeq dataset characteristics, and the additional Long Tiny Imagenet with fourfold number of tasks.}
\label{tab:ds}
\begin{tabular}{@{}lll@{}}
\toprule
                & \textbf{Tiny Imagenet} & \textbf{Long Tiny Imagenet} \\ \midrule
Tasks           & 10                     & 40                          \\
Classes/task    & 20                     & 5                           \\
Train data/task & 8k                     & 2k                          \\
Val. data/task  & 1k                     & 0.5k                        \\
Task selection  & random class           & random class                \\ \bottomrule
\end{tabular}%
\end{table}

\begin{table}[!ht]
\centering
\caption{Long Tiny Imagenet average accuracy (forgetting) of the final model after learning the full sequence. 
Additional adaptation to a significantly different task distribution of SVHN, with plasticity measured by \emph{accuracy for SVHN} and stability by the \emph{average accuracy and forgetting} on the final model.}
\label{tab:tiny_long}
\resizebox{\linewidth}{!}{%
\begin{tabular}{@{}llll@{}}
\toprule
\textbf{Method}               & \textbf{Long Tiny Imagenet}       & \multicolumn{2}{l}{\textbf{Long Tiny Imagenet + SVHN}} \\ \cmidrule(lr){2-2}\cmidrule(lr){3-4}
                              & \emph{avg. acc.(forgetting)} & \emph{acc. SVHN} & \emph{avg. acc.(forgetting)} \\ \midrule
\textbf{joint*}               & 79.40 (\NA)                    &       92.76             &  76.71 (\NA)                              \\
\textbf{finetuning}           & 24.15 (49.89)                  &         93.68           &               22.38 (52.08)                 \\
\addlinespace\textbf{PackNet} & 67.52 (0.00)                   &         26.62           &   66.52 (0.00)                             \\
\textbf{HAT}         & 63.30 (0.00)                   &            81.40        &        63.71 (0.03)                        \\
\textbf{iCARL (4.5k)}         & 67.10 (1.53)                   &            82.27        &        61.88 (7.08)                        \\
\textbf{LwF}                  & 64.03 (3.61)                   &         88.54           &                61.19 (6.96)                \\
\textbf{MAS}                  & 60.43 (8.45)                   &         88.41           &                35.91 (33.44)                \\ \bottomrule
\end{tabular}%
}
\end{table}

\begin{figure*}[!htbp]
    \center{\includegraphics[width=.8\textwidth]{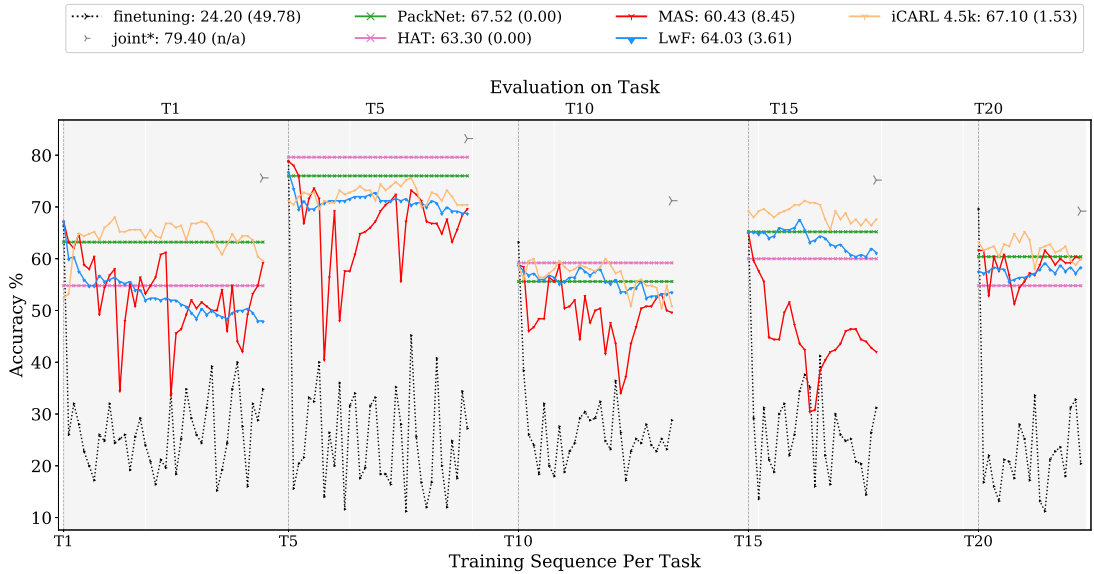}}
          \caption{\label{fig:packnet_capacity_acctrack} Evaluation for a set of task in the Long Tiny Imagenet sequence subdivided in 40 tasks.
          }
  \end{figure*}

\boldspacepar{Finding PackNet's saturation point.} Nonetheless, PackNet inevitably has a limited number of tasks to allocate capacity to.
In order to scrutinize this, Table~\ref{tab:ds} presents an extended Long Tiny Imagenet setup, which comprises a total of 40 tasks. This setup with fourfold length is designed to stress PackNet for retaining any capacity after learning the full sequence.
Surprisingly, Figure~\ref{fig:packnet_capacity_acctrack} shows the results for baselines and best performing methods, with PackNet maintaining the lead with iCaRL for average accuracies measured on the final learned model. 
We observe as from task 15 the iCaRL accuracies superseding PackNet (see midway iCaRL curves in Figure~\ref{fig:packnet_capacity_acctrack} mostly outperforming PackNet), culminating in $70.32\%$ over $68.80\%$ average accuracy after learning task 21.
Note that accuracies are higher compared to the experiments with 10 tasks, as each task now performs 5-way instead of 20-way classification.
We further investigate the used capacity in the layers for PackNet to clarify these positive findings.

\boldspacepar{PackNet Capacity Analysis.} Figure~\ref{fig:packnet_capacity} illustrates for a range of tasks the total capacity used by PackNet in each layer. After decaying twice for the first task in our framework, each task is assigned $27.1\%$ \footnote{The initial pruning rate is $90\%$, which is decayed twice in the first task, resulting in a $90\% \cdot 0.9^2 = 72.9\%$ pruning rate.} of the remaining free weights. 
Figure~\ref{fig:packnet_capacity} indicates saturation after task 11, but PackNet nonetheless has the best average accuracy and forgetting in the Long Tiny Imagenet experiment. 
We hypothesise this is due to Long Tiny Imagenet having relatively similar tasks, which makes PackNet preserve its performance by using the fixed weights learned for previous tasks. HAT saturates more rapidly in the first convolutional layer at task 6, but retains capacity for deeper layers.
To demonstrate the issue of limited capacity in the parameter isolation methods, we introduce a severe distribution shift by means of the Street View House Numbers (SVHN) dataset.

\boldspacepar{Saturation Inhibits Plasticity.} 
After learning the 40 tasks in Long Tiny Imagenet, we present the continual learner with the SVHN dataset.
The results in Table~\ref{tab:tiny_long} find PackNet to severely deteriorate in plasticity, obtaining only $26.62\%$ compared to near $90\%$ accuracy for the other methods on the new SVHN task. 
Evaluating the performance on this final model for the 41 learned tasks, PackNet's average accuracy drops only by $1\%$ due to PackNet fixing its weights once a task is learned.
However, in practice this implies PackNet not being able to learn new tasks deviating from the Long Tiny Imagenet data distribution after capacity saturation.
The parameter isolation methods usually cope with this lack of plasticity by increasing capacity allocated to the new task. In the case of no remaining capacity, this experiment shows that failure to learn the new task is inevitable.
Our analysis of HAT in Appendix~B.5 also supports this for the \textsc{deep} model saturating in the Tiny Imagenet sequence.

\boldspacepar{Hyperparameter decay.} Learning the SVHN task for HAT the hyperparameters decay from $s=800, c=2.5$ to ${s=400}$, ${c=0.625}$ to attain the finetuning accuracy, with HAT obtaining $81.40\%$ accuracy. In contrast to PackNet, the capacity allocation in HAT behaves asymmetric with no capacity left in the early layers, while the latter layers retain $18.8\%$ capacity (see bottom Figure~\ref{fig:packnet_capacity} for FC1). PackNet has no capacity left at all, while HAT can use the final layers capacity.
In summary, the good performance for HAT may be due to efficient reuse of the early layers, while able to reuse the final layers to learn the SVHN task. However, the RecogSeq and iNaturalist tasks differ too much in low-level features, where the asymmetric capacity allocation is a pressing disadvantage, causing cumbersome learning with HAT.

In our framework, regularization methods decrease the regularization strength $\lambda$ when having trouble to adapt to the new task. MAS decays from $\lambda=1.5$ to $\lambda=0.75$, hence sacrificing stability of the previously learned knowledge to learn the SVHN task. Table~\ref{tab:tiny_long} reports $88.4\%$ accuracy for the new task, but severely drops in average performance over all learned tasks from $60.4\%$ to $35.9\%$.

\end{document}